\PassOptionsToPackage{table,prologue}{xcolor}
\documentclass[acmsmall]{acmart}

\usepackage{hyperref}
\usepackage{graphicx}
\usepackage{array}
\usepackage{svg}
\usepackage{varwidth}
\usepackage{tabularx}
\usepackage{booktabs}
\usepackage{cite}
\usepackage{lscape}
\usepackage{afterpage}
\usepackage{multirow}
\usepackage{makecell}
\usepackage{longtable}
\usepackage{makecell}
\usepackage[table]{xcolor}
\usepackage{doi}
\setcitestyle{maxcitenames=1}
\hypersetup{colorlinks=true, linkcolor=blue}

\newcolumntype{T}{>{\tiny}l} 
\newcolumntype{H}{>{\Huge}l} 
\newcolumntype{M}{>{\begin{varwidth}{2cm}}l<{\end{varwidth}}} 

\AtBeginDocument{%
  \providecommand\BibTeX{{%
    \normalfont B\kern-0.5em{\scshape i\kern-0.25em b}\kern-0.8em\TeX}}}

\setcopyright{rightsretained}
\copyrightyear{2023}
\acmDOI{XXXXXXX.XXXXXXX}

\begin{document}

\title{Complex QA \& language models hybrid architectures, Survey}

\author{Xavier Daull}
\affiliation{%
\institution{Naval Group, Toulon Université, Aix Marseille Univ, CNRS, LIS}
  \city{}
  \country{France}
}

\author{Patrice Bellot}
\affiliation{%
  \institution{Aix Marseille Univ, CNRS, LIS, Marseille}
  \city{Marseille}
  \country{France}}

\author{Emmanuel Bruno}
\affiliation{%
 \institution{Toulon Université, Aix Marseille Univ, CNRS, LIS, Toulon}
 \city{Toulon}
 \country{France}}

\author{Vincent Martin}
\affiliation{%
  \institution{Naval Group}
  \country{France}}

\author{Elisabeth Murisasco}
\affiliation{%
  \institution{Toulon Université, Aix Marseille Univ, CNRS, LIS, Toulon}
  \city{Toulon}
  \country{France}}

\addtocontents{toc}{\protect\setcounter{tocdepth}{-1}}
\begin{abstract}
This paper reviews the state-of-the-art of large language models (LLM) architectures and strategies for "complex" question-answering with a focus on hybridization.
LLM based chatbot services have allowed anyone to grasp the potential of LLM to solve many common problems, but soon discovered their limitations for complex questions.
Addressing more specific, complex questions (e.g., "What is the best mix of power-generation methods to reduce climate change ?") often requires specialized architectures, domain knowledge, new skills, decomposition and multi-step resolution, deep reasoning, sensitive data protection, explainability, and human-in-the-loop processes.
Therefore, we review:
(1) necessary skills and tasks for handling complex questions and common LLM limits to overcome;
(2) dataset, cost functions and evaluation metrics for measuring and improving (e.g. accuracy, explainability, fairness, robustness, groundedness, faithfulness, toxicity...);
(3) family of solutions to overcome LLM limitations by (a) training and reinforcement (b) hybridization, (c) prompting, (d) agentic-architectures (agents, tools) and extended reasoning.

\end{abstract}



\maketitle

\pagebreak
\small
\addtocontents{toc}{\protect\setcounter{tocdepth}{2}}

\section{Introduction}
Research in the field of question answering could route from earliest examples of AI system designed to answer questions like ELIZA~\citep{weizenbaumELIZAComputerProgram1966} developed at MIT in the 1960s; or engineering approaches to solve complex problems; or foundational research in the psychology field on how people understand and interpret language, prioritize and focus on relevant information in context~\citep{cuadraOPENINGBLACKBOX1967}, investigate and make decision, retrieve and use information from memory to answer~\citep{normanMemoryKnowledgeAnswering1972}, innovate...
Even Socrates' philosophical approach to questioning and critical thinking has recently been directly used to improve training~\citep{pagnoniSocraticPretrainingQuestionDriven2022}, common sense reasoning~\citep{jungMaieuticPromptingLogically2022}, or helping to solve some complex questions.

When a question or problem is too complex to be solved as it is, such as ``~how does the concept of personal freedom vary between different cultures and societies ?~'', a common strategy is to break it down into solvable questions, and then combine solutions to provide an overall answer if there is consensus, or the possible alternatives and nuances.
We aim to answer complex questions, or problems formulated as a question,  which are non-factoid and so require decomposition (multi-step), multi-source of knowledge combination, higher order reasoning or tasks resolution.
All of these may vary a lot depending on the field and question.
LLMs have demonstrated their ability to outperform an average human on complex QA tasks across knowledge fields~\citep{kazemiBIGBenchExtraHard2025}, and can mimic methods or uncover appropriate methods for a given problem.
However, even most advanced systems will fail on some basic questions~\citep{borjiCategoricalArchiveChatGPT2023}, or could assert totally false or biased knowledge without any caution.
This can seriously impact the credibility of a system by a human.
Involving humans in the question-answering loop, strengthening training, hybridizing with third parties like programs, symbolic AI or other, can greatly improves those models and help ensure ethical and safe outcomes.

To be able to build those efficient hybrid systems, properly trained and aligned to human expectations in order to better solve increasingly complex problems, it is thus necessary to precisely know strengths and limitations of each language models alone or in collaboration.

Therefore we use benchmarks and insights from collective papers as consensus baselines.
Latest evaluations of large community evaluation project HELM~\citep{liangetal.HolisticEvaluationLanguage2022}, and BIG~\citep{bigetal.ImitationGameQuantifying2022} focus on evaluating, democratizing and improving LLM capabilities, particularly on question answering and tasks that will also be useful for more complex questions.
HELM provides a comprehensive analysis and benchmark for evaluating global strengths and weaknesses of reference large language models across a range of different scenarios and complementary metrics.
Training large language models is done by resource-rich organizations and are mostly impossible to train by any individual public research group.
BigScience\footnote{\url{https://bigscience.huggingface.co/}}, a collaboration of hundreds of international researchers, trained on the French government-funded Jean Zay supercomputer during 3.5 months the largest open source multilingual language models in July 2022 called BLOOM and shared all models, codes and results in the BLOOM paper~\citep{bigscienceetal.BLOOM176BParameterOpenAccess2022}.
BIG spots difficult tasks to better assess current skills and limitations of language models.

To overcome identified limitations and tackle more specific or complex questions, after those benchmarks and insights, we review hybrid architectural patterns solutions, training, prompting techniques, and reinforcement strategies to acquire the necessary knowledge, skills (general abilities), tasks (goal oriented actions), methods and human alignment to answer complex questions.

\subsection{Structure of the paper}
This survey covers the main topics related to complex question answering systems.
In \autoref{sec_core_concepts}, we present typical process and architectures to answer questions, and new approach brought by large language models (LLM) and transformers.
Next, we delve into complex QA definition, required tasks \& skills, limitations to overcome (\autoref{sec_skills_tasks_limits}) to answer complex questions.
For each limitation we link to potential resolution strategies presented later.
In \autoref{sec_evaluation}, we review the evaluation metrics, methods and datasets of the scientific community used to assess current tasks, skills and limitations of LLMs, or to further develop them through training.
We then review the complementary resolving strategies which could be combined to solve complex QAs for the target usage.

We first explore training techniques in \autoref{sec_training}, from \emph{pre‑training}, \emph{mid‑training} to \emph{post‑training}, as well as methods for dealing with lack of data, poor quality, learning to reason, and adaptation to new tasks and domains.

Second, in \autoref {sec_architectural_patterns}, we review and classify different hybridization architectural patterns with their pros and cons to augment LLM.
In addition, we present a unified section on modern \emph{agentic meta-architectures} together with \emph{inference-time strategies} (see \autoref{sec_agentic_and_reasoning}).

Third, in \autoref{sec_prompting}, we review prompting techniques and how it can also be used to decompose complex questions down to solvable tasks; we also formalize \emph{in-context learning}, \emph{context engineering} and \emph{protocol engineering} for principled prompt-time assembly of instructions, evidence, tools and memory (\autoref{sec_context_protocol}).

Fourth, in \autoref{sec_agentic_and_reasoning}, we cover \emph{agentic meta-architectures (agents, tools)} \emph{and} the \emph{inference-time strategies} they invoke to allocate extra “reasoning-time’’ on hard queries;
\emph{preference alignment and reinforcement} is treated within training (\autoref{sec_posttraining_alignment}).
We introduce \emph{reasoning-time (inference-time) compute} as a strategy to trade compute for quality on difficult queries (\autoref{sec_reasoning_time}).

Finally, \autoref{sec_limits_and_research} highlight tougher challenges, partial solutions that have been identified, and research avenues to overcome them, including safety and multi‑sensitivity usage.
All along the survey, we provide an extensive bibliography for readers who wish to delve deeper into the subject matter (\autoref{sec_biblio}).

\subsection{Contribution}
This survey provides :
\begin{enumerate}
\item[--] a systematic review \& analysis of literature on complex QA with LLM, including an enriched \textbf{definition} (\autoref{CQA_definition}) and \textbf{taxonomy} (\autoref{sec_skills}, \autoref{CQA_Taxonomy}), and an extensive bibliography of the field.
\item[--] a qualitative analysis of the skills, tasks, and limitations of LLMs (\autoref{sec_skills_tasks_limits}) aimed at better framing complex QA requirements and complexity.
\item[--] an overview of evaluation metrics, methods, datasets and SOTA (\autoref{sec_evaluation}) to better evaluate skills \& tasks, and estimate LLM limits and strategies.
\item[--] a \textbf{classification and aggregation of hybridization architectural patterns} (\autoref{sec_hybridLLMpatterns}) that can augment LLM and overcome their limitations in complex QA.
\item[--] a set of \textbf{resolving strategies to combine} (training: \autoref{sec_training}, hybridization: \autoref{sec_hybridLLMpatterns}, prompting: \autoref{sec_prompting}, post-training preference alignment: \autoref{sec_posttraining_alignment}, and inference-time strategies within agentic meta-architectures: \autoref{sec_agentic_and_reasoning}).
\item[--] a list of major research challenges (\autoref{sec_limits_and_research}) and a focus on some blind spots (\textsl{i.e.} data multi sensitivity).
\end{enumerate}

\subsection{Survey methodology}

To build this survey, \textbf{first}, we  collected surveys in the last two years related to ``~complex question answering~'' or ``~complex problem solving~'' and ``~language models~''  that we cite throughout the article.
From these elements, we systematically extracted the major concepts and their plans (table of contents).
We fused all their plans into one to ensure a complete coverage of major concepts and adopt a similar methodology.

\textbf{Second}, we gathered:
\begin{enumerate}
\item[--] latest challenges from \href{http://nlp.uned.es/clef-qa/}{CLEF QA (nlp.uned.es)}, \href{https://research.nii.ac.jp/ntcir/}{NTCIR (research.nii.ac.jp)}, \href{https://semeval.github.io/}{SemEval (semeval.github.io)} and TREC (microsoft.github.io/msmarco) related to question answering;
\item[--] the list of the major conferences; then, list research papers from latest edition of conferences \href{https://sigir.org/}{SIGIR (sigir.org)}, \href{https://nips.cc/}{NeurIPS (nips.cc)}, \href{https://naacl.org/}{NAACL (naacl.org)}, \href{https://emnlp.org/}{EMNLP (emnlp.org)}, \href{https://iclr.cc/}{ICLR (iclr.cc)}, \href{https://aaai.org/}{AAAI (aaai.org)}, \href{https://www.ijcai.org/}{IJCAI (ijcai.org)}, CIKM, \href{https://www.kdd.org/}{SIGKDD (kdd.org)}, \href{https://www.wsdm-conference.org/}{WSDM (wsdm-conference.org)} about  ``~complex question answering~'' and ``~language models~'';
\item[--] research publications from influential organizations in the field: \href{https://www.deepmind.com/research}{Deepmind}, \href{https://openai.com/publications/}{OpenAI}, \href{https://research.google/pubs/}{Google}, \href{https://www.microsoft.com/en-us/research/publications/}{Microsoft}, \href{https://research.facebook.com/publications/}{Meta}, and \href{https://www.anthropic.com/research}{Anthropic} related to "question answering" and "language models".
\item[--] \textbf{living evaluation sources} that now shape SOTA perception and track reasoning progress like \emph{Arena‑Hard} (and Chatbot Arena Elo), \emph{HELM} updates, \emph{LiveBench}, and harder reasoning sets such as \emph{Big Bench Extra Hard}, \emph{Humanity's Last Exam}~\citep{phanHumanitysLastExam2025}, \emph{MMLU‑Pro}, \emph{GPQA}. These complement frozen test sets for reproducible comparisons.
\end{enumerate}
From these documents, we clustered major LLM limitations which are listed in \autoref{sec_LLMlimits} and solutions in sections \ref{sec_training}, \ref{sec_hybridLLMpatterns}, \ref{sec_prompting}, \ref{sec_ImprovementLoop_and_kg_capitalization}, \ref{sec_limits_and_research}.
\newline

\textbf{Third}, we enriched our bibliography using a search engine of scientific articles to identify all the articles published during the last four years on the main subject of this investigation (\small{\textit{search queries: "complex question answering" AND "language model"; "question answering" AND "language model architecture"; "question answering" AND "language model" AND "hybrid architecture"; "hybrid language models" OR "hybrid language model" OR "hybrid neural language models"; "language model" "hybrid architecture"}}).

The bibliography collected in all the previous steps was then used throughout this survey to ensure that we summarize the main concepts at the state-of-the-art by automatically detecting articles semantically close (subscription to "zeta alpha" and "semantic scholar").

Last but not least, we searched the most cited papers in this bibliography and extended with recent connected papers or some historical papers often cited, and investigated relevant citations.
Through this research, we identified three recent research papers (HELM capabilities, and BIG) each involving hundreds of researchers from many organizations, so we decided to use them as a baseline or major reference in this study.

\section{Core concepts \& architectures}\label{sec_core_concepts}

\subsection{Question answering typical pipeline and modular approaches}
\begin{figure}
\captionsetup{position=top}
\includegraphics[width=\linewidth]{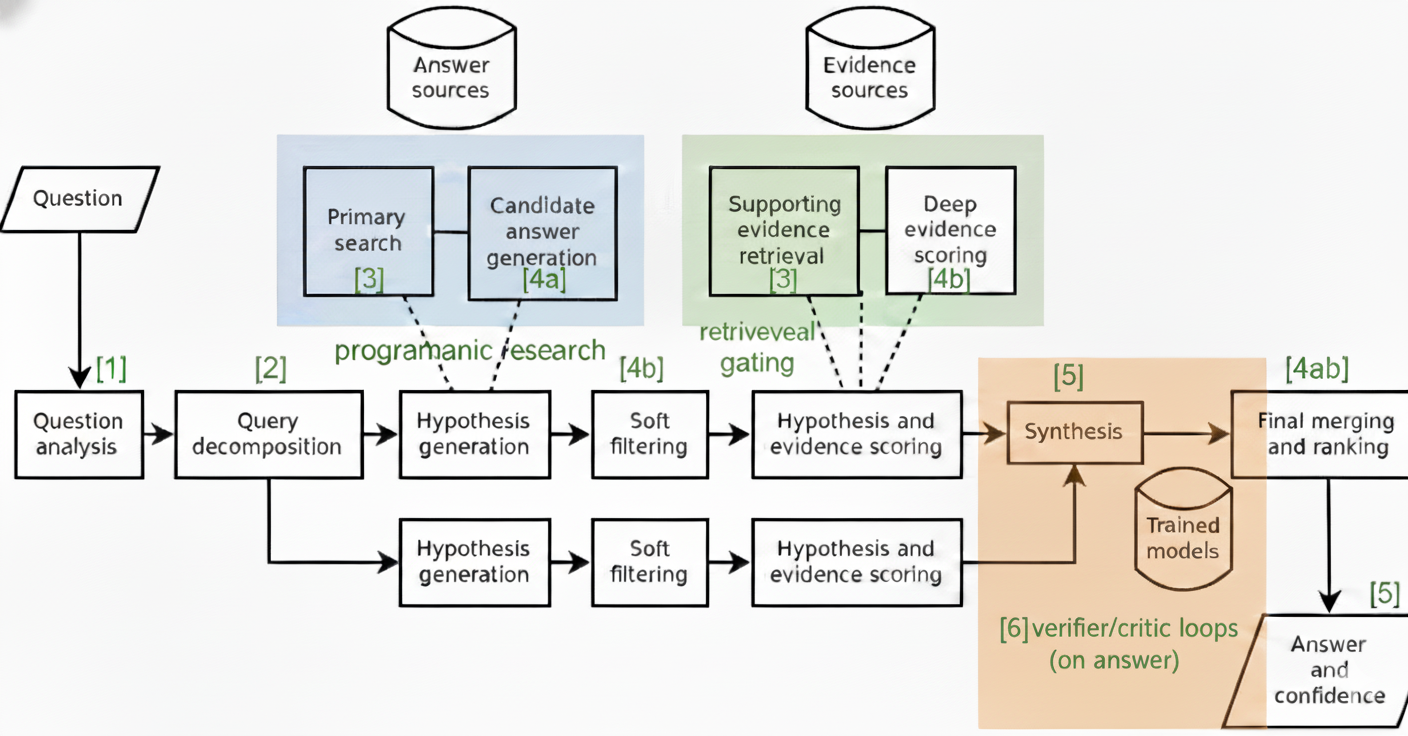}
\caption{IBM DeepQA Architecture (2010)~\citep{ferrucciBuildingWatsonOverview2010} CQA pipeline steps "patched" to reflect new practice with verifier/critic loops, retrieval gating, and programmmatic research actions.}
\label{DeepQA Architecture}
\end{figure}

From question capture and refinement, to answer generation and knowledge capitalisation, question answering (QA) or complex question answering (CQA) pipeline can follow a variable number of steps depending on architecture and features.
Some steps are explicit and well separated, some can be implicit and fused with others in the same model operation.
However, we can identify most frequent steps and options.
The "IBM DeepQA Architecture" (see figure~\ref{DeepQA Architecture}) seems a dated architecture compared to some current end-to-end neural language models but it defines clearly some major steps; in practice, since 2024, many research pipelines \emph{add} explicit \textbf{verifier/critic loops} and \textbf{retrieval gating}, and support \textbf{tool use} (function calling) for programmatic research actions.

\begin{enumerate}
\item \textbf{Question understanding} and analysis, it can include question and context refine, parsing and understanding the question, context, and intent (for task identification).
It can also embed a dialogue management to interact with the user and understand the conversation context and state.
\item \textbf{Query construction} with optional \textbf{decomposition} of complex questions and multi-step queries.
\item \textbf{Information retrieval (IR)} and optional knowledge expansion, to add new knowledge to the system.
\item (a) \textbf{Information extraction}, and (b) \textbf{evaluation, scoring, ranking, filtering}, \textbf{verification/attribution} (\emph{judge} models, quote checks) and \textbf{retrieval gating} (e.g. self-corrective RAG).
\item \textbf{Answer generation}, natural language or defined format (e.g. language program, table, ...)
\item \textbf{Feedback loop \& new knowledge capitalisation}: learning and improving from users and models feedback, plus storing linked and generated knowledge for improving answering skills.
\end{enumerate}
This process is only a baseline as complex questions answering can be a dynamic and progressive process, and can also be collaborative.
Architectures of QA systems have importantly evolved recently with the arrival of transformers architectures and large language models.
We first quickly review typical modular architectures, then the transformers with large language models and, later, the hybrid architectures.
Will we go to gigantic knowledge models or more complex and composed architectures, maybe in network of smaller specialized models and other components ?

Typical architectures of QA systems before transformers were modular approaches such as:
\textbf{Rule-based systems} using a set of predefined rules to answer questions; \textbf{Retrieval-based systems} using a search engine or database to find answers to questions; \textbf{Information extraction systems} using natural language processing (NLP) to extract relevant information from text documents and often leverage information retrieval (IR) systems; \textbf{Knowledge-based systems} storing and retrieving information from a knowledge base; \textbf{Case-based reasoning systems} using a database of previously solved problems to find solutions to new questions; \textbf{Hybrid architectures} could assemble the some of the modules listed (e.g. "IBM DeepQA Architecture, 2010"~\citep{ferrucciBuildingWatsonOverview2010}) to deliver a more advanced QA system integrated with natural language models for understanding initial question for example.

\subsection{LLM with transformers evolution}
The emergence of deep feedforward layers, allowed to learn and infer a much wider range of relationships with inputs; Then the raise of attention mechanism, allowed a model to selectively focus on certain parts of the input for better understanding and contextualizing. It led to language models surpassing humans on certain tasks.
We can group current transformer based language models in three types:
\begin{description}
    \item [Encoders only](e.g. BERT~\citep{devlinBERTPretrainingDeep2019}, RoBERTa~\citep{liuRoBERTaRobustlyOptimized2019}) encode a sequence of a text input into a rich representation (vector embedding) which can be used by a task-specific model or function for classification, named entity recognition (NER), semantic similarity measure used in IR and QA or topic modeling. This is often called bidirectional attention because they can take the context of the words before and after the target word analyzed, which allow them to perform better on some tasks.
BERT is one of the most well-known encoder-only models, and RoBERTa is an optimized version of BERT.
    \item [Decoders only] (e.g. GPT family~\citep{zongSurveyGPT32022, brownLanguageModelsAre2020}) complete an input sequence (mostly text prompt) by the most probable next words (generation).
This left to right generation can be very rich like writing a story or answering a question (e.g. used by ChatGPT).
The input prompt can be formatted with appropriate instructions (prompt \& instructions engineering) to use the generated text as a task-specific model (e.g. classification, summarization, decomposition...).
This is often called causal or autoregressive attention (non-causal decoders exist but has limited adoption in the literature).
GPT family of models are one of the most well-known decoder-only models, known for their ability to generate human-like text.
    \item [Encoders-Decoders] (e.g. T5~\citep{raffelExploringLimitsTransfer2020}, BART~\citep{lewisBARTDenoisingSequencetoSequence2019}) encode the input text and decode it into a different form (text-to-text mapping), they are suitable for translation, summarization, or generating a response to a question.
They consist of both an encoder and a decoder, where the encoder generates a fixed-size representation of the input text, and the decoder generates the output text.
T5 is known for its multi-task ability; BART is mainly used for text generation and summarization.
\end{description}

\noindent\textit{Recently}, closed-source providers, as well as open-weight and open-source community efforts focus on practical inflections for complex QA: (i) \textbf{long-context} models (hundreds of thousands to > 1M tokens) make multi-document “full book’’ reasoning far more feasible; (ii) \textbf{multimodal} models unify text, image, audio and diagrams in a single loop; (iii) \textbf{reasoning-centric} models and practices (deliberate \emph{think-then-answer} decoding, plan/verify loops) substantially improve math, science, and tool-using tasks; (iv) \textbf{mixture-of-experts (MoE)} designs regain traction to balance quality/cost; and (v) \textbf{agentic tool use} via function calling is now default in many deployments.

\section{Analyzing: complexity, skills, tasks, and limits}\label{sec_skills_tasks_limits}
In order to design a system able to answer complex questions, we first propose to analyze targeted questions complexity, identify required skills and tasks, as well as limitations to handle.
This analysis allows to properly define the problem to be solved gradually and constraints to integrate in order to compose among the different complementary solving approaches further reviewed (training: section \ref{sec_training}, hybridization: section \ref{sec_hybridLLMpatterns}, prompting: section \ref{sec_prompting}, experience: section \ref{sec_ImprovementLoop_and_kg_capitalization}).
This analysis could also be done "a posteriori" if a system fails to properly answer in order to better characterize or identify the causes.

\subsection{What are complex questions ?}\label{CQA_definition}
Complex questions can be defined as those that involve multiple factors of complexity, requiring higher levels of cognitive processes and domain-specific knowledge to answer accurately.
\citet{ullrichUsingBloomsTaxonomy2021} suggests using Bloom's taxonomy to evaluate question complexity by integrating required knowledge (ranging from easier to harder: factual, conceptual, procedural, metacognitive) with cognitive processes (ranging from easier to harder: remember, understand, apply, analyze, evaluate).
For instance, a simple factual question such as "What is the capital of France?" requires only basic recall, while a more complex question like "What are the main causes of climate change and their potential solutions?" demands an understanding of multiple concepts and the ability to evaluate different solutions.

We think that a question complexity is also highly dependent on users' or systems' expertise involved in the response.
We propose to assess the complexity of questions by also considering the main difficulties and efforts required for an LLM to solve them:

\begin{itemize}
    \item \textbf{skills and knowledge} required (section \ref{sec_skills}): simple memorization or higher reasoning (e.g. evaluation, constraints solving, deduction, induction, abduction), single or multiple types of logic; prior domain-specific knowledge, retrieve and process one easily accessible piece of information or multiple rare information, reason over long distance between pieces of information to combine; expected answer format and explanation; ambiguity and nuances to handle (especially in non-factoid questions).
    and the need for specific decomposition and multi-step resolution.
    \item \textbf{new or challenging tasks} to solve like specific decomposition, and multi-step resolution (section \ref{sec_tasks}).
    \item major \textbf{limitations of LLMs to overcome} in this context (sections \ref{sec_LLMlimits} and \ref{sec_limits_and_research}).
    \item difficulty in designing \textbf{appropriate metrics and sufficient datasets} to measure LLM skills and knowledge performance (section \ref{sec_evaluation}).
    \item \textbf{training effort} to develop those skills \& tasks (section \ref{sec_training}) in a model, or in different \textbf{hybrid models} (section \ref{sec_hybridLLMpatterns}) to align or train end-to-end to solve questions in a coherent way.
    \item complexity to \textbf{engineer robust prompt questions} on trained models such as additional context and instructions required (section \ref{sec_prompting}), and then \textbf{decompose questions down to solvable tasks}.
    \item progressive \textbf{reinforcement and knowledge capitalization}, or system's experience learning, to solve targeted complex questions (section \ref{sec_ImprovementLoop_and_kg_capitalization}).
\end{itemize}

\afterpage{%
  \clearpage
\begin{landscape}

\begin{figure}[ht]
\centering
\includegraphics[width=1.0\linewidth]{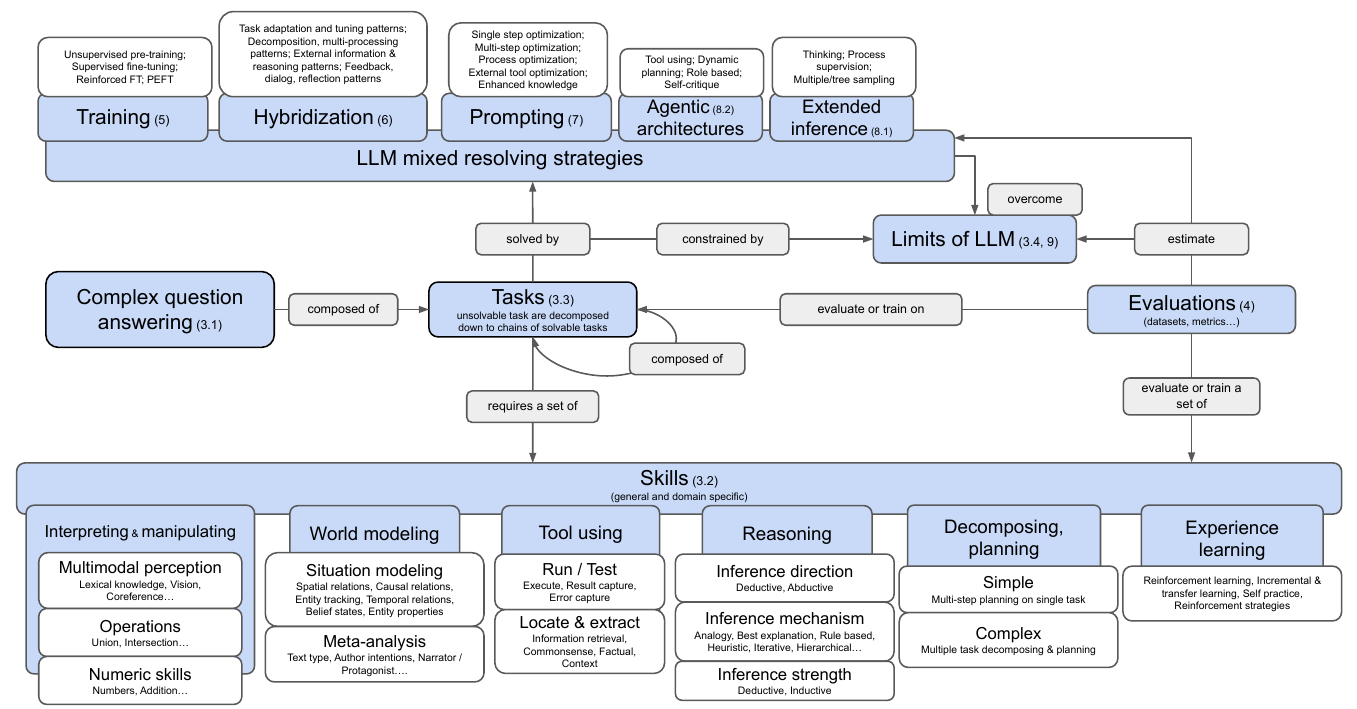}
\caption{QA/CQA taxonomy links skills, tasks and limits of LLMs to complementary resolving strategies (training, hybridization, prompting, agentic architectures, extended reasoning-time), as well as evaluation datasets \& metrics (skills inspired by \citet{rogersQADatasetExplosion2022}) - Each concept is provided with reference to section in this paper.}
\label{CQA_Taxonomy}
\end{figure}

\end{landscape}
  \clearpage
}

\subsection{Skills}\label{sec_skills}

Considering the QA/CQA standard pipeline presented in introduction, a task of question answering requires different complementary skills in the domain of machine reading comprehension (MRC), information retrieval (IR), and also knowledge capitalisation and reinforcement.

We leverage the QA skills taxonomy from \citet{rogersQADatasetExplosion2022} which we augmented with the "Experience learning" skill concept (see figure \ref{CQA_Taxonomy}), important for a CQA system.
This later has shown to be a major skill to enable calibration or alignment to intent and values~\citep{baiConstitutionalAIHarmlessness2022}, and continuous improvement by usage~\citep{chiuKnowledgeGroundedReinforcementLearning2022}.

\subsubsection{Interpreting \& manipulating input}
Like humans, machines should capture the meaning of the individual constituent elements of the input (words, numbers) and the global syntax and semantic, and manipulate them in the context of the task and in respect to the language and other shared system (e.g. mathematics).
This  requires a set of skills, including: 
\begin{itemize}
    \item \textbf{Linguistic skills} - e.g. recognizing word categories and meaning, translating, understanding the context, relationships and implications of words and phrases; it might be decomposed into syntactic, grammatical, and semantic skills.
    \item \textbf{Numeric skills} - e.g. performing calculations, dealing with precise and imprecise numbers.
    \item \textbf{Operation on sets} - e.g. selecting, combining, intersection, operating on elements of a set of input (e.g. Alice, Bob and Charlie are in the room. Bob goes out. Who are the persons in the room?);
\end{itemize}

\subsubsection{Information Retrieval}
It can be summarized as determining whether an answer exists, if yes, to look for it and provide most useful information :
\begin{itemize}
    \item \textbf{Answerability}: ability to identify whether a given query is a valid question and can be answered with provided information.
Optionally identify additional information to correctly answer.
    \item \textbf{Where to look for the required knowledge?}: ability to identify the correct source of knowledge to get the best answer.
 It the required knowledge for the answer is in the question, process is to extract the good piece of information.
Otherwise, we need to know if it is a precise fact or non factual, then where to look for it.
Additionally, a proper answer may require common sense and potential domain information.
\end{itemize}

\subsubsection{Inferring \& reasoning}
Inferring and reasoning can be summarized as the process of drawing logical conclusions from available facts or other premises.
Inferring is used in language models to understand a text, and generate responses to questions posed.
There are three main aspects to inference in language models:
\begin{itemize}
    \item \textbf{Inference Strength}: could draw general conclusions from specific facts (inductive), or draw specific conclusions from general facts (deductive).
    \item \textbf{Inference Mechanism}: draw conclusions from a comparison of two or more elements (analogy), draw conclusions based on the best explanation for a given situation (best explanation)...
    \item \textbf{Inference Direction}: conclusion follows necessarily from the premises or from general to specific (Deductive), conclusion is reached through a process of elimination or reasoning from the specific to the general (abductive).
\end{itemize}

\subsubsection{World modeling}
It can be summarized as  the ability to understand and make decisions based on the understanding of the world.
It is a complex type of question answering skill that requires understanding of physical and mental states, as well as relationships between them.
It involves the following categories: 
\begin{itemize}
    \item \textbf{Spatial} reasoning: understand and reason about objects and their locations in space.
    \item \textbf{Temporal} reasoning: understand and reason about event order, event attribution to time, script knowledge, event duration, temporal commonsense knowledge, factoid/news questions with answers where the correct answers change with time, temporal reasoning in multimodal setting.
    \item \textbf{Belief states}: understand and track beliefs, opinions, and mental states.
    \item \textbf{Causal relations}: understand and reason about the cause-and-effect relationships between events.
    \item \textbf{Other relations between events}: understand and reason about relationships between events, such as sub-events, conditionals, and counterfactuals.
    \item \textbf{Entity properties and relations}: properties of characters, physical properties, numerical properties, social interactions.
    \item \textbf{Tracking entities}: understand and track entities over time, across locations, in co-reference chains.
\end{itemize}

\subsubsection{Decomposing, multi-step}
Complex questions require decomposition down to solvable tasks and resolution in the best chain of action steps.
\textbf{Simple question} may use multi-step resolution but all necessary knowledge are located in one place.
\textbf{Complex questions} rely on several knowledge, necessitating the combination of information across sentences, paragraphs, documents, or other modalities.
It also includes questions that require a combination of context and world knowledge.
It can be even broader than simply combining several facts, and could also be taken as combining the “skills” from different dimensions and different methods of resolution.
This decomposition and multi-step resolution can be resolved inside a model having these skills and all other necessary for the question, or distributed across multiple components.

\subsubsection{Experience learning}
A complex QA system should be able to permanently improve itself through: \textbf{reinforcement} by aligning answers to target intent, format, method, values expectations with solutions which could vary with requester person (e.g. knowledge, culture...) or system; \textbf{capitalization} by integrating new knowledge generated or linked in order to improve knowledge enabling to solve more complex problems.
Experience skills   could be classified under meta-analysis in world modeling skills but it may not fully capture self-modeling, self-practice or  integration of external feedback, incremental learning towards a coherent optimization of all skills.

\subsection{Tasks}\label{sec_tasks}
A task of complex question answering could be solved in one inference task incorporating all the skills viewed in previous section, or subdivided in several sub-tasks, planned, each with a set of skills and, maybe, different domains.

\subsubsection{Integrated (C)QA task}
In this case, the CQA system answers from question using only one inference in the model but could include multi-step reasoning inside the model.
LLM (large language models) should therefore embed all the necessary skills and knowledge for interpretation \& manipulation, information retrieval, world knowledge, reasoning \& inference, decomposition \& multi-step resolution.
Otherwise, the model should be further trained with adapted datasets to acquire those new skills and knowledge, or rely on task decomposition and LLM hybridation.
\citet{bigetal.ImitationGameQuantifying2022} and \citet{liangetal.HolisticEvaluationLanguage2022} provide an overview of limits of integrated QA/CQA.
To make these ceilings concrete, Table~\autoref{tab:big-bench-hard} contrasts human average and expert performance with the best open models on representative hard BBH tasks.

\begin{table}[ht]
\centering
\small
\resizebox{\textwidth}{!}{%
\begin{tabular}{llrrrrl}
\toprule
Domain & Task & Human & Human & Top open LLM & Top open LLM \\
 &  & average & expert & Score (\%) & Name \\
\midrule
\multirow{6}{*}{Logic} & boolean\_expressions & 84.0 & \textbf{100.0} & 96.0 & T3Q-qwen2.5-14b-v1.0-e3 \\
& web\_of\_lies & 71.0 & 92.0 & \textbf{93.2} & calme-3.2-instruct-78b \\
& formal\_fallacies & 67.2 & 90.0 & \textbf{98.0} & Llama-3.2-1B-Instruct \\
& logical\_deduction\_three\_objects & 73.6 & 94.0 & \textbf{98.4} & internlm2\_5-20b-chat \\
& logical\_deduction\_five\_objects & 50.8 & 78.0 & \textbf{82.0} & T3Q-qwen2.5-14b-v1.0-e3 \\
& logical\_deduction\_seven\_objects & 34.0 & 54.0 & \textbf{86.0} & Llama-3.2-1B-Instruct \\
\midrule
\multirow{5}{*}{Algorithmic} & tracking\_shuffled\_objects\_three\_objects & 62.8 & \textbf{86.0} & 79.6 & Llama-3.2-1B-Instruct \\
& tracking\_shuffled\_objects\_five\_objects & 34.0 & 56.0 & \textbf{60.4} & Llama-3.2-1B-Instruct \\
& tracking\_shuffled\_objects\_seven\_objects & 23.2 & 44.0 & \textbf{54.8} & Llama-3.2-1B-Instruct \\
& navigate & 54.0 & 74.0 & \textbf{87.2} & test-2.5-72B \\
& object\_counting & 77.0 & \textbf{96.0} & 89.2 & Llama-3.2-1B-Instruct \\
\midrule
\multirow{5}{*}{Linguistic} & disambiguation\_qa & 76.2 & 89.2 & \textbf{97.6} & Llama-3.2-1B-Instruct \\
& hyperbaton & 73.1 & 90.0 & \textbf{99.6} & Llama-3-Refueled \\
& ruin\_names & 70.0 & 92.0 & \textbf{97.6} & internlm2\_5-7b-chat \\
& salient\_translation\_error\_detection & 68.8 & 88.0 & \textbf{97.6} & Llama-3.2-1B-Instruct \\
& snarks & 78.4 & 91.0 & \textbf{96.6} & Llama-3.2-1B-Instruct \\
\midrule
\multirow{6}{*}{Reasoning} & causal\_judgement & 63.8 & 82.2 & \textbf{87.7} & Llama-3.2-1B-Instruct \\
& date\_understanding & 74.0 & \textbf{95.6} & 94.8 & Llama-3.2-1B-Instruct \\
& movie\_recommendation & 70.8 & 94.0 & \textbf{100.0} & BunderMaxx-1010 \\
& reasoning\_about\_colored\_objects & 81.2 & \textbf{97.2} & 90.8 & ultiima-72B-v1.5 \\
& sports\_understanding & 69.0 & 86.0 & \textbf{97.6} & Llama-3.2-1B-Instruct \\
& temporal\_sequences & 70.0 & 90.0 & \textbf{100.0} & calme-2.3-llama3.1-70b \\
\bottomrule
\end{tabular}%
}
\caption{Big bench hard (BBH) leaderboard - human average and top performance vs top open source LLM models.}
\label{tab:big-bench-hard}
\end{table}

\subsubsection{CQA tasks decomposition and primitives}
Answering by decomposition could be grouped in:
\begin{itemize}
    \item[--] \textbf{Standard sub-tasks} include intent detection, word sense disambiguation, entity recognition (NER) and linking, topic classification, sentiment classification, information extraction, fact retrieval, ranking, and summarization, including query focused summarization...
    \item[--] \textbf{Advanced sub-tasks} include multi-hop \& decomposition, domain oriented tasks, sources \& fact checking, code generation \& program synthesis, causal explanation (or possible consequences), temporal explanation...
    \item[--] \textbf{External sub-tasks} leverage resources out of model like program synthesis~\citep{droriNeuralNetworkSolves2022}, using a solver, or invoking tools (function calling) and retrieval variants (e.g., self-critique/corrective RAG).
\end{itemize}

\citet{wuAIChainsTransparent2022} propose a taxonomy of \textbf{primitive tasks in decomposed and chained LLM} which could be applicable to CQA:
\begin{description}
    \item[--] \textbf{Validate and categorize input} such as \textbf{classification} which assigns the input to categories.
Most useful for branching logic and validation (e.g. is the question answerable?).
    \item[--] \textbf{Gather additional information} from the LLM such as \textbf{Factual Query} to ask LLM for a fact, {Generation} to ask LLM to do some creative “hallucination” on the input, \textbf{Ideation} to ask a list of ideas or examples.
    \item[--] \textbf{Re-organize input} such as \textbf{Information extraction} from the context, \textbf{Rewriting (1-1 mapping)} input to more machine-readable formats (e.g. JSON to natural language) or other usage (e.g. translation), \textbf{Split Points (1-N mapping)} for splitting contexts, digging concepts, {Compose Points (N-1 mapping} to  synthesise, reverse operation of decomposition like merge multiple results back together.
\end{description}

\subsubsection{CQA tasks hybrid program decomposition examples}
To better illustrate how CQA tasks could be decomposed between a LLM and an external software module to better solve complex problems, we invite you to see how: university level math problems (CQA) can be solved by splitting the task between a LLM and a Python language interpreter~\citep{droriNeuralNetworkSolves2022}, physical reasoning question can be solved by splitting the task between a LLM and a physics engine~\citep{liuMindsEyeGrounded2022}.
Those solutions design can be extrapolated to many complex QA challenges as proposed in \autoref{sec_hybridLLMpatterns}.
Each of these patterns can be combined to split sub-tasks of complex problem to leverage most adapted module for the task by ensuring necessary context is provided wherever needed.

\subsection{LLM limits}\label{sec_LLMlimits}
Scaling up language models has been shown to predictably improve performance\citep{weiEmergentAbilitiesLarge2022} on a wide range of downstream tasks.
The HELM~\citep{liangetal.HolisticEvaluationLanguage2022} and BIG~\citep{bigetal.ImitationGameQuantifying2022} studies show that state-of-the-art on most scenarios are led by those very larges models but still lack on different sides (e.g. fairness, robustness across tasks \& domains, reasoning).
Many additional components are used or investigated to face limitations of those models by augmentation or hybridization with LLM.
For example, ChatGPT and InstructGPT~\citep{baiTrainingHelpfulHarmless2022} added reinforcement learning with human feedback to its pre-trained large language model to highly improve their answer performance (e.g. calibration, human expectation alignment).
Therefore we decided to cover in this study the improvement of the language model itself (see next section) and the hybridation patterns which can overcome different limits of base models (below).
Those different limits or challenges have been identified in our systematic review, and we linked them to different potential solutions presented in next section.
Some emerging challenges like resistance to adversarial attacks~\citep{bartoloImprovingQuestionAnswering2021} are not covered.

\label{Tablelimits} 
{
\begingroup
\definecolor{limIntent}{HTML}{E8F4FF}
\definecolor{limDecomp}{HTML}{FFF4E5}
\definecolor{limAlign}{HTML}{F3E8FF}
\definecolor{limFact}{HTML}{EAF7EB}
\definecolor{limReason}{HTML}{E8FFFB}
\definecolor{limLong}{HTML}{FFFBEA}
\definecolor{limMulti}{HTML}{FFEAF3}
\definecolor{limTemp}{HTML}{FFF0F0}
\definecolor{limData}{HTML}{EDEFFB}
\definecolor{limExp}{HTML}{F0F7FF}
\definecolor{limAdapt}{HTML}{E9FBF7}
\definecolor{limBias}{HTML}{F9EEF3}
\definecolor{limEff}{HTML}{F4F6F8}

\definecolor{solTrain}{HTML}{1F77B4}   
\definecolor{solHybrid}{HTML}{2CA02C}  
\definecolor{solPrompt}{HTML}{FF7F0E}  
\definecolor{solAgent}{HTML}{9467BD}   

\newcommand{\solTrainMark}{\textcolor{solTrain}{\rule{0.8ex}{0.8ex}}}
\newcommand{\solHybridMark}{\textcolor{solHybrid}{\rule{0.8ex}{0.8ex}}}
\newcommand{\solPromptMark}{\textcolor{solPrompt}{\rule{0.8ex}{0.8ex}}}
\newcommand{\solAgentMark}{\textcolor{solAgent}{\rule{0.8ex}{0.8ex}}}

\rowcolors{3}{white}{white}

\begin{longtable}{p{0.6\linewidth} >{\tiny}p{0.35\linewidth}}
\\

\multicolumn{2}{r}{\tiny
\solTrainMark\ Training\quad
\solHybridMark\ Hybridization\quad
\solPromptMark\ Prompting\quad
\solAgentMark\ Agentic \& reasoning-time
}\\[-0.1\baselineskip]
Limitation & Potential solutions (see also \autoref{sec_limits_and_research}) \\\midrule
\endhead

\rowcolor{limIntent}
\textbf{Question/intent understanding \& conversational state} — enhancing context, intent, goals identification, through question expansion, clarification, dialog.
&
    \solHybridMark\ Hybridization: \ref{HP13}, \ref{HP3}, \ref{HP12};
    \solPromptMark\ Prompting: \ref{sec_prompting};
    \solTrainMark\ Post-training alignment: \ref{sec_posttraining_alignment};
    \solTrainMark\ Mid-training (retrieval/tool-aware, long-context): \ref{sec_midtraining};
    \solAgentMark\ Agentic \& reasoning-time: \ref{sec_agentic_and_reasoning}, \ref{sec_reasoning_time}.
\\

\rowcolor{limDecomp}
\textbf{Question decomposition \& planning} — breaking down complex questions into simpler sub-questions, enabling multi-hop/step reasoning and action design.
&
    \solPromptMark\ Prompting: \ref{sec_prompting};
    \solHybridMark\ Hybridization: \ref{HP4}, \ref{HP10}, \ref{HP15}, \ref{HP20};
    \solAgentMark\ Agentic \& reasoning-time: \ref{sec_agentic_and_reasoning}, \ref{sec_reasoning_time};
    \solTrainMark\ Mid-training: \ref{sec_midtraining}.
\\

\rowcolor{limAlign}
\textbf{Alignment to human expectations/preferences}~\citep{baiTrainingHelpfulHarmless2022} — ensuring models align with human expectations and values, cultural differences.
&
    \solTrainMark\ Post-training alignment (RLHF / DPO / ORPO / KTO): \ref{sec_posttraining_alignment};
    Reinforcement/experience: \ref{sec_ImprovementLoop_and_kg_capitalization};
    \solHybridMark\ Hybridization: \ref{HP9}, \ref{HP17};
    \solAgentMark\ Agentic workflows: \ref{sec_agentic_and_reasoning};
    \solTrainMark\ Training: \ref{sec_supervisedtraining_active}, \ref{sec_supervisedtraining_instructions}.
\\

\rowcolor{limFact}
\textbf{Factuality, grounding \& attribution (hallucinations/faithfulness)}~\citep{jiSurveyHallucinationNL2023} — reducing hallucination, ensuring answer accuracy, and providing evidence/attribution.
&
    \solHybridMark\ Hybridization: \ref{HP5}, \ref{HP11}, \ref{HP15}, \ref{HP20};
    \solTrainMark\ Post-training alignment \& PRMs: \ref{sec_posttraining_alignment};
    \solAgentMark\ Agentic \& reasoning-time (verify/rerank, self-consistency/cascades): \ref{sec_agentic_and_reasoning}, \ref{sec_reasoning_time};
    \solPromptMark\ Prompting: \ref{sec_prompting};
    \solTrainMark\ Training: \ref{sec_supervisedtraining_multiview}, \ref{sec_improvetraining}.
\\

\rowcolor{limReason}
\textbf{Reasoning} — LLM progressed a lot on reasoning with Chain-of-Thought and reasoning at inference-time strategies but they are nor optimal in time and quality, nor universal on domains. incorporating higher logical reasoning, causality, and learning from code to improve problem-solving capabilities.
&
    \solAgentMark\ Agentic \& reasoning-time: \ref{sec_agentic_and_reasoning}, \ref{sec_reasoning_time};
    \solHybridMark\ Hybridization: \ref{HP3}, \ref{HP7}, \ref{HP8}, \ref{HP14}, \ref{HP15}, \ref{HP20};
    \solTrainMark\ Post-training (RFT/PRMs): \ref{sec_posttraining_alignment};
    \solTrainMark\ Training: \ref{sec_supervisedtraining_multitask}, \ref{sec_pretraining_program}.
\\

\rowcolor{limLong}
\textbf{Handling long-context questions \& long-form answers} — handling long inputs, long-range dependencies, and multi-document summarization/synthesis.
Most LLM have limitation in output size (or drift to goal) and input (question but also all necessary to knowledge to add), as well as in the reasoning length dependencies.
&
    \solHybridMark\ Hybridization: \ref{HP4}, \ref{HP5}, \ref{HP6}, \ref{HP10}, \ref{HP14};
    \solTrainMark\ Mid-training (long-context adaptation): \ref{sec_midtraining};
    \solAgentMark\ Agentic \& reasoning-time (cascades/verification): \ref{sec_agentic_and_reasoning}, \ref{sec_reasoning_time}.
\\

\rowcolor{limMulti}
\textbf{Multi-modal understanding, search and reasoning} - knowledge and world model cannot always be captured with text only and require other forms of information.
&
    \solHybridMark\ Hybridization: \ref{HP16}, \ref{HP7}, \ref{HP8};
    \solTrainMark\ Training: \ref{sec_supervisedtraining_meta}, \ref{sec_pretraining_program}, \ref{sec_supervisedtraining_multitask};
    \solAgentMark\ Agentic orchestration: \ref{sec_agentic}.
\\

\rowcolor{limTemp}
\textbf{Temporal reasoning \& freshness} - handling time-based reasoning, knowledge update, and understanding sequences or workflows.
&
    \solHybridMark\ Hybridization: \ref{HP19}, \ref{HP5}, \ref{HP14};
    \solTrainMark\ Mid-training / updating: \ref{sec_midtraining}, \ref{sec_supervisedtraining_transfer};
    \solAgentMark\ Agentic workflows: \ref{sec_agentic}.
\\

\rowcolor{limData}
\textbf{Data sensitivity protection} - utilizing \& protecting sensitive data, such as private, intellectual property, organizational, or governmental sensitive data.
&
    \solHybridMark\ Hybridization/RAG with access control \& verification: \ref{HP5}, \ref{HP6}, \ref{HP11};
    Deployment patterns: \ref{sec_data_sensitivity};
    \solTrainMark\ Mid-training (offsite/federated): \ref{sec_midtraining};
    \solAgentMark\ Agentic guardrails/gating: \ref{sec_agentic_and_reasoning}.
\\

\rowcolor{limExp}
\textbf{Experience / knowledge and skills capitalization} - like a human, it should be able to continually improve itself by experience on skills and knowledge using implicit and explicit feedback.
GPT-4 technical report highlighted it still does not have experience learning skill~\citep{openaiGPT4TechnicalReport2023}
&
    Reinforcement/experience: \ref{sec_ImprovementLoop_and_kg_capitalization};
    \solHybridMark\ Hybridization: \ref{HP3}, \ref{HP9}, \ref{HP12}, \ref{HP13}, \ref{HP14}, \ref{HP18}.
    \solAgentMark\ Agentic \& reasoning-time: \ref{sec_agentic_and_reasoning}, \ref{sec_reasoning_time}.
deliberate \emph{reasoning-time} scaling (\ref{sec_reasoning_time}) and process-level supervision (PRMs) improve step quality without updating base weights; persistent “experience” still requires explicit memory (\ref{HP14}) or retraining.
\\

\rowcolor{limAdapt}
\textbf{Adaptating and updating} - adapting models to specific domains and tasks, and ensuring they stay updated with new knowledge
&
    \solHybridMark\ Hybridization: \ref{HP2}, \ref{HP3}, \ref{HP5}, \ref{HP6}, \ref{HP18};
    \solTrainMark\ Training: \ref{sec_pretraining_transfert}, \ref{sec_supervisedtraining_transfer}, \ref{sec_supervisedtraining_instructions}, \ref{sec_PETtraining}, \ref{sec_midtraining};
    Reinforcement/experience: \ref{sec_ImprovementLoop_and_kg_capitalization};
    \solPromptMark\ Prompting: \ref{sec_prompting}.
\\

\rowcolor{limBias}
\textbf{Bias}~\citep{liangetal.HolisticEvaluationLanguage2022} - mitigating representational and statistical biases across topics, domains, and contexts
&
    \solTrainMark\ Post-training alignment: \ref{sec_posttraining_alignment}.
    \solHybridMark\ Hybridization: \ref{HP9}, \ref{HP18}.
    Reinforcement/experience: \ref{sec_ImprovementLoop_and_kg_capitalization}.
\\

\rowcolor{limEff}
\textbf{Efficiency \& scaling (cost/latency/energy)}~\citep{liangetal.HolisticEvaluationLanguage2022} — training/inference efficiency and cost control.
&
    \solHybridMark\ Hybridization: \ref{HP5}, \ref{HP12}, \ref{HP17};
    \solTrainMark\ Training: \ref{sec_PETtraining}, \ref{sec_improvetraining}, \ref{sec_midtraining};
    \solAgentMark\ Agentic \& reasoning-time (cascades/adaptive compute): \ref{sec_agentic_and_reasoning}, \ref{sec_reasoning_time}.
\\
\end{longtable}
\endgroup
}

\section{Evaluating: metrics, cost functions, datasets}\label{sec_evaluation}
The performance of language models on question answering can vary greatly depending on factors such as the domain, question complexity, necessary subtasks, bias, fairness, toxicity, and human expectations.
A large language model may perform well overall but struggle with some type of questions, or areas of evaluation,
while a model that is specialized for certain questions or domains may perform poorly on more general tasks.

To orient readers across the most relevant living and frozen evaluations used in complex QA, \autoref{tab:cqa-eval-stack} consolidates widely used suites, what they measure, and their headline metrics.

\begin{table}[htbp]
\footnotesize
\setlength{\tabcolsep}{4pt}
\renewcommand{\arraystretch}{1.15}
\caption{Main living and frozen leaderboards covering the different aspects of LLM complex QA evaluation.}
\begin{tabularx}{\linewidth}{@{} p{3cm} p{2.5cm} p{4.5cm} p{3cm} @{}}
\toprule
\textbf{Aspect} & \textbf{Name (link)} & \textbf{Task description} & \textbf{Metric} \\
\midrule
Capability anchor (reasoning+knowledge)
& \href{https://crfm.stanford.edu/helm/capabilities/latest/}{HELM Capabilities}
& Standardized LLM capabilities evaluations (e.g., MMLU-Pro, GPQA, IFEval, WildBench, Omni-MATH).
& Aggregate, rescaled mean over scenario-specific metrics \\
\addlinespace
\multirow{2}{*}{Academic level QA}
& \href{https://scale.com/leaderboard/humanitys_last_exam}{HLE Humanity's Last Exam}
& 2,500 closed-ended, expert-created questions across >100 subjects; 14\% multimodal (text+image); private held-out split for anti-overfitting
& Accuracy, RMS calibration error \\

& \href{https://artificialanalysis.ai/evaluations/gpqa-diamond}{GPQA-Diamond}
& Expert-vetted, graduate-level science questions.
& Accuracy (\% correct) \\
\addlinespace
Safety anchor
& \href{https://crfm.stanford.edu/helm/safety/latest/}{HELM Safety}
& Multi-risks evaluation (violence, fraud, discrimination, sexual content, harassment, deception).
& Category-specific scores and overall aggregates \\
\addlinespace
Broad academic knowledge
& \href{https://huggingface.co/spaces/TIGER-Lab/MMLU-Pro}{MMLU-Pro}
& Harder, contamination-resistant revision of MMLU across many subjects.
& Accuracy (\% correct) \\
\addlinespace
Math reasoning (annual)
& \href{https://artificialanalysis.ai/evaluations/aime-2025}{AIME-2025}
& 30 AIME 2025 problems; strict final-answer grading.
& Accuracy (\% solved) \\
\addlinespace
Long-context QA/comprehension
& \href{https://longbench2.github.io/}{LongBench v2}
& QA, summarization, dialogue history, code and structured-data understanding with very long contexts.
& Task-specific (mostly accuracy); with/without CoT \\
\addlinespace
Knowledge-intensive QA / RAG
& \href{https://eval.ai/web/challenges/challenge-page/689/leaderboard/}{KILT}
& Unified open-domain QA, fact-checking, entity linking, slot-filling with Wikipedia provenance.
& KILT-EM/F1 gated by retrieval; retrieval R-Precision/Recall@k \\
\addlinespace
& \href{https://github.com/centerforaisafety/CRAG}{CRAG (Comprehensive RAG)}
& Benchmark suite for retrieval-grounded QA that emphasizes corrective retrieval and explicit attribution/citation; evaluates end-to-end RAG quality across domains.
& Composite: retrieval precision/recall, groundedness/attribution, and task-specific answer correctness. \citep{yangCRAGComprehensiveRAG2024} \\
\addlinespace
Long-form factuality \& grounding
& \href{https://www.kaggle.com/benchmarks/google/facts-grounding}{FACTS Grounding}
& Judge-based evaluation of whether long-form answers are fully grounded in a provided document.
& Aggregate factuality/grounding score \\
\addlinespace
Software reasoning / coding
& \href{https://www.swebench.com/}{SWE-bench (Verified / Lite / Bash-Only)}
& End-to-end bug fixing on real OSS repositories; multiple tracks.
& \% Resolved (issues solved) \\
\addlinespace
Trustworthiness (fairness/toxicity/robustness/privacy)
& \href{https://trustllmbenchmark.github.io/TrustLLM-Website/leaderboard.html}{TrustLLM}
& Integrated suite across truthfulness, safety, fairness, robustness, privacy, machine ethics.
& Dimension-specific metrics and aggregates \\
\addlinespace
Human preference (helpfulness)
& \href{https://lmarena.ai/}{LMSYS Chatbot Arena}
& Crowdsourced pairwise battles of model outputs (text \& multimodal).
& Elo-style Arena score (pairwise win rate) \\
\addlinespace
Abstract reasoning “limits”
& \href{https://arcprize.org/leaderboard}{ARC Prize (ARC-AGI-2)}
& Abstraction \& reasoning puzzles; easy for humans, hard for current AI.
& Accuracy on private test (score and cost) \\
\bottomrule
\end{tabularx}
\label{tab:cqa-eval-stack}
\end{table}

The following section will examine different metrics and datasets used for evaluating and training these models.
\subsection{Metrics \& SOTA performance}\label{sec_metricsSOTA}

\subsubsection{Standard metrics}\label{sec_standard_metrics}
There are a variety of metrics that can be used to evaluate the performance of QA models, each with its own strengths and weaknesses.
In this section, we will discuss some of the most common metrics used to evaluate QA models:
\begin{description}
\item [Recall@k] Measures the proportion of relevant answers retrieved by the model among the first k answers.
It is a measure of the model's ability to find all relevant answers, regardless of their position in the ranking. The main weaknesses are that it does not take into account the position of the relevant answer in the ranking and does not penalize irrelevant answers that appear in the top k.
\item [Accuracy@k] Measures the proportion of correct answers among the first k answers returned by the model.
It is a measure of the model's ability to return the correct answer and can be used to evaluate a model's performance in a closed-domain QA task.
As with recall@k, the main weaknesses are that it does not take into account the position of the correct answer in the ranking and does not penalize irrelevant answers that appear in the top k.
\item [nDCG] Normalized Discounted Cumulative Gain, a measure of rank quality that takes into account the relevance and position of responses.
It is often used in information retrieval and web search to measure the effectiveness of a ranking algorithm.
The main weaknesses are that it does not take into account the number of irrelevant answers in the ranking.
\item [MAP] Mean Average Precision, a measure of the quality of a set of ranked responses.
This is a commonly used metric for evaluating QA models in open domain tasks, where the model should return a list of possible answers.
its main weakness is that it does not take into account the position of the correct answer in the ranking.
\item [MRR] Mean Reciprocal Rank, a measure of the quality of a set of ranked answers, with a higher value indicating better performance.
It is often used in information retrieval and web search to measure the effectiveness of a ranking algorithm.
Weakness: It only considers the position of the first correct answer in the ranking.
\item [CWS] Cross-entropy word-level perplexity, a measure of the model's ability to predict the next word in a sentence.
It is often used to evaluate the quality of the language model.
Weakness: it cannot be used for answer quality and perplexity is not well correlated with a language model tasks performance.
\item [F1 (macro, micro)] F1 score, a measure of a model's accuracy that takes into account both precision and recall.
It is used to evaluate the model's performance in a classification task.
F1 macro averages the per-class F1 scores, used for imbalanced datasets, F1 micro computes metrics globally by counting total true positives, false negatives, and false positives, used for balanced datasets.
Weakness: It does not take into account the relative importance of false positives and false negatives.
\item [EM] Exact Match, a binary metric that measures whether the model's answer is exactly the same as the reference answer.
It is often used in closed-domain QA tasks where the correct answer is a single word or phrase.
Weakness: It is not able to capture the semantic similarity between the model's answer and the reference answer.
\item [Jaccard similarity] measures similarity between two sets of data, based on the size of the intersection divided by the size of the union of the sets.
Weakness: It does not take into account the overall size of the sets, which can lead to errors in similarity measurements.
\item [Cosine similarity] measures similarity between two non-zero vectors based on the cosine of the angle between them.
Weakness: It is sensitive to the magnitude of the vectors (e.g. zero), which can lead to errors in similarity measurements.
\item [BLEU] Bilingual Evaluation Understudy, a measure of the similarity between a model's answer and a reference answer.
It is commonly used in machine translation and natural language generation tasks.
Weakness: It does not take into account the meaning of the words (semantic similarity).
\item [ROUGE] Recall-Oriented Understudy for Gisting Evaluation, a measure of the similarity between a model's answer and a reference answer.
It is commonly used in natural language summarization tasks.
Additionally to BLEU, it takes into account the longest common sequence (LCS).
Weakness: it does not take into account the meaning of the words (semantic similarity).
\item [METEOR] Metric for Evaluation of Translation with Explicit ORdering, a measure of the similarity between a model's answer and a reference answer.
It is similar to BLEU, but takes into account word alignment and synonymy.
Weakness: It is computationally expensive and can be sensitive to the reference translations used.
\item [LLM-as-judge]~\citep{zhengJudgingLLMasaJudgeMTBench2023} Uses a LLM to evaluate the quality, helpfulness, or factuality of another model's output, often by scoring or pairwise comparison. Weakness: Can exhibit position bias, self-consistency bias (favoring its own style), and requires careful prompting and \emph{calibration} against human preferences to be reliable.
\item [Human evaluation] human judges provides a subjective measure of the quality of the model's answers.
It is considered the gold standard for evaluating QA models.
Weakness: it is time-consuming and can vary greatly depending on the individual evaluators and their level of expertise.
\end{description}

\subsubsection{Metrics for free-form/natural language QA}
In the context of free-form QA, standard metrics limits question complexity~\citep{chenEvaluatingQuestionAnswering2019} and do not capture many good answers semantically close.
So new metrics have been proposed using PLM having higher correlations to human expectations:
\begin{description}
\item [BERTScore]~\citep{zhangBERTScoreEvaluatingText2020} compute a semantic similarity score through a sum of cosine similarities between the contextualized embeddings of answer tokens and those of the reference text.
\item [BARTScore]~\citep{yuanBARTScoreEvaluatingGenerated2021} similar to BERTScore, it uses the more recent model BART pre-trained using a more robust technique (denoising autoencoding).
A BARTScore variant adds faithfulness in the measure.
\item [MAUVE]~\citep{pillutlaMAUVEMeasuringGap2021}: similar to BERTScore and add divergence frontiers.
It claims to better correlate with human judgement and identify quality differences.
\item [T5Score] this hybrid metric~\citep{qinT5ScoreDiscriminativeFinetuning2022} based on mT5 model is not yet compared to date with MAUVE but is globally more robust than BERTScore and BARTScore.
\end{description}

However \citet{heBlindSpotsModelBased2022} highlight that all those PLM based metrics have flaws, they could assign a high likelihood to degenerate, repetitive text and could be insensitive to perturbations such as word order shuffling, negation, etc.
We can balance those blind spots by composing with complementary metrics.

\subsubsection{Multi-metrics and scenarios}
\citet{liangetal.HolisticEvaluationLanguage2022} referenced a large space of targeted use cases of LLMs with a focus on QA, and identifies 7 key categories of metrics required to create useful systems: accuracy, calibration, robustness, fairness, bias, toxicity, and efficiency (speed/cost). For accurate definition of theses categories of metrics please refer to HELM taxonomy.
\citet{liangetal.HolisticEvaluationLanguage2022} highlights that performance of models are unequal and even best performer is not the best choice depending on metrics preference regarding your needs.
The relationship between accuracy and calibration depends on the scenario (task \& domain) and adaptation procedure, with some scenario showing trade-offs between accuracy and calibration.
Across all scenarios, there is a strong correlation between accuracy, robustness, and fairness, but trade-offs where the most accurate model is not the most robust or fair.
The study also found performance disparities in models when demographic metadata is available, and low overall average biases and toxicity in model generations, but notes that targeted evaluations are needed to obtain a more detailed characterization.
The study also found that there is not a strong trade-off between accuracy and efficiency, but as models become larger, accuracy improves but with higher training and inference cost.
Only a subset of all models are on the accuracy-efficiency Pareto frontier for each scenario.
There is no model leading on all metrics, QA performances also vary depending on the scenario (task \& domain) and model, so weighting or defining a decision tree among the 7 metrics then evaluating on target scenario is necessary for choosing your model.

\citet{liangetal.HolisticEvaluationLanguage2022} introduced a holistic, multi-metric view (accuracy, calibration, robustness, fairness, bias, toxicity, efficiency) that remains relevant for complex QA. However, \textbf{static 2022 leaderboards are no longer representative} of the frontier landscape in 2024–2025. Current practice complements HELM with \emph{harder} reasoning tests and \emph{living} leaderboards—e.g., \textbf{MMLU‑Pro}, \textbf{GPQA}, \textbf{Arena‑Hard}/Chatbot Arena Elo, \textbf{Humanity's Last Exam}, and \textbf{LiveBench}—to avoid saturation and contamination and to better probe complex QA.
Accordingly, any summary figures based solely on 2022 snapshots (e.g., “InstructGPT best on many axes”) should be replaced by panels that track these evolving sources; the core HELM finding (“no single winner across metrics and scenarios”) still holds.

In practice, we orient readers to the evaluation landscape summarized in Table~\ref{tab:cqa-eval-stack}, which consolidates widely used living and frozen leaderboards, what they measure, and headline metrics.
  
\subsubsection{Which metrics for "complex" QA?}

Considering that complex QA is not well defined and answers highly depends on human expectations and values, we could not identify standard metrics in the literature.
\citet{ullrichUsingBloomsTaxonomy2021} propose to use Bloom's taxonomy to assess question complexity but does not offer a metric for measuring the relevance of answers to question.
Metrics identified in previous section for free-form QA could be used when there is a gold/reference answer but shall be unstable considering that a good non-factoid answer can be semantically distance to gold answer.
In this survey we aim to focus on complex questions with the following characteristics: non-factoid, multi-step (requiring decomposition), multi-source of knowledge, higher order of reasoning questions.
We could separately measure each skill of a language model on those characteristics to estimate the capacity to answer using decomposition.
Using data from BIG bench, we created a summary evaluation of similar QA capacities comparing human performance to open source LLM models (see \ref{tab:big-bench-hard}).
However, it will not assess the end-to-end capacity to provide a relevant final answer.
Current systems like ChatGPT which solves some complex questions, with high differences in quality but a clear improvement curve, mainly used human feedback for alignment and evaluation~\citep{openaiGPT4TechnicalReport2023}.
A path investigated in papers WebGPT~\citep{nakanoWebGPTBrowserassistedQuestionanswering2022} and Constitutional AI~\citep{baiConstitutionalAIHarmlessness2022} is to build Elo‑style preference maps (e.g., helpfulness vs harmlessness, compute efficiency) and seek Pareto frontiers.

\subsubsection{Explainability, truthfulness, hallucination metrics}
Papers reviewed often highlight the recurrent problem of hallucination~\citep{jiSurveyHallucinationNL2023} and urge for explainability~\citep{leiterExplainableEvaluationMetrics2022, wiegreffeTeachMeExplain2021} and truthfulness~\citep{linTruthfulQAMeasuringHow2022} when delivering an answer.

Conventional metrics measuring quality of writing including answers are not adequate for quantifying the level of hallucination~\citep{jiSurveyHallucinationNL2023}.
Except human evaluation, there are no mature metrics:
\begin{itemize}
    \item[--] statistical metrics mainly focus on lexical matching~\citep{jiSurveyHallucinationNL2023} such as PARENT-T, bag-of-vectors sentence similarity (BVSS)~\citep{martindaleIdentifyingFluentlyInadequate2019}, Knowledge F1.
    \item[--] model based metrics expect to handle more complex syntactic and semantic variations: they can compare LLM performance on known gold answer in different scenarios (e.g. QuestEval~\citep{scialomQuestEvalSummarizationAsks2021}) or on free parameters using LLM-as-judge approach.
When using LLM-as-judge to scale evaluation, it is necessary to add a \emph{human based calibration} with human reviews to mitigate bias and ensure alignment.
    \item[--] \emph{faithfulness/attribution} metrics for retrieval-grounded QA are increasingly used in practice (e.g., \textbf{FActScore} for atomic factual precision; \textbf{RAGAS} for groundedness and context/answer relevance) to complement human judgments; these are now de facto in many RAG evaluation stacks.
    \item[--] human evaluation is the most commonly used method considering the currently imperfect automatic methods, most common usages are: (1) scoring: annotator rate the evaluation level out of a range; (2) comparing: annotator compares the output texts with baselines or ground-truth references.
\end{itemize}

We identified three types of explanation in the literature~\citep{wiegreffeTeachMeExplain2021}: highlights, free-text, and structured explanations.
Those explanations could be intrinsic, explaining LM internal logic, or extrinsic, related to external sources and proofs.
Therefore some metrics for the extrinsic explanation could be the number of sources, authority and reliability of sources.
The intrinsic explanation could be measured trough their quality: compactness (short and coherent), sufficiency, comprehensiveness.
Some indirect metrics could be related to the explaination task performance (source identification, fact-checking, coherence...).
Explaination are usually evaluated on plausibility and faithfulness (coherent decision process), a common approach is to provide a chain of facts that detail the reasoning steps to reach an answer.

For further details, we invite you to look into the taxonomy in \autoref{Table Explainable Evaluation Metrics}~\citep{leiterExplainableEvaluationMetrics2022}, main references~\citep{wiegreffeTeachMeExplain2021, leiterExplainableEvaluationMetrics2022, jiSurveyHallucinationNL2023, linTruthfulQAMeasuringHow2022}, and research topics in the further section "Hallucination \& credibility".

\setcounter{table}{1}

\begin{table}\centering
\caption{Taxonomy of existing explainable evaluation metrics (extracted from table of \citet{leiterExplainableEvaluationMetrics2022})}\label{Table Explainable Evaluation Metrics}
\begin{tabular}{llll}
\toprule
Work & Type & Method & Goals \\ \midrule
Eval4NLP 2021: (Fomicheva et al. 2021) & Various \\
Rubino, Fujita, and Marie (2021) & FI & Expl. by Design & AL \\
Treviso et al. (2021) & FI & Various & AL \\
SemEval 15/16: (Agirre et al. 2015, 2016) & Various \\
Magnolini, Feltracco, and Magnini (2016) & CAl & Neural Networks & AL, E \\
Yuan, Neubig, and Liu (2021) & CA & Generation Prob. & E \\
Adversarial Attacks (Section 7) & EbE & Perturbations & D, E \\
Kaster, Zhao, and Eger (2021) & EbS/CA & Linear Regression & D, E \\
Sai et al. (2021) & CA & Perturbations & B, D, E \\
\end{tabular}
\caption*{
The first column is the research work reference.
The second is the explanation types: Concept Attribution (CA), Chunk Alignment (CAl), Feature Importance (FI), Explanation by Example (EbE) and Explanation by Simplification (EbS).
The column “Goals” specifies which aspect is measured amongst (B)ias detection, (D)iagnosis, (E)xpressiveness and automated labeling(AL).}
\end{table}

\subsubsection{Domain/task matrix of performance}
As performance of a model is unequal depending on knowledge domain and tasks~\citep{liangetal.HolisticEvaluationLanguage2022, bigetal.ImitationGameQuantifying2022}, metrics assessment should be segmented per knowledge domain \& key tasks within a matrix of comparison or database like in the 42 scenarios of HELM~\citep{liangetal.HolisticEvaluationLanguage2022}. 

\subsection{Cost functions}
Cross-entropy (negative log-likelihood) is the primary objective used to train and fine-tune language models. This objective is adapted to each training phase (see sections \ref{sec_pretraining}, \ref{sec_supervisedtraining}, \ref{sec_PETtraining}, \ref{sec_improvetraining}) and admits richer variants for knowledge distillation~\citep{wuOneTeacherEnough2021}.
We will see in later sections that it is frequent to add a reinforcement learning mechanism to align QA with humans’ expectations.
Importantly, this RLHF/RFT stage is \emph{not} a simple additive term of “CE $+$ reward.” After supervised fine-tuning, the policy is optimized to \emph{maximize} an estimated reward while staying close to a reference policy via a KL penalty; e.g.,
\[
\max_{\theta}\; \mathbb{E}_{x,y\sim \pi_{\theta}}[\,r_{\phi}(x,y)\,] \;-\; \beta\,\mathrm{KL}\!\left(\pi_{\theta}(\cdot\!\mid\!x)\,\|\,\pi_{\text{ref}}(\cdot\!\mid\!x)\right),
\]
as in PPO-based RLHF \citep{ouyangTrainingLanguageModels2022}. Cross-entropy is used in earlier stages (pre-training/SFT) and may be mixed with PPO gradients (e.g., PPO-ptx), but the RL objective itself is distinct from CE.

Early large language models were optimized with the negative log-likelihood (cross-entropy) loss, which remains the foundation for pre-training and supervised fine-tuning of QA systems~\citep{wuOneTeacherEnough2021}.  
Since 2023, however, the design of cost functions has diversified to address alignment, reasoning, and hybrid retrieval-based architectures.  
The current landscape can be grouped into four main families:
\begin{enumerate}
  \item \textbf{Likelihood-based objectives.}  Token-level cross-entropy or its masked variants are still used during pre-training, mid-training, and task-specific fine-tuning (\S\ref{sec_pretraining}–\S\ref{sec_PETtraining}).  They model next-token prediction or reconstruction and provide the base likelihood signal for all later objectives.
  \item \textbf{Preference-based objectives.}  Reinforcement learning from human feedback (RLHF)~\citep{ouyangTrainingLanguageModels2022} has evolved toward lighter, reinforcement-free formulations that directly optimize preference data.  Notable examples include \emph{Direct Preference Optimization (DPO)}~\citep{rafailovDirectPreferenceOptimization2023}, \emph{Odds-Ratio Preference Optimization (ORPO)}~\citep{hongORPOMonolithicPreference2024}, and \emph{Kahneman–Tversky Optimization (KTO)}~\citep{ethayarajhModelAlignmentProspect2024}.  These objectives are widely adopted in 2024–2025 due to their simplicity and lower cost, especially in open-source pipelines; however, multiple studies report that carefully tuned PPO-based RLHF can outperform DPO on several benchmarks, so the choice is task- and resource-dependent~\citep{xuDPOSuperiorPPO2024,ivisonUnpackingDPOPPO2025}.
  \item \textbf{Reinforcement and reasoning-oriented objectives.}  A recent line of work termed \emph{Reinforcement Fine-Tuning (RFT)} optimizes verifiable rewards that target reasoning quality (e.g., correctness and format) rather than only final answers.  The \emph{Group Relative Policy Optimization (GRPO)} algorithm used in DeepSeek-R1~\citep{guoDeepSeekR1IncentivizesReasoning2025} exemplifies this trend; GRPO removes the critic and estimates baselines from groups of samples, and reports strong gains on mathematical and scientific QA.
  \item \textbf{Process-level and hybrid objectives.}  Step-wise supervision with \emph{process reward models (PRMs)}~\citep{lightmanLetsVerifyStep2023} and verifier-guided learning introduces auxiliary losses on intermediate reasoning steps, improving faithfulness and robustness.  In retrieval-augmented and agentic settings, models are also trained with \emph{RAG-aware losses}, e.g., \emph{Self-RAG}~\citep{asaiSelfRAGLearningRetrieve2023}, which adds control tokens and reflection losses for retrieval and self-critique decisions.
\end{enumerate}
In contemporary systems, these objectives are most often optimized \emph{sequentially} across stages (pre-training $\rightarrow$ SFT $\rightarrow$ preference tuning $\rightarrow$ RL/RFT) rather than as a single simultaneous loss~\citep{ouyangTrainingLanguageModels2022,zieglerFineTuningLanguageModels2020}. When objectives are optimized \emph{jointly} (e.g., monolithic preference alignment or multi-task RAG/verifier training), a composite objective can be written as:
\[
  \mathcal{L}_{\text{total}}
  = \mathcal{L}_{\text{CE}}
  + \lambda_1\,\mathcal{L}_{\text{pref}}
  + \lambda_2\,\mathcal{L}_{\text{RL/RFT}}
  + \lambda_3\,\mathcal{L}_{\text{proc/RAG}} ,
\]

where the weights balance next-token likelihood, preference alignment, reasoning reinforcement, and retrieval/verification feedback. A concrete joint example is ORPO, which appends a log-odds–based preference penalty to the standard NLL during SFT in a single stage~\citep{hongORPOMonolithicPreference2024}.

By contrast, PPO-based RLHF typically optimizes an expected-reward objective with a KL penalty to a reference policy rather than an additive CE term:
\[
\max_{\theta}\; \mathbb{E}_{x,y\sim \pi_{\theta}}[\,r_{\phi}(x,y)\,] \;-\; \beta\,\mathrm{KL}\!\left(\pi_{\theta}(\cdot\!\mid\!x)\,\|\,\pi_{\text{ref}}(\cdot\!\mid\!x)\right),
\]
with optional interleaving of SFT-style updates (e.g., PPO-ptx)~\citep{ouyangTrainingLanguageModels2022,zieglerFineTuningLanguageModels2020}.  

This evolution of objectives parallels the move from single-model language modeling toward hybrid, agentic, and verifier-coupled architectures capable of tackling complex question answering.

\subsection{Datasets}
To train and evaluate QA/CQA systems, a variety of datasets have been developed to cover main skills, tasks and knowledge which we reference in the following sections.
They can assess current performance or train model from end-to-end question answering on different field and complexity, to specific logic or task, such as decomposition.
We also cover the generation of datasets or improvement of existing ones to overcome quality issues or challenge to create domain or logic specific dataset.
In addition to legacy QA sets, community practice now relies on harder multi‑task reasoning (e.g. \textbf{MMLU‑Pro}, \textbf{GPQA} (graduate‑level, “Google‑proof” science)), plus \emph{living} leaderboards such as Arena‑Hard and LiveBench for open‑ended QA.
For long‑context/multidoc and multimodal QA, newer suites (e.g., LongBench variants, MMMU/MathVista) are increasingly used alongside earlier multi‑hop sets.
Long-form free answers of QA datasets above a page (> 250 words) was rare and are difficult to evaluate, but increasingly needed. Table~\autoref{Table_LongformQA_datasets} compares representative long-form QA/summarization datasets to help select resources by answer length, attribution requirements, and evaluation metrics.

\setcounter{table}{1}
\begin{table}\centering
\caption{Comparison of representative long-form answer datasets used for QA-style evaluation.}\label{Table_LongformQA_datasets}
\resizebox{\textwidth}{!}{%
\begin{tabular}{p{2cm} p{2cm} p{3cm} p{3cm} p{2.2cm} p{3cm} p{2cm} p{1.5cm}}
\toprule
\textbf{Dataset name} & \textbf{Type} & \textbf{Task} & \textbf{Format} & \textbf{Answer length} & \textbf{Evaluation metric} & \textbf{Made by} & \textbf{\# Instances} \\
\midrule
\href{https://facebookresearch.github.io/ELI5/}{ELI5} & Open-domain LFQA & Explanatory answers from web evidence & Free-form paragraphs; evidence docs & $\approx$130 words (answers) & ROUGE-1/2/L; human fluency/correctness/preference & FAIR & 270k \\
\href{https://arxiv.org/abs/2309.07852}{EXPERTQA} & Expert LFQA & Long answers with citations & Attributed answers + claim judgments & $\approx$160 words & ROUGE; QA-based factuality (QAFactEval); claim-level FActScore & UPenn, UW & 2.2k \\
\href{http://participants-area.bioasq.org/datasets/}{BioASQ (Task B, ideal)} & Biomedical expert QA & Ideal summary + exact answers & Multi-sentence \emph{ideal} + evidence & \emph{Cap} $\leq$200 words (ideal) & ROUGE-2/ROUGE-SU4 + manual (ideal); EM/F1 (exact) & BioASQ consortium & $\sim$5.4k (train) \\
\href{https://gov-report-data.github.io/}{GovReport} & Long-doc \textbf{summ.} (gov) & Summarize govt.\ reports & Report $\to$ multi-paragraph abstractive summary & $\approx$553 words (summary) & ROUGE-1/2/L; QA-based APES\textsubscript{src}; human & UMich \& UIUC & 19.5k docs \\
\href{https://aclanthology.org/2022.acl-long.589/}{SummScreen} & Dialogue \textbf{summ.} (TV) & Summaries of TV episode transcripts & Transcripts $\to$ human recaps & $\sim$114 / 381 \emph{tokens} (FD/TMS recaps) & ROUGE; entity-centric (character/pair overlap); human & TTIC \& Duke Univ. & $\sim$27k episodes \\
\href{https://arxiv.org/abs/2205.11465}{SQuALITY} & Question-focused long-doc \textbf{summ.} & Summaries that answer questions & Story + question $\to$ 4 ref.\ summaries & $237$ \emph{tokens} (resp.) / $442$ \emph{tokens} (plot) & Human ratings (correctness/coverage/overall); ROUGE/METEOR/BERTScore (low corr.) & NYU (MLL) & 500 Q / 2k refs \\
\href{https://arxiv.org/abs/2010.12694}{AQuaMuSe} & Query-based \textbf{MDS} & qMDS across web docs & Query + $\sim$6 docs $\to$ summary & $\approx$106 words (summary) & ROUGE-1/2/L & Google Research & 5.5k \\
\href{https://arxiv.org/abs/2105.08209}{BookSum (chapter)} & Long-narrative \textbf{summ.} & Chapter-level narrative summarization & Chapters $\to$ abstractive summaries & $\sim$505 \emph{tokens} (summary) & ROUGE-1/2/L; BERTScore; human Likert & Salesforce Research & 12.6k chapters \\
\href{https://arxiv.org/abs/2105.08209}{BookSum (book)} & Long-narrative \textbf{summ.} & Book-level narrative summarization & Books $\to$ multi-paragraph summaries & $\sim$1{,}167 \emph{tokens} (summary) & ROUGE-1/2/L; BERTScore; human Likert & Salesforce Research & 405 books \\
\bottomrule
\end{tabular}%
}
\end{table}

\subsubsection{Legacy QA/CQA text datasets (monomodal)}
By “legacy monomodal” we refer to text–only QA corpora that shaped modern information retrieval (IR), machine reading comprehension (MRC), and open-domain QA prior to retrieval-grounding and multimodality. These resources remain canonical testbeds for span extraction, short-answer and multiple-choice QA, and dialogue QA, and are still widely used for pretraining, ablations, and backwards-compatible comparisons.
\begin{description}
    \item [MS Marco] (Microsoft Machine Reading Comprehension Dataset): largest publicly available collection of relevance judgments, with 100,000 to 1,000,000 human generated QA, it has been central to the progress in neural IR/QA over the past several years (standard QA task, human performance: Rouge-L: 0.539, BLEU-1: 0.485)~\citep{liangetal.HolisticEvaluationLanguage2022}.
    \item [SQUAD] (Stanford  Question  Answering Dataset): 100,000+ questions posed by crowdworkers on a set of Wikipedia articles (human performance F1-score: 86.8\%).
    \item [SQuAD v2]: add 50,000 unanswerable questions written adversarially by crowdworkers to look similar to answerable ones, to avoid training on unreliable guesses on questions (human performance F1-score: 86.8\%).
    \item [TriviaQA]: 650K question-answer-evidence triples.
TriviaQA includes 95K question-answer pairs authored by trivia enthusiasts and independently gathered evidence documents, six per question on average, with distant supervision for answering. In comparison to other QA datasets in 2017, TriviaQA (1) had relatively complex, compositional questions, (2) considerable syntactic and lexical variability between questions and corresponding answer-evidence sentences, and (3) required more cross sentence reasoning to find answers.
    \item [MMLU] (Measuring Massive Multitask Language Understanding): 15,908 multiple-choice questions packages across 57 different tasks/datasets from subjects in the humanities, social sciences, hard sciences, and many others (unspecialized human performance accuracy: 34.5\%).
    \item [NarrativeQA]: 46,765 human generated questions \& answers requiring understanding of stories (books, movie scripts) requiring summarization (human performance: Bleu-1:44.3, Bleu-4:18.9, Meteor:24.0, Rouge-L:57.1)
    \item [NaturalQuestions] (closed-book, open-book): 323,000 questions and documents (full dataset is 42Gb) consisting of real anonymized, aggregated queries from  Google search engine providing several long documents (e.g. 5 wikipedia pages) with a long answer (e.g. a paragraph) and a short answer (one  or  more  entities) if present on the pages, or marks null if no long/short answer is present (standard human performance: short answers F1: 57.5\%, long answer F1: 73.4\%).
    \item [QuAC]: >100,000 questions and their corresponding answers, based on dialogues between two persons where many questions requires understanding of the dialog (human performance F1: 80.9\%).
    \item [Semi-structured datasets]: semi-structured data with \textbf{tables-and-text} are abundant on the web and in companies.
\citet{wangSurveyTableandTextHybridQA2022} list the following datasets: HybridQA, OTT-QA, GeoTSQA, FinQA, TAT-QA, TAT-HQA, MultiHiertt.
Semi-structured dataset could also be samples of JSON, XML....
For \textbf{Structured, graph, table only or SQL like data} (not bundled with text), \citet{rogersQADatasetExplosion2022} list the following datasets: WikiTableQuestions, TableQA, WikiSQL, WikiOps.
    \item [CQA on knowledge bases datasets]: knowledge bases like ontologies and knowledge graphs offers valuables structured symbolic data (e.g. Wikidata, all resources from lod-cloud.net...) not always easy to query to non experts.
\citet{lanComplexKnowledgeBase2022} list the following datasets: WebQuestions, ComplexQuestions, WebQuestionsSP, ComplexWebQuestions, QALD series, LC-QuAD, LC-QuAD 2.0, MetaQA Vanilla, CFQ, GrailQA, KQA Pro.
    \item [Domain specific datasets] are numerous and can be specific to sectors and perimeters (e.g. Covid19 in medical sector, Qasper QA about research papers in NLP~\citep{dasigiDatasetInformationSeekingQuestions2021}).
\end{description}

\subsubsection{Hard tasks datasets}.
\label{sec_Hard_tasks_datasets}
We use "hard tasks" to denote \emph{stress tested tasks} that probe different types of reasoning, multi-step deduction, long-range dependencies, and robustness under minimal spurious cues~\citep{liangetal.HolisticEvaluationLanguage2022}. They complement legacy monomodal QA sets by focusing on capabilities that typically limit end-to-end complex QA systems. In practice, frozen test sets are paired with \emph{living} leaderboards that resist contamination and track progress over time; see the evaluation stack in \autoref{tab:cqa-eval-stack}.

\textbf{BBH and BBEH}: \emph{BIG-bench}~\citep{bigetal.ImitationGameQuantifying2022} collects 200+ tasks across linguistics, math, science, and commonsense; its "Hard" selection (BBH) aggregates tasks that remained difficult for earlier LLMs and serves as a standard reasoning stress test.
\emph{BIG-bench Extra Hard (BBEH)}~\citep{kazemiBIGBenchExtraHard2025} further isolates subsets that remain challenging for frontier models, and is increasingly reported alongside BBH to measure reasoning progress.
The BIG-bench family mixes algorithmic, logical, linguistic and temporal competencies that are directly implicated in complex-QA decomposition and verification, and interacts well with inference-time strategies (e.g., self-consistency) and verifier-based selection discussed elsewhere in this survey. For concrete ceilings, \autoref{tab:big-bench-hard} summarizes representative BBH tasks (human averages/experts vs.\ top open models); Appendix \autoref{tab:bbeh} contrasts multiple recent models side-by-side on selected BBEH tasks.

\textbf{Very-hard academic QA (closed-ended).} To complement BBH/BBEH with \emph{exam-style} difficulty, recent suites target graduate and upper-undergrad knowledge with contamination-resistant procedures and calibration reporting. Notably, \emph{Humanity’s Last Exam (HLE)}~\citep{phanHumanitysLastExam2025} provides \(\sim\)2,500 expert-authored, closed-ended questions across \(>100\) subjects (with a modest multimodal share), and reports both \emph{accuracy} and \emph{calibration} (e.g., RMS calibration error), using a private held-out split to discourage overfitting. This positions HLE as a high-signal “very hard” anchor for complex QA beyond BBH/BBEH. See \autoref{tab:cqa-eval-stack} for its placement among living leaderboards and metrics.

\subsubsection{Multimodal QA datasets}
When answering a question, humans build a coherent understanding of the world by actively exploring and  reasoning over multiple evidences (multi-hop) from different modalities (multimodal) like illustrated by \citet{yangEnhancingMultimodalMultihop2022}.
Therefore, natural QA requires to leverage more than text (natural or structured) like images, videos, sensors...
Training \& evaluation datasets are emerging like:

\begin{description}
    \item [Image QA datasets]: \citet{zakariVQAVisualReasoning2022} provide a list of images/visual question-answering (VQA) including reasoning tasks.
    \item [Audio QA datasets]: DAQA~\citep{fayekTemporalReasoningAudio2019} on audio temporal reasoning, Clotho-AQA~\citep{lippingClothoAQACrowdsourcedDataset2022} on binary and multi-choice audio QA.
    \item [Video QA datasets]: such as VideoQA~\citep{zhongVideoQuestionAnswering2022} for multi-domain, MovieQA~\citep{tapaswiMovieQAUnderstandingStories2016}/MovieFIB~\citep{maharajDatasetExplorationModels2017}/TVQA~\citep{leiTVQALocalizedCompositional2019}/KnowIT VQA~\citep{garciaKnowITVQAAnswering2019} for movies and shows, MarioQA~\citep{munMarioQAAnsweringQuestions2017} for games, PororoQA~\citep{kimDeepStoryVideoStory2017} for cartoons, TurorialVQA~\citep{colasTutorialVQAQuestionAnswering2020} for tutorials, CLEVRER~\citep{maoCLEVRERHumansDescribingPhysical2022} for physical \& causal reasoning.
    \item [Multi-modal QA datasets]: MultiModalQA/MMQA~\citep{talmorMultiModalQAComplexQuestion2021} for multi-modal and multi-hop QA, WebQA~\citep{changWebQAMultimodalMultihop2022} on web multi-modal QA, MAQA focus on negation learning and testing~\citep{liMAQAMultimodalQA2023}.
    \item [Unified dataset format] is proposed by \citet{xieUnifiedSKGUnifyingMultiTasking2022} to unify multiple formats of different modality to enable training, inference and evaluation on multi-tasks and sources.
\end{description}

\subsubsection{Structured knowledge datasets}.
As seen in first section "QA/CQA text datasets (monomodal)", we can use structured (e.g. RDBMS, graph, table) or semi-structured (e.g. table and text, JSON samples) datasets to learn to extract factual information from structured knowledge sources in natural language.
These datasets are also key to improve general or domain specific reasoning abilities as we see in further section \ref{reasoning_dataset}.

\begin{table}
\newcolumntype{T}{>{\tiny}c} 
\captionsetup{skip=1pt} 
\caption{Multi-hop QA datasets (base: \citet{maviSurveyMultihopQuestion2022}, \textbf{updated to 2025}) - OD: Open domain context, MCQ A: multi-choice question with A being the number of possible answers}
\label{TableMultiHopDatasets}
\scriptsize
\begin{tabular}{T T T T T T T T T}
\toprule

\makecell{Context \\ granularity} & Dataset & \makecell{Total context \\ of dataset} & Context source & Domain & \makecell{Number of \\ questions} & \makecell{Context \\ per question} & \makecell{Average \\ \# hops } & Answer type \\ \midrule
Passage & HotpotQA & Wikipedia & Wikipedia & Generic & 112,779 & 10/0 ODa & 1/2/3 & Span \\ 
\makecell{Table,\\ Passage} & HybridQA & \makecell{Tables: 13k \\ Passages: 293k} & \makecell{Wikitables, \\ Wikipedia} & Generic & 69611 & \makecell{1 table \\ passages} & 2/3 & Span \\ 
Sentence & NarrativeQA & \makecell{Books: 783 \\ Movies: 789} & Multiple & Fiction & 46765 & 1 story & - & Generative \\
Sentence & MultiRC & 871 & Multiple & Generic & 9872 & 1 passage & 2.37 & MCQ A : 5.44 \\ 
Passage & Medhop & Medline & Medline & Medicine & 2508 & OD & - & MCQ A : 8.9 \\ 
Passage & Wikihop & |Wikipedia| & Wikipedia & Generic & 51318 & OD & - & MCQ A1:19.8 \\ 
Sentence & QASC & \makecell{Core: 928 \\ Other: 7672} & WorldTree & Science & 9980 & OD & 2 & MCQ A:8 \\
Sentence & OpenBookQA & \makecell{Core: 1326 \\ Other: 6000} & WorldTree & Science & 5947 & OD & 2 & MCQ A:4 \\ 
Passage & 2WikiMultiHopQA & \makecell{Wikipedia + \\ Wikidata} & \makecell{Wikipedia, \\ Wikidata} & Generic & 192{,}606 & \makecell{10 paragraphs \\ (distractor)} & 2/3/4 & \makecell{Span / Yes–No; \\ evidence triples} \\
Passage & MuSiQue-Ans & Wikipedia & Wikipedia & Generic & 24{,}814 & \makecell{20 paragraphs} & 2/3/4 & Span \\
\makecell{Multi-doc} & FanOutQA & English Wikipedia & Wikipedia & Generic & 1{,}034 & \makecell{$\ge$5 docs \\ (avg $\sim$7)} & $\approx$7 & \makecell{Short generative \\ (list/number)} \\
Passage & MoreHopQA & \makecell{Derived from \\ HotpotQA, 2Wiki, MuSiQue} & Wikipedia & Generic & 1{,}118 & varies & \makecell{2–4 + extra step} & Generative \\
\makecell{Multi-doc} & MultiHop-RAG & \makecell{English news \\ article corpus} & News & News & 2{,}556 & \makecell{2–4 evidence \\ pieces} & 2–4 & \makecell{Short answer} \\
Passage & MEQA & \makecell{WikiEvents-based \\ documents} & Wikipedia/IE & \makecell{Event-centric} & 2{,}243 & \makecell{intra-/inter-doc} & multi-step & \makecell{Entities (lists) \\ + explanations} \\
\bottomrule
\end{tabular}
\end{table}

\subsubsection{Decomposition and multi-hop datasets}.
Decomposition skill is required for CQA to break down complexity, and the related ability to resolve it in multiple hops or steps.
Table~\autoref{TableMultiHopDatasets} summarizes widely used multi-hop QA datasets~\citep{maviSurveyMultihopQuestion2022} with hops, context granularity, and answer types; use it to select datasets aligned with your target reasoning depth and supervision format.
\citet{bigetal.ImitationGameQuantifying2022} also provide advanced decomposition and multi-step tasks datasets in \ref{sec_Hard_tasks_datasets} such as strategyQA or multistep arithmetics.
In order to improve problem specific decomposition and resolution ability, emerging datasets are providing reasoning decomposition examples to be provided in context like chain-of-thoughts (e.g. FLAN CoT dataset\citep{chungScalingInstructionFinetunedLanguage2022}).
They are mainly used as examples to be provided with the question but could be also used at training. In a different manner, Meta's Galactica model was trained on scientific papers where step-by-step reasoning were wrapped between 2 tokens <WORK>\citep{taylorGalacticaLargeLanguage2022} both to explicitly learn reasoning and activate working memory which lacks in standard LLM.

\subsubsection{Instructions (IFT, CoT) datasets}.
Instructions fine-tuning (IFT) are collection of written instructions used to teach model user intent declaration to solution logic \& format which can be model generated such as: Unnatural Instructions:~\citep{honovichUnnaturalInstructionsTuning2022}, large community effort Super-natural instructions~\citep{wangSuperNaturalInstructionsGeneralizationDeclarative2022}, small high-quality crafted~\citep{wangSelfInstructAligningLanguage2022}, converted existing large datasets to instructions~\citep{iyerOPTIMLScalingLanguage2023}, NaturalInstructions~\citep{mishraCrossTaskGeneralizationNL2022}.

\subsubsection{Reasoning datasets}\label{reasoning_dataset}.
Dedicated datasets for specific reasoning abilities~\citep{qiaoReasoningLanguageModel2022} have been developed, or existing sets could be derived to take advantage of different abilities.
\begin{enumerate}
    \item general combination of reasoning such as \href{github.com/facebookarchive/bAbI-tasks}{bAbI} tasks categorized by type \& complexity;
    \item spatial or temporal reasoning: spatial question (e.g. SpartQA), event order (e.g. QuAIL, TORQUE),
    event attribution to time (e.g. TEQUILA, TempQuestions), event duration (e.g. MCTACO, QuAIL),
    temporal commonsense knowledge (e.g. MCTACO, TIMEDIAL),
    questions where the correct answers change with time (e.g. ArchivalQA, SituatedQA), temporal reasoning in multimodal setting (e.g. DAGA, TGIF-QA);
    additionally, recent synthetic, contamination-resistant temporal logic suites (e.g., \emph{Test of Time}) systematically probe LLM temporal reasoning outside world knowledge memorization~\citep{fatemiTestTimeBenchmark2024}.
    \item belief states (e.g. Event2Mind, QuAIL);
    \item causal relations (e.g. ROPES, QuAIL, QuaRTz, ESTER);
    large-scale, structured causal benchmarks such as \emph{COLD} (closed daily-activity graphs)~\citep{joshiCOLDCausalReasOning2024} and long-chain visual causal QA like \emph{CausalChaos!}~\citep{parmarCausalChaosDatasetComprehensive2024} stress multi-step causal inference with explanations.
    \item other relations between events such as subevents, conditionals, and counterfactuals (e.g. ESTER);
    event-centric, multi-hop question answering with explicit explanation chains (e.g., \emph{MEQA}) targets reasoning over event–entity and event–event relations~\citep{liMEQABenchmarkMultihop2024}.
    \item entity properties and relations : social interactions (e.g. SocialIQa), properties of characters (e.g. QuAIL), physical properties (e.g. PIQA, QuaRel), numerical properties (e.g. NumerSense);
    complex aggregative reasoning that composes extracted facts into tabular calculations is captured by \emph{TACT} (text\,$\rightarrow$\,tables with arithmetic/aggregation)~\citep{caciularuTACTAdvancingComplex2024}.
    \item tracking entities: across locations (e.g. bAbI), in coreference chains (e.g. Quoref, resources in the Winograd Schema Challenge family).
    \item multi-document / multi-hop reasoning datasets designed to discourage extractive shortcuts and test genuine composition:
    \emph{MoreHopQA} extends Hotpot/2Wiki/MuSiQue with generative, higher-difficulty multi-hop questions~\citep{schnitzlerMoreHopQAMoreMultihop2024};
    \emph{MultiHop-RAG} benchmarks retrieval \emph{and} reasoning over multi-evidence chains for RAG systems~\citep{tangMultiHopRAGBenchmarkingRetrievalAugmented2024}.
\end{enumerate}

\subsubsection{Explainable \& truthfulness QA datasets}.
The veracity and explainability of an answer is a significant challenge for language models where answers are mostly provided without evidence, logic, confidence/trust.
Explainability can be trained by models or evaluated through different tasks like source highlighting or URL providing and importance, claim check, commonsense check, answer explanation, logic check...\newline

\citet{leiterExplainableEvaluationMetrics2022} propose three types of ground truth explanations: highlights (rationales or feature importance), free-text explanations, and structured explanations.
In \autoref{TableExplDatasets} we enriched an existing comparison~\citep{wiegreffeTeachMeExplain2021} listing important \textbf{datasets with explainability tasks} in field with number of instances, mode of creation, explanation type and format, task.

\citet{rogersQADatasetExplosion2022} propose an "evidence format" for the explainable part of a dataset composed of Modality (Unstructured text, Semi-structured text, Structured knowledge, Images, Audio, Video, Other combinations) and Amount of evidence (Single source, Multiple sources, Partial source, No sources).

\begin{table}
\newcolumntype{T}{>{\tiny}c} 
\captionsetup{skip=1pt} 
\caption{Explainability tasks datasets (data from \citet{wiegreffeTeachMeExplain2021} enriched)}
\label{TableExplDatasets}
\scriptsize
\begin{tabular}{T T T T T T}
\toprule
Ex type & Dataset & Task & Ex Format & Made by & \# Instances \\ \midrule
\multirow{13}{*}{highlight} & TRUTHFULQA & check false belief \& ref in QA & URL ref & experts & 817 \\ 
& TriviaQA with evidence or filtered & QA with evidence & URL ref & crowd \& experts & 95k / 2099 \\ 
& FEVER  or filtered (GopherCite) & verifying claims from text & sentences or URLs & crowd & ~136K \\ 
& WIKIQA & open-domain QA & sentence & crowd + authors & 1,473 \\ 
& MULTIRC & reading comprehension QA & sentences & crowd & 5,825 \\ 
& HOTPOTQA & reading comprehension QA & sentences & crowd & 112,779 \\ 
& Hanselowski et al. & verifying claims from text & sentences & crowd & 6,422 (varies) \\ 
& CoQA & conversational QA & none & crowd & ~127K (1 or 3) \\ 
& COS-E v1.0 [100] & commonsense QA & none & crowd & 8,56 \\ 
& COS-E v1.11 & commonsense QA & none & crowd & 10,962 \\ 
& BOOLQ & reading comprehension QA & none & crowd & 199 \\ 
& SCIFACT & verifying claims from text & 1-3 sentences & experts & 995 (1-3) \\ 
& Kutlu et al. & webpage relevance ranking & 2-3 sentences & crowd & 700 (15) \\ 
\hline
\multirow{11}{*}{free-text} & NaturalQuestions & QA with evidence & sentences & crowd \& experts & 323k / 7638 \\
& or filtered version (GopherCite) \\
& ELI5 or filtered (GopherCite) & long QA & sentences & crowd \& experts & 270k / 3999 \\ 
& Jansen et al. & science exam QA &  & authors & 363 \\ 
& Ling et al.  & solving algebraic word problems &  & auto + crowd & ~101K \\ 
& LIAR-PLUS & verifying claims from text &  & auto & 12,836 \\ 
& COS-E v1.0 [100] & commonsense QA &  & crowd & 8,56 \\ 
& COS-E v1.11 & commonsense QA &  & crowd & 10,962 \\ 
& ECQA & commonsense QA &  & crowd & 10,962 \\ 
& PUBHEALTH & verifying claims from text &  & auto & 11,832 \\ 
& ESPRIT & reasoning about qualitative physics &  & crowd & 2441 (2) \\ 
\hline
\multirow{11}{*}{structured} & ProofWriter & Reasoning QA with proof & rules, QA, chain of facts &  & 500k \\ 
& WORLDTREE V1 & science exam QA & explanation graphs & authors & 1,68 \\ 
& OPENBOOKQA & open-book science QA & 1 fact from WORLDTREE & crowd & 5,957 \\ 
& WORLDTREE V2 & science exam QA & explanation graphs & experts & 5,1 \\ 
& QED & reading comp. QA & inference rules & authors & 8,991 \\ 
& QASC & science exam QA & 2-fact chain & authors + crowd & 9,98 \\ 
& EQASC & science exam QA & 2-fact chain & auto + crowd & 9,980 (~10) \\ 
& Ye et al. & SQUAD QA & semi-structured text & crowd + authors & 164 \\ 
& Ye et al. & NATURALQUESTIONS QA & semi-structured text & crowd + authors & 109 \\ 
& R4C & reading comp. QA & chains of facts & crowd & 4,588 (3) \\ 
& STRATEGYQA & implicit reasoning QA & reasoning steps w/ highlights & crowd & 2,780 (3) \\  

\bottomrule
\end{tabular}
\end{table}

\subsubsection{Local \& multi-lingual datasets}.
Multilingual datasets can not only help to assess or expand the answering skills to new languages, but also to transfer or share concepts between languages: CCQA focuses on general QA pre-training, MLQA on extractive question answering, MKQA on open domain QA, TydiQA is a topologically diverse QA dataset to learn more robust concept and answer without the need of translation, XQuAD demonstrates performance on pre-training mono-lingual and fine-tuning on new languages, MGSM deals with solving math problems and transfering reasoning abilities across 10 languages.

\subsubsection{Dialogue datasets}.
The "Dialogue System Technology Challenge" started as an initiative to provide a common testbed on dialog state tracking and is now a reference in terms of dialog dataset~\citep{zhangAutomaticEvaluationModeration2021} with each year several tacks (challenge).
The most recent 10 tracks were released for \href{https://dstc10.dstc.community/tracks}{DSTC10} and \href{https://dstc11.dstc.community/tracks}{DSTC11 in 2022}.
Some other reference datasets are: CommaQA on complex tasks solving by talking to agents, SODA with million exchange with social commonsense contextualization, DeliData for multi-party problem solving with deliberation, TIMEDIAL for temporal commonsense reasoning.

\subsubsection{Synthetic or improved datasets by generation}.
Creating datasets is very expensive but often necessary for domain adaptation.
A growing trend is the \textbf{generation of synthetic QA datasets} from models~\citep{jeronymoInParsv2LargeLanguage2023} or unstructured text using different techniques.
\citet{yuCOCODRCombatingDistribution2022} compares ICT, GPL, GenQ, Promptagator and COCO-DR.
Dataset generation can also be used to robustify reasoning skills learning. \citet{trivediTeachingBroadReasoning2022} convert an existing QA dataset provided with question decomposition meaning representation (QDMR) to generate new contexts and questions.
Some other techniques like natural language augmentation~\citep{dholeNLAugmenterFrameworkTaskSensitive2022} aims at enriching existing datasets for a more robust training through transformation and data filtering. An interesting paper from \citet{yuanReStructuredPretraining2022} highlights the "signals" present in datasets for learning knowledge and capabilities and propose to restructure formats of pre-training datasets to improve learning for different of signals (e.g. skills or knowledge type).

\textbf{New practice:} for retrieval‑grounded complex QA, synthetic data pipelines increasingly include \emph{retrieval/critique labels} (e.g., Self‑RAG control tokens and reflection signals) so that generated corpora train \emph{when to retrieve, how much to retrieve, and how to self‑evaluate}. In parallel, \emph{programmatic optimizers} (e.g., DSPy) are used to automatically compile prompts, demonstrations, and retrieval policies that yield higher‑quality QA data and more robust evaluation harnesses~\citep{asaiSelfRAGLearningRetrieve2023, khattabDemonstrateSearchPredictComposingRetrieval2023}.

\section{Solving with training}\label{sec_training}
Now that we surveyed the skills we need to develop, the tasks and challenges to solve, the datasets needed for training: how to train for complex QA ? We will see the importance of pre-training, domain adaptation and fine-tuning.

\subsection{Training dataset quality}
Even before pre-training, a very important step is to maximize the quality of the input datasets (eg, accuracy, completeness, consistency, relevance, uniformity).
It is commonly said that machine learning project could spend up to 80\% of time on data preparation with a major part dedicated to data cleaning~\citep{pfistererHumanCenteredAutoML2019}.
The efficiency of a model task is directly and heavily affected by the quality of the training dataset and its improvement~\citep{budachEffectsDataQuality2022}.
We will not dig this subject because, surprisingly, our review methodology did not preselect any scientific article with the word "quality" in their title and is anecdotal in their abstracts.
It might be that it is not specific to QA/CQA or language models training but an assumption in any machine learning related subject.
Developers of GPT-3 spent important efforts to filter on a high-quality pre-training dataset~\citep{zongSurveyGPT32022}, and \citet{sunImportanceBuildingHighquality2022} highlight this important preparation task in similar search tasks.

\subsection{Type of LLMs training}
As introduced in the "core concepts" section, training can apply to:
\begin{enumerate}
    \item a \textbf{pretrained language model (PLM)} which is trained without supervision (unsupervised or self-supervised) mainly on large text (e.g. reddit, Wikipedia...) to discover general knowledge and logic.
It could then be re-used, optionally augmented with a "head", to be further trained (mid or post-training) on specific tasks (supervised training), on preferences (alignment), on new reasoning capabilities.
It can also be trained on very large corpus of data to uncover enough knowledge and logic to be used "as is" without additional training but oftenly some additional instructions to better align with requester expectations (e.g. AI chatbot services).
\citet{kalyanAMMUSSurveyTransformerbased2021} identify several type of pre-training: Pretraining from Scratch (PTS), Continual Pretraining (CPT - initialization from an existing PLM), Simultaneous Pretraining (SPT - synchronized mix of general and domain-specific corpus from beginning), Task Adaptive Pretraining (TAPT - continually adapts mix of general and specific training examples), Knowledge Inherited Pretraining (KIPT - adds knowledge distillation in the process).
    \item \textbf{fine-tuned model} on specific task(s) in a supervised manner (each training example is provided with input and expected output solution), by re-purposing a pre-trained language model.
    \item adapting an existing model (PLM or fine-tuned model) to a new domain of knowledge (\textbf{domain adaptation} - e.g. COVID19 terms and facts) or to new task(s) (\textbf{knowledge transfer}) leveraging existing knowledge and logic in the model.
    \item a distinct \textbf{mid‑training} stage (continued/skill‑targeted training) that sits between generic pre‑training and post‑training alignment: (i) \emph{CPT/DAPT/TAPT} to expand domain or task coverage with largely unlabeled corpora; (ii) \emph{retrieval‑/tool‑aware continued training} (e.g. learning \emph{when} to retrieve, \emph{how much} to retrieve, and \emph{how} to critique) to support hybrid pipelines; (iii) \emph{long‑context adaptation} to sustain multi‑document reasoning; and (iv) \emph{privacy‑/federation‑constrained adaptation} (e.g. offsite/federated recipes) for sensitive deployments (see \autoref{sec_midtraining}).
\end{enumerate}

\subsection{Pre-training techniques}\label{sec_pretraining}

\subsubsection{Self-supervised learning (SSL)}
This type of machine learning technique, largely used for PLM, trains on a dataset without any pre-labeled outputs.
Instead, the model must learn to generate the correct output itself, using information from the input data.
It is often based on an unlabeled training converted to a supervised coherence task.
\citet{kalyanAMMUSSurveyTransformerbased2021} identify three major techniques:
\begin{description}
    \item [Generative SSL], depending on chosen technique, the model learns to predict different scenarios: (1) next token based on current tokens (CLM (causal language model) is used by GPT-1, GPT-2, GPT-3 models); (2) masked tokens in a phrase or MLM (masked language model is the most used technique, but variants exists such as TLM (translation language model) or Seq2SeqLM used by RoBERTa, XLM, XLM-R models); (3) reconstruct original text which has been corrupted (denoising autoencoder, DAE, is used by BART, mBART models).
    \item [Contrastive SSL] augments learning by comparison.
It is not used alone but to further improve a model like in continual pretraining, to learn sentence-level semantics.
Different techniques exist such as next-sentence prediction NSP~\citep{devlinBERTPretrainingDeep2019}, sentence order prediction (SOP), simple contrastive learning like SimCSE~\citep{gaoSimCSESimpleContrastive2022}, bootstrapping with BYOL~\citep{grillBootstrapYourOwn2020}, cross lingual contrastive pretraining with XLCo~\citep{chiInfoXLMInformationTheoreticFramework2021}.
    \item [Adversarial SSL] learns by distinguishing corrupted tokens (replaced or shuffled), can be used alone or in continual pretraining like contrastive.
Different techniques exist: replaced token detection (RTD - used by ELECTRA model), multi-lingual replaced token detection (MRTD) is used by XLM-E model, translation replaced token detection (TRTD), shuffled token detection (STD) is used by RoBERTa model.
    \item [Hybrid SSL] uses more than one type of SSL - e.g. U-Palm uses up to 7 denoising objectives as \textbf{mixtures-of-denoisers}~\citep{tayUL2UnifyingLanguage2022}, BERT uses MLM (generative) and NSP (contrastive), ALBERT used MLM and SOP (contrastive), infoXLM uses TLM (generative) and XLCo (contrastive).
\end{description}

\subsubsection{Transfer learning, domain adaptation, knowledge distillation}\label{sec_pretraining_transfert}
Those techniques are also used as supervised learning \& finetuning, we cover them in the next section.

\subsubsection{Program execution learning}\label{sec_pretraining_program}
This technique\citep{piReasoningProgramExecutors2022} learn to mimic inputs to ouputs logics of program to capture new logic or more general skills like numerical reasoning, logical reasoning, better multi-hop reasoning. This technique useful at pre-training stage can be viewed as a self-supervised learning.

\subsection{Mid-training: continued pre-training for retrieval, long-context and deployment constraints}\label{sec_midtraining}
We call \emph{mid‑training} the family of \emph{continued, largely unlabeled} training procedures applied \emph{after} generic pre‑training but \emph{before} (or alongside) post‑training alignment. The goal is to endow a pre-trained LLM (PLM) with capabilities that complex QA systems rely on at inference time—\emph{retrieval policy, critique/verification hooks, long‑context robustness, and privacy‑constrained adaptation}—without requiring fully supervised task labels.

\paragraph{(A) CPT/DAPT/TAPT as a stage.} Beyond the definitions given earlier, we group \emph{continual pre‑training} and its domain/task‑adaptive variants as a distinct stage to expand knowledge and styles relevant to downstream QA without overfitting small datasets. This reduces the burden on post‑training alignment for domain transfer and improves robustness in knowledge‑intensive pipelines.

\paragraph{(B) Retrieval‑/tool‑aware continued training.} Mid‑training can teach models to \emph{decide when to retrieve, how much to retrieve, and how to critique} generated answers. \emph{Self‑RAG} trains special control tokens for retrieval decisions and self‑evaluation and shows reliable factuality/grounding gains on QA~\citep{asaiSelfRAGLearningRetrieve2023}. \citet{salemiEvaluatingRetrievalQuality2024}'s methodology further clarifies how to \emph{evaluate retrieval quality inside RAG} systems. CRAG benchmark~\citep{yangCRAGComprehensiveRAG2024} operationalizes corrective retrieval and attribution requirements that modern mid‑training aims to satisfy.

\paragraph{(C) Long‑context adaptation.}~\citep{chenLongLoRAEfficientFinetuning2023, moMidTrainingLargeLanguage2025} For multi‑document reasoning and book‑length contexts, mid‑training includes position‑handling and \emph{train‑short, test‑long} strategies (e.g., ALiBi/positional interpolation) combined with continued training on long inputs so that later SFT/RLHF does not shoulder length generalization alone. This ties directly to long‑context QA evaluations highlighted in \autoref{sec_evaluation}.

\paragraph{(D) Privacy‑/federation‑constrained adaptation.} In sensitive settings, mid‑training adopts \emph{offsite} or federated recipes so adaptation occurs where the data lives. \citet{hongDPOPTMakeLarge2024} introduced \emph{DP‑OPT}, which privately learns prompts/adapters client‑side and transfers them to cloud models, offering a practical bridge between pre‑training and post‑training in privacy‑constrained QA deployments.

\paragraph{When to use mid‑training vs.\ SFT/PEFT?} Use mid‑training when you need \emph{behavioral priors} (retrieval/tool use, long‑context coping, privacy constraints) that generalize across tasks and reduce label needs. Use SFT/PEFT rather for \emph{formatting and instruction following} or tight training latency constraints. In practice, many strong QA stacks follow: \emph{pre‑training} $\rightarrow$ \emph{mid‑training for retrieval/long‑context/constraints} $\rightarrow$ \emph{SFT} $\rightarrow$ \emph{preference optimization (DPO/ORPO/KTO) or RLHF}, with verifiers/PRMs applied at inference.

\subsection{Post-training: Supervised fine-tuning}\label{sec_supervisedtraining}
Supervised learning is the ancestor and most well-known ML technique.
It trains on labeled dataset to predict the expected ouput from given input.
This allows the model to learn from the data and make predictions about new, unseen data but similar task.
This assumes that dataset is representative of new, unseen data.
We will see in sections \ref{sec_PETtraining} and \ref{sec_prompting} that task specific fine-tuning, which can require a lot of compute and examples, can be avoided via complementary strategies like prompt engineering, tuning adapters, soft prompts prefix, late prompts.

\subsubsection{(Task specific) Vanilla Fine-Tuning} is commonly used to refer to the standard method of fine-tuning only the few layers near the output of the pre-trained model ("the head") with a task-specific loss.
\citet{kalyanAMMUSSurveyTransformerbased2021} highlight that the main drawback is that PLM having large parameters are prone to overfit on small task specific datasets.
Intermediate fine-tuning or multi-task fine-tuning overcome this.

\subsubsection{Multi-task learning}\label{sec_supervisedtraining_multitask}
trains a model to perform {multiple different tasks} to help the model learn~\citep{kalyanAMMUSSurveyTransformerbased2021} more generalizable features (regularization effect), and improve its performance on multiple tasks (transverse knowledge and skills acquired from multiple datasets).
\citet{raffelExploringLimitsTransfer2020} demonstrated how one model can reach state-of-the-art on many different tasks, and the field did stop since.
This learning can be done {simultaneously} on all tasks~\citep{liuMultiTaskDeepNeural2019}, in {sequence}~\citep{mahajanIdentificationSemanticallySimilar2020}, {mixed}~\citep{piergiovanniAnswerMeMultiTaskOpenVocabulary2022}, or {optimized per task} (e.g. hypernetwork~\citep{jiPatientOutcomeZeroshot2023}).
Multi-task learning can spread on {similar tasks from different domains and cross-language} (e.g. similar summarization tasks), or {related auxiliary tasks}~\citep{jinHooksHeadlineLearning2020} to improve different skills.

\subsubsection{Instruction fine-tuning}\label{sec_supervisedtraining_instructions}
can highly increase the number of tasks, reasoning capabilities and global performance of a model~\citep{chungScalingInstructionFinetunedLanguage2022} by finetuning a pre-trained multi-task model with example with instructions to better align with intent, expected task reasoning and format.

\subsubsection{Transfer learning, knowledge \& domain adaptation, continual learning} \label{sec_supervisedtraining_transfer}
leverage knowledge and/or logic of an already trained model to re-use it in a target domain or application, which we then fine-tune.
This enables faster adaptation to target usage and allows to address a task even when there is not enough data available.
Transfer learning could be further divided into inductive (related task) and transductive (same task, new domain) learning, unlabeled to labeled transfer (similar to unsupervised pre-training to fine-tuning), and feature and parameter transfer (capture high level concepts of domain).
A model can be very efficient on a given knowledge domain in 2021 but will be later enable to process new questions requiring new or updated facts, this can be addressed by continual learning techniques~\citep{keContinualLearningLanguage2023}, or adding context to question prompt (\autoref{single_pass_prompt_engineering}), or by training to use an external information retriever (e.g. \autoref{HP5}).

\subsubsection{Knowledge distillation (KD)} \citep{boreshbanImprovingQuestionAnswering2023}\label{sec_supervisedtraining_distillation}
enables a smaller, more efficient model (student) to be trained to imitate the predictions of a larger, more complex model (teacher), leveraging the knowledge learned by the larger model.
For a given QA question/answer pair, it can not only provide answer but rich information like confidence, attention map and activated features.
The KD can be jointly used with active learning to reduce even more the training examples needed.

\subsubsection{Active learning}\label{sec_supervisedtraining_active}
enables a language model to be trained on a small initial set of examples and then, in a iterative manner, the model can request additional labeled data based on its own uncertainty, in order to improve its accuracy on a given task with minimal effort (e.g. \citet{jukicSmoothSailingImproving2022}).
This highly reduces the manual creation and training time of a dataset when required.
This can also help to craft better examples and avoid over-fitting due to excessive examples on same subject.

\subsubsection{Meta learning} \label{sec_supervisedtraining_meta}
gives the ability to learn faster new tasks with lesser data and time, like "learning to learn"~\citep{thrunLearningLearnIntroduction1998}.
This ability is well illustrated in the capacity to learn instructions to be addressed to a language model with the example of Tk-INSTRUCT supporting >1600 NLP tasks from 76 types reaching a performance near SOTA supervised tasks~\citep{wangSuperNaturalInstructionsGeneralizationDeclarative2022}.

\subsubsection{Multi-view learning} \citep{liLearningDiverseDocument2022}\label{sec_supervisedtraining_multiview}
learns multiple representations or "views" of the same input data to improve the model's performance on a specific task.
The idea is to leverage those different representations to better capture different knowledge facets or aspects of the data, leading to a more nuanced and effective representation for the task at hand in order to improve the performance of the model.

\subsection{Post-training: Preference alignment and reinforcement}
\label{sec_posttraining_alignment}\label{sec_ImprovementLoop_and_kg_capitalization}
How to create a system able to align to human expectations, enable LLM to develop different reasoning strategies, solve questions step by step, to use tools more efficiently?
Recent progress in this last step of the LLM training pipeline has markedly improved capabilities: with the same pre-training knowledge, models learn to better capture human intent and expected answer structure, to reason step-by-step, to better parallel and sequence tool calls, and to discover new solving strategies; over time this forms a kind of “experience’’ for tackling harder questions.
Our survey identifies two main levers: preference alignment and reinforcement.

In practice, a strong post-training stack follows:
\begin{enumerate}
  \item \textbf{Supervised fine-tuning (SFT)} on expected answers optionally with rationales/instructions.
  \item \textbf{Preference optimization} to better match human preferences: RL-based (e.g., RLHF)~\citep{baiTrainingHelpfulHarmless2022}, PPO) or RL-free methods such as \textbf{DPO}~\citep{rafailovDirectPreferenceOptimization2023}, \textbf{ORPO}~\citep{hongORPOMonolithicPreference2024}, and \textbf{KTO}~\citep{ethayarajhModelAlignmentProspect2024}; efficient RL variants such as \textbf{ReMax}~\citep{liReMaxSimpleEffective2023}. Recent “\textbf{reinforcement fine-tuning (RFT)}’’ strengthens \emph{reasoning} itself (e.g., GRPO used in DeepSeek-R1~\citep{guoDeepSeekR1IncentivizesReasoning2025}).
  \item \textbf{Train verifiers, process reward models (PRMs)} to score final answers or \emph{intermediate steps}; these models are trained during post-training, and then \emph{applied at inference} together with \emph{self-consistency} decoding~\citep{cobbeTrainingVerifiersSolve2021, wangSelfConsistencyImprovesChain2022}. See also \autoref{sec_reasoning_time}.
\end{enumerate}

\subsubsection{Supervision sources: human-in-the-loop (RLHF) and AI-in-the-loop (RLAIF)}
\begin{figure}
\includegraphics[width=1.0\linewidth]{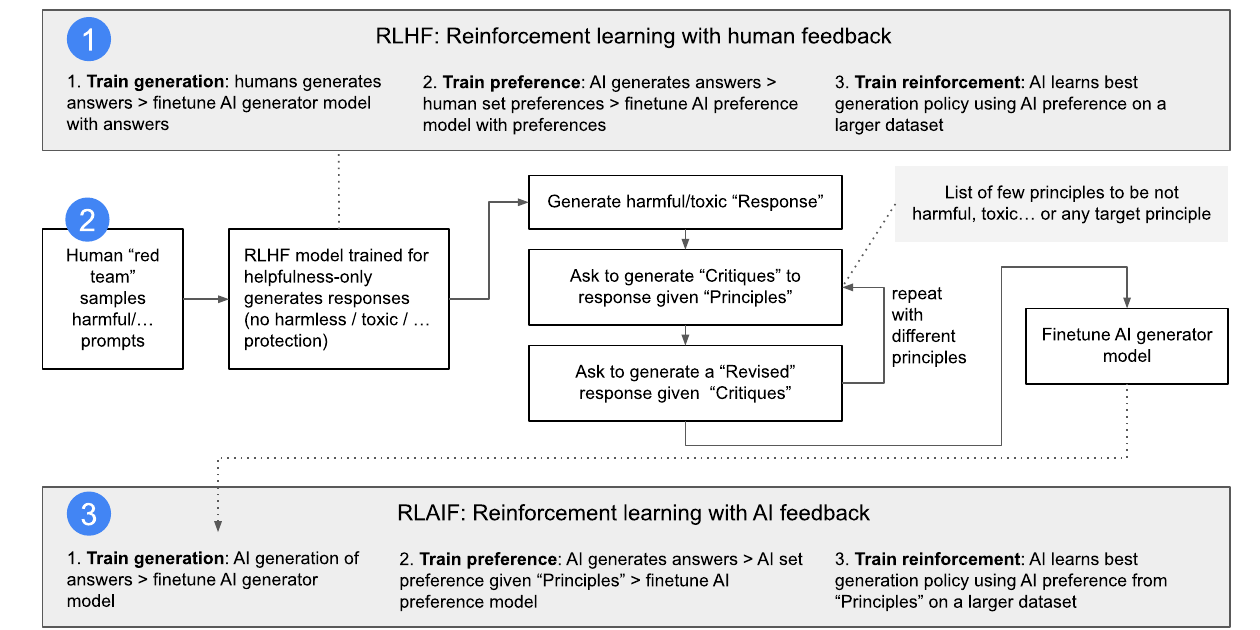}
\caption{From reinforcement learning with human feedback to AI feedback in order to scale and maximize helpfulness vs harmless tradeoff (\citet{ouyangTrainingLanguageModels2022, baiConstitutionalAIHarmlessness2022})}.
\label{RLHF CAI}
\end{figure}

Human-in-the-loop is meant to improve the outcome of an event or process via user input in the system loop.
Humans can intervene at many steps in a QA system from task definition or data creation, to final answer assessment.
We focus on the QA feedback loops.
Human can explicitly validate, rank, correct, provide guidance for answering (instructions), or implicitly rate via click-through.
The outcome of each question should fit the user’s intention, explicit (following instructions) or implicit (helpful, truthful, safe)~\citep{baiTrainingHelpfulHarmless2022}.
We therefore need to:
\begin{itemize}
    \item capture user rich explicit and implicit feedback through different human-in-the-loop input feedback.
    \item estimate and maximize user explicit and implicit intentions satisfaction for each answer and also in total through diverse reinforcement learning (RL) techniques or RL-free.
\end{itemize}
This is typically done through RLHF (reinforcement learning with human feedback)~\citep{ouyangTrainingLanguageModels2022a, baiTrainingHelpfulHarmless2022, daniels-kochExpertiseProblemLearning2022, tamkinTaskAmbiguityHumans2022, ganguliRedTeamingLanguage2022}.
While RLHF (often with PPO) is foundational but expensive, it is frequently complemented or replaced by simpler, more stable \textbf{RL-free preference learning methods} (\textbf{DPO}/\textbf{ORPO}/\textbf{KTO}) that learns directly from preferences without a separate reward model or on-policy RL~\citep{rafailovDirectPreferenceOptimization2023, hongORPOMonolithicPreference2024, ethayarajhModelAlignmentProspect2024}.
These include \textbf{Direct Preference Optimization (DPO)}~\citep{rafailovDirectPreferenceOptimization2023}, \textbf{Odds Ratio Preference Optimization (ORPO)}~\citep{hongORPOMonolithicPreference2024}, and \textbf{Kahneman-Tversky Optimization (KTO)}~\citep{ethayarajhModelAlignmentProspect2024},
which learn from preference data directly without a separate reward model or complex reinforcement learning.
RLHF can be enhanced by an AI supervision process to better scale (RLAIF),
reducing human workload and biases, as illustrated in the figure~\ref{RLHF CAI}.

Compared to GPT-3 with 175B parameters (SOTA general LLM model in 2022), InstructGPT with only 1.3B parameters but built with RLHF is more than 100 times smaller and answers are 85\% preferred by humans~\citep{ouyangTrainingLanguageModels2022}.

However, learning from human expertise has limits:
\begin{enumerate}
    \item \textbf{scaling cost selection}: manual labeling of data is slow and expensive so maybe restricted to some wealthy organizations or labeled with less expertise.
    \item \textbf{labeler biases}: longer RLHF training can bias language model with stronger political views and undesired goal pursuit.
    \item \textbf{expertise problem}: some queries require specific expertise; query–teacher selection matters.
    \item \textbf{harmlessness vs helpfulness trade-off}: optimizing one can hurt the other~\citep{baiConstitutionalAIHarmlessness2022}.
\end{enumerate}

\subsubsection{Reinforcement learning formulations (background)}
We can categorize reinforcement learning (RL)~\citep{suttonReinforcementLearningSecond2018} in this survey as follows:
\begin{itemize}
     \item \textbf{Reward-based reinforcement learning} (RL standard technique): the LM is trained to maximize a reward signal (e.g. positive or negative feedback) that is provided by a human or some other external source.
This could involve providing the model with a fixed reward whenever a correct answer is generated, or using more complex reward functions that take into account the quality and specificity of the model's answers.
A prominent modern application of this approach is \textbf{Reinforcement Fine-Tuning (RFT)}.
Pushing an LLM to reason via RFT can highly improve the answer quality of the same model.
RFT is often implemented using specific algorithms like \textbf{Group Relative Policy Optimization (GRPO)}, which optimizes policies with \emph{relative, group-wise baselines} to explicitly encourage longer, higher-quality \emph{reasoning tokens}.
For example, \textsc{DeepSeek-R1} used GRPO to \emph{incentivize reasoning} and reported strong gains without explicit chain-of-thought supervision~\citep{guoDeepSeekR1IncentivizesReasoning2025}.
     \item \textbf{Imitation learning} (includes procedure cloning~\citep{yangChainThoughtImitation2022}): the LM is trained to imitate the behavior of an expert (e.g. human, system...).
This can be a useful way to incorporate domain knowledge, search methodology, or other types of expertise into the model, and can help the model learn to generate high-quality answers more quickly by mimicking.
     \item \textbf{Inverse reinforcement learning}~\citep{zhouInverseReinforcementLearning2020}: here the reward is not direct, the LM attempts to infer the reward function from indirect human feedback or other forms of guidance.
This can be a more flexible approach, as it allows the model to learn from a wider range of feedback signals and to adapt to changing requirements over time.
Inverse RL could be classified as a type of imitation learning.
\end{itemize}

\subsubsection{Where to learn in the CQA pipeline}
This loop acts at multiple CQA stages; for each, we indicate whether it is mainly \textbf{Alignment} (pref.\ learning), \textbf{Reinforcement} (policy optimization), and/or \textbf{Reasoning} (step quality):
\begin{enumerate}
   \item \textbf{Question understanding / context} — learn to ask for clarifications and follow instructions \emph{(Alignment; Reinforcement)} via SFT + preferences/RLHF; e.g., instruction-following from \citet{ouyangTrainingLanguageModels2022a}.
   \item \textbf{Decomposition strategies} — learn to plan steps/tools, reward good plans \emph{(Reinforcement; Reasoning)}; PRMs can score intermediate steps; self-consistency encourages robust plans~\citep{wangSelfConsistencyImprovesChain2022}.
   \item \textbf{Query construction (prompting)} — optimize prompts, formats, and tool call arguments \emph{(Alignment)}; preference optimization (DPO/ORPO/KTO) improves adherence to desired styles~\citep{rafailovDirectPreferenceOptimization2023, hongORPOMonolithicPreference2024, ethayarajhModelAlignmentProspect2024}.
   \item \textbf{Information retrieval} — learn search/browse/citation behaviors \emph{(Reinforcement; Reasoning)} with browser-assisted QA and retrieval-grounded scoring (e.g., WebGPT)~\citep{nakanoWebGPTBrowserassistedQuestionanswering2022}; ReAct-style search couples reasoning and acting.
   \item \textbf{Answer generation} — align helpfulness/harmlessness and factuality \emph{(Alignment; Reinforcement; Reasoning)}: preference-based alignment (RLHF / DPO/ORPO/KTO), and verifier/PRM scoring trained post-training and \emph{applied at inference}; verifiers improve selection among candidates~\citep{cobbeTrainingVerifiersSolve2021}.
   \item \textbf{Knowledge capitalization} — prefer answers with evidence and coherent chains \emph{(Alignment; Reasoning)}; PRMs trained on step quality; at inference, \emph{self-consistency} and reranking select robust solutions~\citep{wangSelfConsistencyImprovesChain2022}.
\end{enumerate}

\subsection{Parameter-efficient tuning (PEFT)}\label{sec_PETtraining}
As per HELM study~\citep{liangetal.HolisticEvaluationLanguage2022} and compared models in the survey, larger models lead to better performance although some architecture allows to get them a bit smaller.
Those large-scale PLM are very expensive to retrain or just fine-tune.
The parameter-efficient tuning alternative targets a small fraction of model parameter update with similar performance than full-model fine-tuning, sometimes better~\citep{dingDeltaTuningComprehensive2022}.
We frame PEFT as selecting an \emph{update family} and \emph{deployment recipe} along four axes:
(A) memory/throughput budget, (B) structure and placement of updates, (C) task portfolio and domain shift, and (D) system constraints (context length, federation, personalization).

\paragraph{A. Budget: quantization‑aware and memory‑efficient recipes.}
Use low‑precision bases plus adapters when VRAM is the bottleneck. QLoRA combines 4‑bit quantization with low‑rank adapters~\citep{dettmersQLoRAEfficientFinetuning2023};
\emph{LoftQ} aligns quantization and adapter training~\citep{liLoftQLoRAFineTuningawareQuantization2023}.
\emph{LoRAM} reduces training memory via optimizer/checkpoint schedules~\citep{zhangTrainSmallInfer2024}.
Compressed designs (\emph{NOLA}, \emph{VB‑LoRA}) shrink adapter families by random‑basis combination or shared vector banks~\citep{koohpayeganiNOLACompressingLoRA2023, liVBLoRAExtremeParameter2024}.
\emph{Light‑PEFT} prunes early to preserve accuracy under tight budgets~\citep{guLightPEFTLighteningParameterEfficient2024}.
 
\paragraph{B. Structure: how to place and shape updates.}
Baseline LoRA injects rank‑$r$ adapters into attention/MLP projections~\citep{huLoRALowRankAdaptation2021};
expressivity results guide rank/depth choices~\citep{zengExpressivePowerLowRank2023}.
For more capacity at similar parameter counts, choose asymmetric factorization (\emph{HydraLoRA})~\citep{tianHydraLoRAAsymmetricLoRA2024}, Hadamard high‑rank transforms (\emph{HiRA})~\citep{huangHiRAParameterEfficientHadamard2024}, or gradient‑approximate updates (\emph{LoRA‑GA})~\citep{wangLoRAGALowRankAdaptation2024}.
If layer utility is heterogeneous, use rank allocation (\emph{ALoRA})~\citep{liuALoRAAllocatingLowRank2024};
randomized bases (\emph{RandLoRA}) regularize training~\citep{albertRandLoRAFullRank2024}.
Structured alternatives like \emph{RoCoFT} constrain updates to row/column subspaces to improve stability~\citep{kowsherRoCoFTEfficientFinetuning2025}.

\paragraph{C. Portfolio: multi‑task, domain, and composition.}
Adapters enable composition/fusion across tasks and languages~\citep{pfeifferMADXAdapterBasedFramework2020,heEffectivenessAdapterbasedTuning2021}.
For domain breadth, combine universal and domain-specific updates (\emph{MoDULA})~\citep{maMoDULAMixtureDomainSpecific2024}, and consider multi-task LoRA sharing with task-conditioned heads (\emph{MTL-LoRA})~\citep{yangMTLLoRALowRankAdaptation2025}.
Sparse expert routing integrates with LoRA for multi-domain applications (\emph{MOELoRA})~\citep{liuWhenMOEMeets2024}.
Achievement-based training across LoRA experts (\emph{C-LoRAE}) supports complementary specialization~\citep{yuanCollaborativeMultiLoRAExperts2025}. For structure-heavy summarization and extended discourse, use \emph{RST-LoRA} and \emph{LongLoRA}~\citep{puRSTLoRADiscourseAwareLowRank2024,chenLongLoRAEfficientFinetuning2023}.

\paragraph{D. System constraints: federation, personalization, and retrieval.}
In federated settings, residual adapter aggregation respects data locality and communication limits (\emph{FRLoRA})~\citep{yanFederatedResidualLowRank2024}.
For personalization, adapter selection and capacity control must balance privacy, drift, and latency~\citep{bragaPersonalizedLargeLanguage2024}.
If the shift is knowledge-intensive, retrieval may dominate or complement PEFT; comparative studies offer guidance on PEFT vs.\ retrieval integration~\citep{ficekGPTVsRETRO2024,leeInferenceScalingBridging2025}.

\paragraph{Editing rather than tuning.}
Where the goal is targeted concept adjustment with minimal retraining, representation editing (\emph{RED}) can be effective~\citep{wuAdvancingParameterEfficiency2024}.
Activation-sparsity control (\emph{DEFT}) reduces effective compute during adaptation~\citep{runwalPEFTDEFTParameter2025}.

\paragraph{Practical recipe.}
\emph{Default:} QLoRA or LoftQ for memory budget; LoRA with rank guided by expressivity and validation~\citep{dettmersQLoRAEfficientFinetuning2023,liLoftQLoRAFineTuningawareQuantization2023,zengExpressivePowerLowRank2023}. 
\emph{If capacity is limiting:} asymmetric/high-rank (\emph{HydraLoRA}, \emph{HiRA}) or dynamic rank (\emph{ALoRA})~\citep{tianHydraLoRAAsymmetricLoRA2024,huangHiRAParameterEfficientHadamard2024,liuALoRAAllocatingLowRank2024}. 
\emph{For many related tasks/domains:} composition (\emph{MoDULA}, \emph{MTL-LoRA}, \emph{MOELoRA}, \emph{C-LoRAE})~\citep{maMoDULAMixtureDomainSpecific2024,yangMTLLoRALowRankAdaptation2025,liuWhenMOEMeets2024,yuanCollaborativeMultiLoRAExperts2025}. 
\emph{For long-context:} \emph{LongLoRA}, \emph{RST-LoRA}~\citep{chenLongLoRAEfficientFinetuning2023,puRSTLoRADiscourseAwareLowRank2024}. 
\emph{For edge/federated:} \emph{FRLoRA}~\citep{yanFederatedResidualLowRank2024}. 
Classic adapters and prompt/prefix tuning remain strong baselines and compose with the above~\citep{heEffectivenessAdapterbasedTuning2021,lesterPowerScaleParameterEfficient2021}.

\subsection{Techniques for improving training}\label{sec_improvetraining}
Additionally to the different training options at each phase, these complementary techniques have been proven effective for improving training speed, quality, efficiency and safety:

    \emph{Regularization and adaptive computation}: as in other deep learning settings, dropout, weight decay and early stopping curb over‑fitting during training.
    For \emph{inference‑time} efficiency at similar accuracy, confidence‑based \emph{early‑exit cascades} and adaptive decoding allocate more compute only to hard inputs~\citep{dohanLanguageModelCascades2022, schusterConfidentAdaptiveLanguage2022}.

    \emph{Mixture-of-denoisers}: a small extra round of training ($\approx 0.1\%$ of pre‑training compute) with \emph{mixtures of denoisers} yields sizable gains in accuracy and reasoning~\citep{tayTranscendingScalingLaws2022}.

    \emph{Compute‑optimal scaling \& token budget}: training at the \emph{Chinchilla} compute‑optimal point (balancing model size and total tokens) systematically improves downstream performance for a fixed budget~\citep{hoffmannTrainingComputeOptimalLarge2022}.

    \emph{Domain/data‑mixture routing}: mixing domains explicitly during pre‑training (e.g., \textbf{DEMix} domain experts) and embarrassingly‑parallel expert training with \textbf{Branch‑Train‑Merge} reduce negative interference and improve generalization under domain shift~\citep{gururanganDEMixLayersDisentangling2021, liBranchTrainMergeEmbarrassinglyParallel2022}. Sparse \emph{Mixture‑of‑Experts} further increases capacity at fixed FLOPs~\citep{fedusSwitchTransformersScaling2022}.

    \emph{Length generalization (train‑short, test‑long)}: positioning schemes such as \textbf{ALiBi} enable training on short sequences while generalizing to much longer contexts at inference, reducing training cost for long‑context models~\citep{pressTrainShortTest2022}.

    \emph{Tasks disambiguation for generalization}: \citet{tamkinTaskAmbiguityHumans2022} train on ambiguous examples in different contexts to dramatically improve the accuracy of LLMs trained without large-scale human feedback training (RLHF).

    \emph{Post‑training pruning}: \citet{frantarSparseGPTMassiveLanguage2023} prune model size by 50\% of a very large model (GPT family) in one-shot without any retraining with a minimal loss of accuracy.

    \emph{Cross‑lingual learning}: training with cross‑lingual objectives (e.g., InfoXLM) helps learn language‑agnostic features and improves transfer and robustness across languages~\citep{chiInfoXLMInformationTheoreticFramework2021}.

    \emph{Red teaming for safety}: automatic red teaming generates stress‑tests to uncover unsafe failure modes and improve safety alignment~\citep{perezRedTeamingLanguage2022}.

    \emph{Data restructuring and curation}: re‑structuring pre‑training corpora to expose clearer supervision signals (e.g., metadata, task‑like formats) improves learning efficiency and downstream QA quality~\citep{yuanReStructuredPretraining2022}.

    \emph{Training on synthetic data and teacher/student}: leveraging high-quality synthetic data, often generated by more capable "teacher" models, to distill complex reasoning, coding, or mathematical skills into a smaller "student" model. This "textbooks are all you need" approach has proven highly effective for creating strong, specialized models~\citep{gunasekarTextbooksAreAll2023}.

    \emph{Parameter‑efficient adaptation}: as per previous section, for low‑budget specialization, parameter‑efficient fine‑tuning (e.g. LoRA) require very low training memory and duration while retaining quality; see \autoref{sec_PETtraining} for details~\citep{dettmersQLoRAEfficientFinetuning2023, liLoftQLoRAFineTuningawareQuantization2023}.

\section{Solving with hybridization patterns}\label{sec_hybridLLMpatterns}\label{sec_architectural_patterns}

To address the different limits of LLMs and skills identified for complex QA, we referenced architectural components which could augment a general-purpose base LLM such as a task-specific fine tuned model, a search engine, a software, a code interpreter...
To help design the different ways to increase an LLM both at inference or training, even if some may overlap, we propose to classify them into the following list of key hybrid architectural patterns each with description, strengths (\textbf{S}), weaknesses (\textbf{W}), references (\textbf{e.g.}).


{
\begin{longtable}{p{1\linewidth}}
\label{Table2LLMArchPatterns}\\

{\subsection{LLM base transformer}\label{HP1}
It is the general-purpose backbone, pretrained on web-scale corpora, ready for hybridization by extra layers/tools/retrieval/verifications/... to extend capabilities \citep{devlinBERTPretrainingDeep2019, brownLanguageModelsAre2020, raffelExploringLimitsTransfer2020}.}
\\\midrule
{\textbf{Strengths}: leverage knowledge from large unstructured data, can handle wide and versatile knowledge, some long-range dependencies and reasoning, and serves as the foundation for all subsequent architectures.
 \newline \textbf{Weaknesses}:} training is too long/costly to allow frequent update, prones to hallucinate with confidence, limited reasoning without massive scale, no mechanism to protect sensitive data.
 \newline \textbf{e.g.} encoders BERT~\citep{devlinBERTPretrainingDeep2019} or its optimized version RoBERTa~\citep{liuRoBERTaRobustlyOptimized2019}; decoders GPT2/GPT-3~\citep{brownLanguageModelsAre2020}; encoders-decoders T5~\citep{raffelExploringLimitsTransfer2020} or BART. \\
\end{longtable}}

\subsection{ADD Task specific adaptation, tuning, routing}
These patterns specialize or route the base model for \emph{specific tasks/domains}.
{
\begin{longtable}{p{1\linewidth}}
{\subsubsection{LLM + Task-specific head}\label{HP2}:
attach supervised head (classification/QA/CRF, etc.) to specialize base representations for a target task by training on a labeled dataset~\citep{lewisBARTDenoisingSequencetoSequence2019}} \\\midrule
{\textbf{Strengths}: achieves SOTA performance for targeted task and/or domain with lower computational resources than retraining an entire LLM.    
 \newline \textbf{Weaknesses}:} requires structured dataset for training, limited to the specific task it is designed for, may struggle with more general or open-ended tasks.
 \newline \textbf{e.g.} BART with classification head for question answering~\citep{lewisBARTDenoisingSequencetoSequence2019} \\

{\subsubsection{LLM + Prompt tuning module}\label{HP3}:
learns best prompt, context, instructions to query a LLM for given tasks and domains (or to better fine-tune) without modifying LLM weights}~\citep{khattabDSPyCompilingDeclarative2023, lesterPowerScaleParameterEfficient2021}. Prompting is further studied in \ref{sec_prompting}. \\\midrule
{\textbf{Strengths}: improve LLM performance on single or multiple task with no retraining, and can dynamically adapt reasoning skills according to context.
 \newline \textbf{Weaknesses}:} highly sensitive to slight prompt variations, may require substantial context, finding the right/robust prompt can be complex.
 \newline \textbf{e.g.} prompt optimization programming~\citep{beurer-kellnerPromptingProgrammingQuery2022, khattabDSPyCompilingDeclarative2023, yuksekgonulTextGradAutomaticDifferentiation2024, chengTraceNextAutoDiff2024}, programmatic retrieval-augmented in-context learning~\citep{khattabDemonstrateSearchPredictComposingRetrieval2023}, instructions generation~\citep{wangSelfInstructAligningLanguage2022}. \\

{\subsubsection{LLM + Router or Task discriminator}\label{HP12}:
routes task/domain to best model/tool/expert with prompt instructions \& context}~\citep{xuUniversalDiscriminatorZeroShot2022, liBranchTrainMergeEmbarrassinglyParallel2022} \\\midrule
{\textbf{Strengths}: can accelerate LLM training, improve inference LLM performance by routing to the most appropriate model (performance vs resource) with most appropriate instructions.
\newline \textbf{Weaknesses}:} complex to implement and maintain, shared reasoning and long term dependencies might be compromised.
\newline \textbf{e.g.} universal discriminator for zero-shot generalization~\citep{xuUniversalDiscriminatorZeroShot2022}, Branch-Train-Merge for fast parallel training of experts LM~\citep{liBranchTrainMergeEmbarrassinglyParallel2022}.
\\
\end{longtable}}

\subsection{ADD Decomposition, multi-processing \& fusion}
These patterns enable \emph{multi-step problem solving} by planning/decomposing tasks, chaining calls, composing/ensembling models, and adding generator–verifier loops for robust selection or fusion~\citep{yaoReActSynergizingReasoning2022, dohanLanguageModelCascades2022, wangSelfConsistencyImprovesChain2022, cobbeTrainingVerifiersSolve2021}.

{
\begin{longtable}{p{1\linewidth}}

{\subsubsection{LLM + Question/task decompose, plan, act module}\label{HP4}:
efficient break down of complex tasks into addressable subtasks following a plan, optionally revised from observations}.
\\\midrule  
{\textbf{Strengths}: efficient at solving more complex tasks requiring multiple steps or sources by converting into several manageable subtasks and an efficient resolution plan (a priori, iterative or recursive).
It can benefit from incorporating external knowledge sources or reasoning capabilities.
\newline \textbf{Weaknesses}:} more time and resources to implement, depending on implementation may struggle with long context and reasoning dependencies.
\newline \textbf{e.g.} agentic dynamic decomposition, like plan–act–observe–reflect, is a major practice (e.g., ReAct/ToT-style planners) using LLM or specific models for the task~\citep{shenHuggingGPTSolvingAI2023}, iterated decomposition w/ reasoning process supervision~\citep{reppertIteratedDecompositionImproving2023}, separated reasoning and acting~\citep{yaoReActSynergizingReasoning2022}, unsupervised QA decomposition~\citep{perezUnsupervisedQuestionDecomposition2020}, successive prompting decomposition for CQA~\citep{duaSuccessivePromptingDecomposing2022}. \\

{\subsubsection{Cascaded / chained / looped LLM}\label{HP10}:
solves by multiple LLM calls—possibly iterative/recursive—mirror and enable confidence-/cost-adaptive cascades \citep{dohanLanguageModelCascades2022, wuAIChainsTransparent2022}.}. \\\midrule  
{\textbf{Strengths}: facilitates human control over design and execution process, can solve higher complexity problems than LLM core skills by breaking down tasks \& design, provides a causal chain useful for explainability, allows optimization of the pipeline and leverage specialization to avoid potential bottlenecks or inefficiencies.
\newline \textbf{Weaknesses}:} may be less effective at tasks requiring extensive context and long reasoning dependencies.
\newline \textbf{e.g.} solving by cacasding language models~\citep{dohanLanguageModelCascades2022}, AI Chains~\citep{wuAIChainsTransparent2022}, in a collaborative visual chain of prompt~\citep{wuPromptChainerChainingLarge2022}, logical and robust reasoning with selection-inference~\citep{creswellSelectionInferenceExploitingLarge2022}, human readable multi-step logical deduction on scientific QA improving accuracy and faithfulness~\citep{creswellFaithfulReasoningUsing2022}, iterative prompting an LLM~\citep{wangIterativelyPromptPretrained2022, duaSuccessivePromptingDecomposing2022}. \\

{\subsubsection{LLM Ensembling and LLM composing}\label{HP17}:
combines diverse LLMs/modules via consensus or learned composition to boost robustness and reduce single-model idiosyncrasies \citep{liComposingEnsemblesPretrained2022, liBranchTrainMergeEmbarrassinglyParallel2022}.}.
\\\midrule
{\textbf{Strengths}: improves inference accuracy, generalization and stability by combining diversity.
 \newline \textbf{Weaknesses}:} cost \& complexity, be aware of trade-offs (e.g. additional computation).
 \newline \textbf{e.g.} ensemble learning for validation and explanation~\citep{huyAutoencodingLanguageModel2022},
    parallel training of expert LLM~\citep{liBranchTrainMergeEmbarrassinglyParallel2022},
    ensembles of LLMs via iterative consensus~\citep{liComposingEnsemblesPretrained2022}, automatic neural module composition~\citep{andreasNeuralModuleNetworks2016}.
\\

{\subsubsection{LLM + Generators/Verifiers}\label{HP15}:
couples a \emph{generator} proposing solutions with a \emph{verifier/judge} (plus self-consistency) that filters/reranks for improved solutions~\citep{cobbeTrainingVerifiersSolve2021, wangSelfConsistencyImprovesChain2022, liAdvanceMakingLanguage2022}.}
\\\midrule
{\textbf{Strengths}: can solve complex tasks (large gains on math/code, factual QA), out of training knowledge, by combining generation and adversarial skills.
 \newline \textbf{Weaknesses}:} resource-intensive \& costly; may not always lead to performance improvements proportional to effort.
\newline \textbf{e.g.} AlphaCode~\citep{liCompetitionLevelCodeGeneration2022}, DiVeRSe~\citep{liAdvanceMakingLanguage2022}, CoRe (cooperative reasoning)~\citep{zhuSolvingMathWord2022}, self-consistency for chain of thought~\citep{wangSelfConsistencyImprovesChain2022}, training verifiers to solve math~\citep{cobbeTrainingVerifiersSolve2021}.
\emph{self-grading RAG} (generate–critique–select) variants are often paired with retrieval.
\\
\end{longtable}}

\subsection{ADD External information \& reasoning}
They leverage external knowledge and tools:
These patterns \emph{augment} the LLM with external knowledge, tools and computation—memories~\citep{borgeaudImprovingLanguageModels2022, izacardAtlasFewshotLearning2022, schickToolformerLanguageModels2023, gaoPALProgramaidedLanguage2023, openaiGPT4TechnicalReport2023}.
{
\begin{longtable}{p{1\linewidth}}

{\subsubsection{LM + Memory}\label{HP14}:
an external working memory stores and retrieves facts, summaries, and states across steps to overcome context limits and support long-horizon reasoning/dialogue \citep{schuurmansMemoryAugmentedLarge2023, xuGoldfishMemoryLongTerm2021}.}
\\\midrule
{\textbf{Strengths}: without LLM modification can simulate any algorithm, process unbounded inputs, strengthen controllability, long-term dependencies (for reasoning, dialogs, summarization, retrieval, algorithmic...) and robustness by incorporating counterfactual \& irrelevant contexts.
\newline \textbf{Weaknesses}:} scalability issues with increasing model size.
\newline \textbf{e.g.} universal memory augmented large LM~\citep{schuurmansMemoryAugmentedLarge2023}, Recurrent Memory Transformer~\citep{bulatovRecurrentMemoryTransformer2022}, long-term open-domain conversations~\citep{xuGoldfishMemoryLongTerm2021}. \\

{\subsubsection{LLM + Semantic-unstructured Information Retriever}\label{HP5}:
retrieves (by similarity) contextual information from a source to be added to a question or task sent to an LLM in order to augment its knowledge~\citep{borgeaudImprovingLanguageModels2022, izacardAtlasFewshotLearning2022}.}
\\\midrule  
{\textbf{Strengths}: incorporates any up-to-date external sources without increasing LLM size, allowing much smaller models (RETRO is 1/25 size of GPT-3 for same perforamnce), control over sources (sensitivity, explainability, knowledge update) and a variety of retrieval techniques.
 \newline \textbf{Weaknesses}:} may struggle with tasks requiring abstract or creative reasoning, may be limited by the quality and coverage of the external sources; in practice, \emph{self-/corrective-RAG} couples generation with critique and retrieval-backed verification to mitigate these issues.
 \newline \textbf{e.g.} Deepmind RETRO~\citep{borgeaudImprovingLanguageModels2022}, Atlas~\citep{izacardAtlasFewshotLearning2022}, training RL agents to query external knowledge~\citep{liuAskingKnowledgeTraining2022}; RAG is now a default backbone in many QA systems (often combined with verifiers).
\\

{\subsubsection{LLM + Symbolic-structured Information Retriever}\label{HP6}:
queries structured sources (ontologies, graphs, database) to extract structured information to be added to a question or task sent to an LLM in order to augment its knowledge~\citep{xieUnifiedSKGUnifyingMultiTasking2022, pramanikUNIQORNUnifiedQuestion2022}.}
\\\midrule  
{\textbf{Strengths}: allows cold start tasks and fast domain adaptation with low data, follows rules and concepts more effectively.
 \newline \textbf{Weaknesses}:} neuro-symbolic integration is complex, creating symbolic data is time-consuming and labor-intensive.
 \newline \textbf{e.g.} UNIQORN~\citep{pramanikUNIQORNUnifiedQuestion2022}, Heterformer~\citep{jinHeterformerTransformerArchitecture2022}, UnifiedSKG~\citep{xieUnifiedSKGUnifyingMultiTasking2022}; \textbf{RAG with Symbolic/Structured Retriever} (e.g., \emph{GraphRAG}) for graph-guided global+local search across narrative corpora.
 \\

{\subsubsection{LLM + Tools}\label{HP7}:
delegates sub-tasks to external tools (search engines, solvers or simulators) on tasks hard for LLM \citep{nakanoWebGPTBrowserassistedQuestionanswering2022, schickToolformerLanguageModels2023, liuMindsEyeGrounded2022}.}
\\\midrule
{\textbf{Strengths}: Leverages proven performance and robustness of external software/services in information retrieval (IR), logic, world modelling. 
\newline \textbf{Weaknesses}:} challenging end-to-end learning and potential complex integration (e.g. additional pre-processing or post-processing steps) \newline \textbf{e.g.} physics with MindsEye~\citep{liuMindsEyeGrounded2022}, WebGPT~\citep{nakanoWebGPTBrowserassistedQuestionanswering2022}, SeeKeR~\citep{shusterLanguageModelsThat2022}, Toolformer~\citep{schickToolformerLanguageModels2023}; mainstream APIs expose \emph{function calling} to orchestrate tools.
\\

{\subsubsection{LLM + Code interpreter}\label{HP8}:
generates code to be executed in order to delegate complex or heavy processing tasks (LLM can also learn complex logics by program's input/output traces)}~\citep{gaoPALProgramaidedLanguage2023, droriNeuralNetworkSolves2022, haluptzokLanguageModelsCan2022}
\\\midrule  
{\textbf{Strengths}: leverage robust reasoning \& algorithmic capabilities, and language ecosystem.
\newline \textbf{Weaknesses}:} may struggle with tasks requiring deeper understanding of context or concepts, dependency on external code interpreters.
\newline \textbf{e.g.} PAL~\citep{gaoPALProgramaidedLanguage2023}, solving math problems via cooperative reasoning~\citep{zhuSolvingMathWord2022} or program synthesis~\citep{droriNeuralNetworkSolves2022}, LM self-improve its programming capabilities~\citep{haluptzokLanguageModelsCan2022}, AlphaCode~\citep{liCompetitionLevelCodeGeneration2022}.
\\

{\subsubsection{LLM + Multimodal}\label{HP16}:
integrates and reasons over non-textual inputs (image, audio, video, sensors...).}~\citep{openaiGPT4TechnicalReport2023, shenHuggingGPTSolvingAI2023, yangEnhancingMultimodalMultihop2022}
\\\midrule  
{\textbf{Strengths}: leverages knowledge from non-textual sources and combines each modality to enhance understanding, reasoning and problem-solving.
 \newline \textbf{Weaknesses}:} integration complexity (representation, alignment, reasoning, generation...) \& cost, increased difficulty in addressing explainability and hallucination issues
 \newline \textbf{e.g.} foundation models and recent trends in multimodal ML~\citep{shenHuggingGPTSolvingAI2023, yangEnhancingMultimodalMultihop2022, openaiGPT4TechnicalReport2023}.
\\

{\subsubsection{LLM + Veracity/evidence checker}\label{HP11}:
provides sources and facts assessment} \citep{menickTeachingLanguageModels2022, chenLORENLogicRegularizedReasoning2022, guoSurveyAutomatedFactChecking2022}
\\\midrule
{\textbf{Strengths}: guarantees information credibility \& reliability while mitigating hallucinations by providing verifiable sources and evidence assessment.
\newline \textbf{Weaknesses}:} limited by the quality and coverage of external sources, do not eliminate risk of misinformation.
\newline \textbf{e.g.} GopherCite supports answers with verified quotes~\citep{menickTeachingLanguageModels2022}, logic-regularized reasoning for interpretable fact verification~\citep{chenLORENLogicRegularizedReasoning2022}, survey on automated fact-checking~\citep{guoSurveyAutomatedFactChecking2022}, hallucinated content detection~\citep{zhouDetectingHallucinatedContent2021}, RL approach for explainability using entailment trees~\citep{liuRLETReinforcementLearning2022}; \emph{self-/corrective-RAG} couple generation with critique and retrieval-backed verification.
\\

{\subsubsection{LLM + Temporal/spatial reasoning}\label{HP19}:
provides temporal (and spatial) reasoning layers, helping with ordering, duration and event queries}~ \citep{wangTimeBERTExtendingPreTrained2022, fayekTemporalReasoningAudio2020}
\\\midrule
{\textbf{Strengths}: allows temporal estimation, ranking \& clustering, reasoning, incoherence detection, addresses knowledge forget and update.
\newline \textbf{Weaknesses}:} few resources, efficient integration might be challenging.
\newline \textbf{e.g.} TimeBERT: Extending Pre-Trained Language Representations with Temporal Information~\citep{wangTimeBERTExtendingPreTrained2022}, improves temporal reasoning through added audio modality~\citep{fayekTemporalReasoningAudio2020}.
\\
\end{longtable}}

\subsection{ADD Feedback, dialog, reflection \& learning from teacher}
These patterns incorporate \emph{critique and guidance} to align behavior and improve reasoning~\citep{ouyangTrainingLanguageModels2022, baiTrainingHelpfulHarmless2022, huTeacherStudentArchitectureKnowledge2022, pengGODELLargeScalePreTraining2022, shinnReflexionAutonomousAgent2023}.
{
\begin{longtable}{p{1\linewidth}}
{\subsubsection{LLM + Human/AI RL feedback}\label{HP9}:
learns alignment/preference and optimal policy for goals (answer quality, safety, data sources...)}.
\\\midrule  
{\textbf{Strengths}: support in most critical challenge of LLMs design such as human expectation alignment and personalization, safety and quality control.
\newline \textbf{Weaknesses}: human feedback and convergence is highly time-consuming. This could be mitigated by incorporating active learning and AI feedback learning, but that means added complexity.
\newline \textbf{e.g.} reinforcement learning with human feedback RLHF~\citep{ouyangTrainingLanguageModels2022, baiTrainingHelpfulHarmless2022, daniels-kochExpertiseProblemLearning2022}, with AI feedback RLAIF~\citep{baiConstitutionalAIHarmlessness2022}, experts imitation learning~\citep{yangChainThoughtImitation2022}, web search RL in WebGpt~\citep{nakanoWebGPTBrowserassistedQuestionanswering2022}, citation RL in GopherCite~\citep{menickTeachingLanguageModels2022}.} \\

{\subsubsection{LLM + Dialog module}\label{HP13}:
enable dialog/state tracking, clarification, disambiguation, and long-context control for interactive complex QA \citep{pengGODELLargeScalePreTraining2022, chenOPALOntologyAwarePretrained2022, khotLearningSolveComplex2021}}
\\\midrule
{\textbf{Strengths}: enhances understanding of complex contexts, problems and concepts through human interaction, guidance, progressive refinement and problem-solving.
\newline \textbf{Weaknesses}: might not fit targeted usage format, more time and resources implementation.
\newline \textbf{e.g.} GODEL~\citep{pengGODELLargeScalePreTraining2022}, OPAL~\citep{chenOPALOntologyAwarePretrained2022}, CommaQA~\citep{khotLearningSolveComplex2021}, ChatGPT.} \\

{\subsubsection{LLM + (multi) Teacher}\label{HP18}:
develop new LLM capacities through expert teaching or supporting on domains/skills/tasks~\citep{huTeacherStudentArchitectureKnowledge2022, wuOneTeacherEnough2021}}
\\\midrule  
{\textbf{Strengths}: accelerates and improves knowledge learning, expansion, adaption, multi-tasking, reinforces reasoning capabilities (e.g. temporal).
\newline \textbf{Weaknesses}: cost \& complexity, control of teacher knowledge transfer (e.g. domain scope, biases).
\newline \textbf{e.g.} Teacher-student architecture survey~\citep{huTeacherStudentArchitectureKnowledge2022}, learning from multiple teachers~\citep{wuOneTeacherEnough2021}.} \\

{\subsubsection{LLM + Reflection module}\label{HP20}:
improves reasoning and reliability by self-critic/analysis to revise its own steps/outputs using episodic memory\citep{shinnReflexionAutonomousAgent2023}.} \\\midrule
{\textbf{Strengths}: boost LLM internal capacities by self-awareness, thoughts, given goal and reasoning trace.
\newline \textbf{Weaknesses}: require additional resources (e.g. external memory and management, heuristic to trigger self-reflection).
\newline \textbf{e.g.} \citet{shinnReflexionAutonomousAgent2023} present a reflection agent allowing a LLM to beat previous results by a large margin.} \\

\end{longtable}}

\small


\section{Solving with prompting}\label{sec_solvingCQA}\label{sec_prompting}
Now that we have trained models to acquire the required skills to solve a complex QA or rely on hybridization, let's see how to design questions (i.e. prompt) with proper instructions and context.
At the outset, we distinguish \emph{tool-use prompting}, covered separately in \autoref{sec_tool_use_prompting}, because it explicitly bridges prompting with hybridization and agentic control (\autoref{HP7}, \autoref{HP8}, \autoref{sec_agentic}).

\subsection{Prompt design (basics)}
LLM can highly improve its ability to solve a problem by leveraging information provided with a question ("problem well stated is half solved").
Engineering a good prompt (text provided to a LLM when posing a question) can rival or beat model finetuning on QA in many cases~\citep{wangSuperNaturalInstructionsGeneralizationDeclarative2022}.
Different information can be added to improve answer success probability while posing a question to a LLM such as \textbf{additional context and knowledge, constraints, instructions and examples}...
Question can also be \textbf{designed to be answered in multiple steps}, for example just by asking to answer step by step~\citep{weiChainThoughtPrompting2022, kojimaLargeLanguageModels2023} or reasoning compositionally like humans~\citep{drozdovCompositionalSemanticParsing2022},~\citep{zhouLeasttoMostPromptingEnables2022, duaSuccessivePromptingDecomposing2022}.

\subsection{Context and protocol engineering}\label{sec_context_protocol}
Beyond single strings, modern prompting increasingly treats the \emph{context} $C$ as a structured payload that is \emph{assembled} at inference time from typed components:
(i) system/task instructions,
(ii) external knowledge (retrieved or generated),
(iii) tool/function interfaces,
(iv) short/long-term memory,
and (v) dialog/agent state.
The discipline of \emph{Context Engineering (CE)} systematizes this: retrieve/generate the right information, process/organize it, then manage/compress/persist it under context-length and latency constraints~\citep{meiSurveyContextEngineering2025}.
Concretely, CE decomposes into: \emph{(a) retrieval \& generation} (e.g., RAG, self-generated hints), \emph{(b) processing} (segmentation, ranking, self-refine, structured/relational integration), and \emph{(c) management} (memory hierarchies, compression, KV-cache policies)~\citep{meiSurveyContextEngineering2025}.
In practice this unifies patterns already surveyed here (RAG: \autoref{HP5}, memory: \autoref{HP14}, tools: \autoref{HP7}, multi-agent orchestration: \autoref{sec_agentic}). A key empirical observation is an \emph{asymmetry}: LLMs—augmented by CE—parse complex contexts well, yet often struggle to emit equally sophisticated, long, structured outputs without additional scaffolding~\citep{meiSurveyContextEngineering2025}.

\paragraph{Typed context assembly at prompt-time.} Useful operational decomposition is:
\[
  C = \underbrace{c_{\text{instr}}}_{\text{rules \& rubric}}
    \;\oplus\; \underbrace{c_{\text{know}}}_{\text{retrieved evidence}}
    \;\oplus\; \underbrace{c_{\text{tools}}}_{\text{function schemas}}
    \;\oplus\; \underbrace{c_{\text{mem}}}_{\text{episodic/summary memory}}
    \;\oplus\; \underbrace{c_{\text{state}}}_{\text{dialog/plan state}} .
\]
This “fielded prompting’’ supports: (i) \emph{retrieval gating} and \emph{self-critique} (e.g., Self-RAG control/reflection tokens for when/how to retrieve and whether the answer is grounded)~\citep{asaiSelfRAGLearningRetrieve2023}; (ii) \emph{schema-constrained outputs} (JSON, tables, checklists) that reduce drift; (iii) \emph{memory hooks} to carry forward distilled facts (\autoref{HP14}). Evaluation should pair \emph{answer correctness} with \emph{attribution/groundedness} (e.g., FActScore, RAGAS; see \autoref{sec_evaluation}).

\paragraph{From CE to \emph{Knowledge Protocol Engineering} (KPE).} For expert domains, factual context (RAG) is insufficient if the model lacks the \emph{procedure}. \emph{KPE} proposes translating human SOPs/manuals into a \emph{machine-executable knowledge protocol} that constrains and guides the LLM’s reasoning and tool use~\citep{zhangKnowledgeProtocolEngineering2025}. The goal is \emph{methodology injection}: endow a generalist LLM to “think like a specialist’’ by encoding workflow steps, decision trees, admissible tool calls, success/failure checks, and required artifacts (e.g., citations, tables, calculations).

\paragraph{KPE—minimal workflow.} (1) \emph{Protocol extraction}: elicit or mine domain SOPs (tasks, sub-tasks, invariants, acceptance criteria). (2) \emph{Encoding}: compile into structured, prompt-time templates (typed fields, role/rubric, tool schemas) and step validators (unit checks, equation or API assertions). (3) \emph{Execution}: drive the LLM/agent to produce \emph{typed intermediates} and invoke tools under the protocol; use retrieval to fill \emph{method} gaps as well as facts. (4) \emph{Verification \& audit}: ensure evidence-linked outputs and process-level logs (plan/act/observe) for human review. In regulated or high-stakes settings, the protocol acts as a \emph{methodological contract} that enables transparency and step-wise audit~\citep{zhangKnowledgeProtocolEngineering2025}.

\paragraph{When to favor KPE.} Use KPE when (i) domains have stable procedures (e.g., legal HHI analysis; scientific/biomedical pipelines), (ii) outputs must be auditable/reproducible, or (iii) multi-step tasks benefit from explicit role artifacts (requirements → design → code → tests) with typed checks. KPE composes naturally with CE (RAG/memory/tools) and agentic controllers (\autoref{sec_agentic}), and complements post-training/RFT (\autoref{sec_posttraining_alignment}, \autoref{sec_reasoning_time}).

\noindent \textit{Summary.} Context engineering provides the prompt-time \emph{information logistics} (what to include, how to compress/manage); KPE provides the \emph{methodological scaffold} (how to proceed and verify). Together they turn provide prompting with \emph{structured contextual information assembly and protocolized execution}, dovetailing with hybrid/agentic and inference-time strategies later in this survey~\ref{sec_agentic_and_reasoning}.

\subsection{In-Context Learning (ICL)}\label{sec_icl}
\emph{In-Context Learning}~\citep{brownLanguageModelsAre2020} refers to adapting a frozen LLM at \emph{inference time} by adding task instructions, information, demonstrations directly in the prompt. ICL is therefore an \emph{inference-time learning via a prompting strategy}, not training. ICL can be paired with retrieval capabilities to dynamically include \emph{(a)} demonstrations and \emph{(b)} evidence snippets (\autoref{HP5}); pair with verifier/attribution metrics in evaluation (e.g., FActScore/RAGAS).
ICL can (i) rapidly teach formats (e.g., evidence$\rightarrow$claim, tool-call JSON), (ii) steer reasoning styles (brief CoT, tabular working), and (iii) adapt to domain terminology without fine-tuning. It composes naturally with \emph{agentic} controllers that reuse demonstrations across steps and with \emph{reasoning-time} strategies that sample multiple chains then verify (\autoref{sec_reasoning_time}).

\textbf{When to prefer fine-tuning/PEFT instead ?} If the task requires strict latency, or needs domain generalization and generalization beyond information retrieval, prefer SFT/PEFT (\autoref{sec_PETtraining}, \autoref{sec_supervisedtraining_instructions}) and keep ICL for light adaptation and formatting control.

\textbf{Common failure modes (and mitigations).}
\begin{itemize}
  \item \emph{Spurious cues / overfitting to exemplars.} Mitigate with diverse demos and retrieval gating; add a critic/verifier step (\autoref{HP11}, \autoref{HP15}).
  \item \emph{Context overrun / dilution.} Trim boilerplate and prefer structured fields; chunk evidence and gate retrieval (\emph{self-/corrective-RAG}, \autoref{HP5}).
  \item \emph{Evaluation pitfalls.} Avoid leakage between demo pool and test; report with and without CoT; for grounded QA, include attribution metrics (FActScore/RAGAS) alongside accuracy.
\end{itemize}

\noindent\textbf{Summary.} We treat ICL as a first-class \emph{prompting} tool: fast, controllable, and synergistic with retrieval and reasoning-time allocation. It stays here (not in training), with cross-references to PEFT (when to train) and to agentic controllers (when to route or verify).

\subsection{Tool-use prompting (program-aided \& API calls)}\label{sec_tool_use_prompting}
Beyond optimizing the text of a prompt, a distinct family of prompting patterns \emph{instructs models to invoke external tools} (e.g., code interpreters, APIs, simulators) during inference. This bridges prompting with hybridization/agents by turning the model into a planner that delegates sub-steps to executors~\citep{schickToolformerLanguageModels2023, yaoReActSynergizingReasoning2022, shenHuggingGPTSolvingAI2023}.
Representative instances include using a \textbf{code interpreter} to carry out precise computation and structured transformations~\citep{chenProgramThoughtsPrompting2022, madaanLanguageModelsCode2022, gaoPALProgramaidedLanguage2023}, and \textbf{domain simulators} (e.g., physics engines) for grounded reasoning about the world~\citep{liuMindsEyeGrounded2022, jacksonNaturalLanguageSimulations2023}.
This section focuses on the prompting interfaces (how to elicit tool calls); architectural aspects of tools are detailed in \autoref{HP7}, \autoref{HP8}, and agentic control in \autoref{sec_agentic}.

This is a fast growing field such as the use of code interpreter to enrich LLM answers for complex structures and calculations~\citep{lyuFaithfulChainofThoughtReasoning2023, chenProgramThoughtsPrompting2022, madaanLanguageModelsCode2022, gaoPALProgramaidedLanguage2023}; simulators (e.g. physics engine) to simulate processes and aid LMs in real-world reasoning~\citep{liuMindsEyeGrounded2022, jacksonNaturalLanguageSimulations2023}; or any dedicated software tool or AI model~\citep{schickToolformerLanguageModels2023, shenHuggingGPTSolvingAI2023}.

\subsection{Enhancing LLM knowledge or skills via prompting}
Prompt can be engineered in order to enhance model capacities such as (according to taxonomy from \citet{qiaoReasoningLanguageModel2022}):
\textbf{knowledge retrieval} which could be divided into implicit knowledge to enrich LLM answers through few-shot prompting and reinforcement learning~\citep{liuRainierReinforcedKnowledge2022, liuGeneratedKnowledgePrompting2022}, and explicit knowledge to retrieve and provide in-context labeled examples in prompt to improve explicit knowledge, reduce hallucination and enrich LLM answers~\citep{suSelectiveAnnotationMakes2022};
\textbf{Arithmetic reasoning} with many different type of reasoning and approaches (e.g. \citep{chenProgramThoughtsPrompting2022, huangLargeLanguageModels2022, liAdvanceMakingLanguage2022, gaoPALProgramaidedLanguage2023, beurer-kellnerPromptingProgrammingQuery2022});
\textbf{Commonsense reasoning} on general background knowledge and reasoning (e.g. \citep{madaanLanguageModelsCode2022, liuRainierReinforcedKnowledge2022, kojimaLargeLanguageModels2023, wangSelfConsistencyImprovesChain2022, liuGeneratedKnowledgePrompting2022});
\textbf{Creativity reasoning} to support inventive process like automated diverse prompting ideas (e.g. \citep{leePromptiverseScalableGeneration2022});
\textbf{Logical reasoning} learnt from synthetic rule bases, entailment trees, or diagnostic benchmarks~\citep{creswellSelectionInferenceExploitingLarge2022, creswellFaithfulReasoningUsing2022}; 
\textbf{Symbolic reasoning}~\citep{gaoPALProgramaidedLanguage2023, kojimaLargeLanguageModels2023, wangSelfConsistencyImprovesChain2022} uses examples that contain symbolic rationales, rules;
\textbf{Multimodal reasoning} trained on dataset such as ScienceQA, ALERT (e.g. \citep{luLearnExplainMultimodal2022}).

\subsection{Optimization}
In this section, we review standard prompt based optimization strategies.
\subsubsection{Single-pass prompt optimization} aims to get the best answer from one prompt:\label{single_pass_prompt_engineering}
    \begin{description}
            \item [Zero-shot prompting] provides a prompt for task without prior training on this task and no additional context or guidance.
            \item [In-Context learning and Few-shot Prompting] provides prompt with relevant context~\citep{brownLanguageModelsAre2020} or demonstration~\citep{minRethinkingRoleDemonstrations2022} for expected task helping LLM to better answer.
Most well-known example is few-shot learning which provides a prompt with a few examples of expected task for helping to generate the best answer.
            \item [(Hard) prompt tuning] adjusts the initial prompt, often by trial and error, to improve answer accuracy.
This improving process can be manual, automated or even programmed~\citep{beurer-kellnerPromptingProgrammingQuery2022, chengTraceNextAutoDiff2024, khattabDSPyCompilingDeclarative2023}.
            \item [Soft prompt tuning] creates soft prompts~\citep{lesterPowerScaleParameterEfficient2021} which are concatenated to the input text.
Tokens of this soft prompt are learned vectors optimized end-to-end over a training dataset.
\autoref{sec_PETtraining} provides some additional details.
Some innovative examples are:
                \begin{itemize}
                    \item[--] {Exploring Universal Intrinsic Task Subspace via Prompt Tuning}: adapt to many NLP tasks with small-scale data by optimizing only a few free parameters in a unified low-dimensional intrinsic task subspace~\citep{qinExploringUniversalIntrinsic2022}.
                    \item[--] {Compositional Task Representations} learn specific codebook for compositional tasks~\citep{shaoCompositionalTaskRepresentations2023}.
                \end{itemize}
            \item [Chain-of-thought prompting]~\citep{weiChainThoughtPrompting2022} ask to reason step by step and can provide relevant examples of multi-steps of reasoning/thoughts up to the solution to improve reliability or more easily spot errors in the result.
It largely outperforms the state-of-the-art results with zero and few-shots learning with the same model on many advanced natural language processing tasks and fine-tuned models trained with hundreds times of examples, with the advantage of being interpretable.
            \item [Chain-of-hindsight or contrastive prompting]~\citep{liuChainHindsightAligns2023} provides examples with qualitative feedback (e.g. comparison and critiques) to better align the model output to preferences.
It is mostly used for finetuning model but could be used when prompting.
            \item [Self-reflection]~\citep{shinnReflexionAutonomousAgent2023} prompts the model to improve its next action given its previous track and mistakes.
    \end{description}

\subsubsection{Multi-step prompt optimization} designs a solving process in progressive prompt steps:
    \begin{description}
        \item [Least-to-most prompting]~\citep{zhouLeasttoMostPromptingEnables2022} improves chain-of-thought with multi-step examples that gradually becomes more specific or complex; Chain-of-thought often performs poorly on tasks requiring to solve problems harder than those in demonstration examples.
To tackle this, LtM first reduces a complex problem into a list of easier subproblems, and then sequentially solves these subproblems with gradual complexity.
LtM can be combined with self-consistency to improve robustness.
\textbf{Dynamic least-to-most prompting (compositionality)}~\citep{drozdovCompositionalSemanticParsing2022} refines it with the following steps:(1) prompts LLM to perform a synthatic parsing to create a tree-structured decomposition, (2) select matching demonstration examples, (3) linearize decomposition tree and prompt to sequentially generate answers to subproblems.
        \item [Successive prompting]~\citep{duaSuccessivePromptingDecomposing2022} develops successive prompting decomposing a complex problem into a first simple problem, with each next subproblem prediction having access to the answers to each previous subproblems.
        \item [Maieutic prompting]~\citep{jungMaieuticPromptingLogically2022} is inspired by Socratic way of questioning, it generates a tree of logical explanations up to the truth values that max-satisfy these relations to verify its veracity.
It surpass many approaches and provides intrinsic interpretations of inference.
    \end{description}

\subsubsection{Process- and verification-based optimization} is designed to follow a parallel or iterative process optimizing final output.
    \begin{description}
        \item [Self-Optimization] covers self refining processes (e.g. calibrators, filters)~\citep{yeUnreliabilityExplanationsFewshot2022}.

        \item [Ensemble-Optimization] encompasses ensembling techniques used to generate more consistent answers by majority vote or ensembling decision process.
A good example  is \textbf{Self-consistency}~\citep{wangSelfConsistencyImprovesChain2022} which generates multiples prompts, verifies and votes; in practice, ensembling is often paired with
\textbf{verifier/PRM-based reranking} that scores final answers or intermediate steps before selection~\citep{cobbeTrainingVerifiersSolve2021, lightmanLetsVerifyStep2023}.

        \item [Iterative-Optimization] iteratively fine-tunes to improve reasoning process and answers~\citep{wangIterativelyPromptPretrained2022, huangLargeLanguageModels2022}.
    \end{description}

\section{Solving with agentic architectures \& inference-time strategies}
\label{sec_agentic_and_reasoning}
Agentic meta-architectures, using agents and tools, and inference-time strategies are \emph{different} strategies but both are dynamic mechanisms at inference.
The former provides a \emph{control-plane} (plan–act–observe–reflect; routing to tools and memories), while the latter is a \emph{compute-allocation knob} that decides how much “thinking’’ (branching, verification, retries) to spend on a given step or query.
Each can be used on its own: a simple agent may execute a straight-through plan without extra reasoning, and a single model can use CoT/self-consistency without any agent controller.
Used together, controllers can \emph{trigger deep thinking only when needed on its different LLM processing steps}, and reasoning-time methods improve the \emph{quality and faithfulness} of each agent step, leading to new high potential QA solving capabilities. A good example are different deep research architectures~\citep{xiongOutliningHeterogeneousRecursive2025, schmidgallAgentLaboratoryUsing2025, yangDocAgentMultiagentSystem2025, hanDeepResearcherTestTime2025} using inference-time reasoning for most complex steps in their answering generation process.

\subsection{Extended inference-time compute strategies}
\label{sec_reasoning_time}
Complex QA often benefits from allocating \emph{more thinking at inference} to refine its understanding, internal knowledge extraction, structuring its answering plan, evaluating hypothesis, selecting and refining answer.
This \emph{inference-time compute and reasoning} strategies can be scaled along three practical axes agreed upon across recent reviews:
\begin{enumerate}
    \item \textbf{Deliberate decoding and multi-sample search}: generate multiple chains/trees of thought (CoT/ToT/Graph-of-Thoughts), then \emph{self-consistency} or a verifier selects among candidates; cost grows with samples/branches~\citep{yaoTreeThoughtsDeliberate2023, wangSelfConsistencyImprovesChain2022}.
    \item \textbf{Process supervision}: learn or score intermediate steps with \emph{process reward models} (PRMs) or step-level preferences; improves stability on math/science and reduces unfaithful reasoning (e.g., verifiers and entailment-tree supervision)~\citep{lightmanLetsVerifyStep2023, dalviExplainingAnswersEntailment2021}.
    \item \textbf{Adaptive thinking allocation}: trigger “deep thinking’’ (extra tokens, tool calls, solver code, retrieval passes) \emph{only} when uncertainty/critique signals are high (e.g., a verifier rejects, or attribution is missing), via cascades/adaptive decoding~\citep{schusterConfidentAdaptiveLanguage2022, elbayadDepthAdaptiveTransformer2020}.
\end{enumerate}
Reasoning-time scaling complements (not replaces) training-time improvements (SFT, RLHF/RLAIF, DPO) and hybridization (retrievers/tools).
In deployment, it is common to route easy queries through “fast paths’’ and reserve intensive loops for hard cases to balance cost, latency and quality.

\subsection{Agentic meta-architectures (agents, tools)}
\label{sec_agentic}
\textbf{In this context, \emph{agents} are LLM based architectures having each a role and could plan, act, or coordinate other agents, while \emph{tools} are callable modules or services that perform specific operations delegated by the agent}. Tools include retrieval over text/graphs, web/file access, code/SQL execution, parsers, solvers, and verifiers (\autoref{HP5}–\autoref{HP8}, \autoref{HP11}, \autoref{HP16}). \emph{Multi‑agent} designs distribute roles (e.g., planner, researcher, writer, reviewer) to enable \emph{division of labor}, arbitration, and self‑critique \citep{venkatramanCollabStoryMultiLLMCollaborative2024, zhangChainofagentsLargeLanguage2024}; \emph{hierarchical} designs let a manager decompose work and spawn sub‑agents \citep{yangDocAgentMultiagentSystem2025}.

Across systems we observe a stable template that the controller instantiates:
\emph{(i)} plan \& route tasks (\autoref{HP4}, \autoref{HP12});
\emph{(ii)} invoke tools with structured calls (\autoref{HP5}–\autoref{HP8}, \autoref{HP16});
\emph{(iii)} verify \& revise with judges/PRMs and process-level supervision (\autoref{HP11}, \autoref{HP15});
\emph{(iv)} use \textbf{structured communication} and \textbf{typed artifacts} between roles (schemas rather than free chat);
\emph{(v)} persist memory/artifacts for reuse (\autoref{HP14});
\emph{(vi)} learn from dialog/feedback for alignment (\autoref{HP13}, \autoref{HP9});
\emph{(vii)} allocate \emph{reasoning-time} adaptively (self-consistency, cascades; \autoref{sec_reasoning_time}).
These patterns let readers map concrete systems to common building blocks without committing to a single implementation.

\noindent\textbf{Role- and structure-driven variants.} Beyond generic “roles,” multi-agent systems might explicitly encode agreed rich structures like \emph{Standard Operating Procedures (SOPs)} so that each role must produce structured intermediate artifacts (e.g., PRD $\rightarrow$ interface/design $\rightarrow$ code $\rightarrow$ tests) and communicate over a shared message bus (publish/subscribe) with \emph{executable feedback} (run tests, debug/repair). A representative design is \emph{MetaGPT}~\citep{hongMetaGPTMetaProgramming2023}, which encodes SOPs into prompts, constrains outputs to schemas (PRD, design docs, API specs), provides a shared message pool, and executes code to iteratively fix errors—reducing idle chatter and cascading hallucinations and reporting strong coherence/executability on software-engineering benchmarks. We treat SOP-style orchestration as a specialization of the control-plane above that tightens planning, verification, and persistence via schema-constrained artifacts, and is particularly useful when tasks naturally decompose into well-defined deliverables (e.g., long-form writing, software).

Hybridization (\autoref{sec_hybridLLMpatterns}) gives the \emph{data‑plane} building blocks (retrievers, executors, verifiers, memories). A \emph{meta‑architecture} supplies the \emph{control‑plane}: (a) \textbf{plan \& route} tasks (\autoref{HP4}, \autoref{HP12}); (b) \textbf{invoke tools} (\autoref{HP5}–\autoref{HP8}, \autoref{HP16}); (c) \textbf{verify \& revise} (\autoref{HP11}, \autoref{HP15}); (d) \textbf{persist artifacts/memory} (\autoref{HP14}); and (e) \textbf{learn from dialog/feedback} (\autoref{HP13}, \autoref{HP9}). This \emph{control‑plane/data‑plane} view \emph{does not define} specific systems; rather, our surveyed systems are \textbf{instances} that \emph{instantiate different points} in this design space. It underlies recent long‑form scientific QA/writing agents~\citep{shaoAssistingWritingWikipedialike2024, jiangUnknownUnknownsEngaged2024, xiongOutliningHeterogeneousRecursive2025, yangDocAgentMultiagentSystem2025, schmidgallAgentLaboratoryUsing2025, luAIScientistFully2024, yamadaAIScientistv2WorkshopLevel2025} and \emph{ecosystem‑level} platforms that organize evaluation and capitalization \citep{zhangAiXivNextGenerationOpen2025, schmidgallAgentRxivCollaborativeAutonomous2025} which we review in table \ref{tab:longform_writingsystems}.

\begin{table}
\centering
\small
\caption{Representative agentic systems as meta‑architectures (rows 1–6) and ecosystem‑level platforms for valuation \& capitalization (rows 7–8).}
\label{tab:longform_writingsystems}
\begin{tabularx}{\linewidth}{>{\raggedright\arraybackslash}p{0.15\linewidth} >{\raggedright\arraybackslash}p{0.15\linewidth} X}
  \toprule
  \textbf{Scope} & \textbf{System} & \textbf{Control‑plane $\rightarrow$ Data‑plane mapping (our patterns)} \\
  \midrule
  \multirow{6}{*}{\makecell[l]{Long‑form \\ generation \\ (planner \\ + workers)}}
  & \textbf{STORM}~\citep{shaoAssistingWritingWikipedialike2024} &
  Multi‑LLM \emph{perspective‑guided} research and outline; retrieve (\autoref{HP5}) $\rightarrow$ sectioned rafting; consolidation with citations; artifacts cached (\autoref{HP14}). \\
  & \textbf{Co‑STORM}~\citep{jiangUnknownUnknownsEngaged2024} &
  STORM + \emph{human‑in‑the‑loop moderator} guiding breadth/depth (\autoref{HP9}, \autoref{HP13}); same retrieve–draft–verify toolchain (\autoref{HP5}, \autoref{HP11}). \\
  & \textbf{WriteHere} (hierarchical planner)~\citep{xiongOutliningHeterogeneousRecursive2025} &
  \emph{Recursive} plan/execute across retrieval–reasoning–composition; task‑graph scheduling (\autoref{HP4}); section‑level arbitration; memory of intents/edits (\autoref{HP14}). \\
  & \textbf{Agent Laboratory}~\citep{schmidgallAgentLaboratoryUsing2025} &
  Stage‑wise controller for literature$\rightarrow$experiments$\rightarrow$paper; code tools and execution \autoref{HP7}, \autoref{HP8}); reviewer/rubric loops (\autoref{HP11}); human oversight (\autoref{HP9}). \\
  & \textbf{AI Scientist}~\citep{luAIScientistFully2024, yamadaAIScientistv2WorkshopLevel2025} &
  End‑to‑end \emph{propose–run–write–review}; parallel tree‑search experiments, VLM‑aided figure checks; generate–critique–revise loops (\autoref{HP11}, \autoref{HP15}). \\
  & \textbf{Deep Researcher (TTD‑DR)}~\citep{hanDeepResearcherTestTime2025} &
  Draft‑as‑diffusion controller: alternating \emph{denoising with retrieval} and \emph{component‑wise elf‑evolution} under LLM‑as‑judge calibration; closes cost–quality Pareto gaps vs.\ public DR agents by coupling lan–search–revise loops with attribution‑aware synthesis (\autoref{HP5}, \autoref{HP11}, \autoref{HP15}). \\
  \midrule
  \multirow{2}{*}{\makecell[l]{\textbf{Ecosystem level} \\ evaluation \& \\capitalization}}
  & \textbf{aiXiv}~\citep{zhangAiXivNextGenerationOpen2025} &
  \emph{Closed‑loop} platform for proposals/papers: RAG‑grounded multi‑agent reviewers (\autoref{HP5}, \autoref{HP11}), pairwise assessment, prompt‑injection defenses; APIs/MCP orchestrate heterogeneous agents (\autoref{HP7}). \\
  & \textbf{AgentRxiv}~\citep{schmidgallAgentRxivCollaborativeAutonomous2025} &
  Shared preprint server for \emph{agent labs} to upload/retrieve results; enables \emph{capitalization} of easoning methods and measurable cross‑lab improvement; persistent memory of field progress (\autoref{HP14}), with valuation signals feeding back into planners. \\
  \bottomrule
\end{tabularx}
\end{table}

Rows 1–6 are \emph{controller‑level} meta‑architectures that plan–act–observe–reflect over tools and retrieval to produce long‑form answers (articles, reports) \citep{shaoAssistingWritingWikipedialike2024, jiangUnknownUnknownsEngaged2024,yangDocAgentMultiagentSystem2025, schmidgallAgentLaboratoryUsing2025, luAIScientistFully2024, hanDeepResearcherTestTime2025}. Rows 7–8 are \emph{one meta level higher}: they provide an \emph{ecosystem} that \textbf{evaluates}, \textbf{refines}, and \textbf{capitalizes} outputs from many agents, driving field‑level complex QA via iterative review (aiXiv) and a shared memory of results (AgentRxiv) \citep{zhangAiXivNextGenerationOpen2025, schmidgallAgentRxivCollaborativeAutonomous2025}. As emphasized by our long‑form paragraph (§\ref{sec_evaluation}), multi‑agent collaboration and hierarchical planning improve coherence and factuality over single‑pass prompting \citep{venkatramanCollabStoryMultiLLMCollaborative2024, zhangChainofagentsLargeLanguage2024, yangDocAgentMultiagentSystem2025}.

\emph{Optimization and feedback;} the answering meta-architecture could be optimized online (at execution) or offline (between execution).
Deep Researcher~\citep{hanDeepResearcherTestTime2025}, when answering to complex question requiring long answers and planning, progressively select among multiple answers and refine its plan, content, and RAG resources at inference time, improving progressively its long answer.
For this, it requires to \emph{calibrate} its LLM-as-judge feedback against human preferences.
It shows Pareto gains in cost–quality trade-offs via denoising-with-retrieval and self-evolution.
Those meta-architectures can also be automatically \emph{optimized} offline with different feedback (pairwise preferences, rubric scoring, LLM‑as‑judge) via programmatic optimizers (e.g. DSPy) on \emph{compiled prompts, retrieval, and routes}, or even the code of agents and architecture using generative optimizers (e.g. Trace). 

Research‑grade long‑form QA seem to evolve to a robust stack of multi-agents with dynamic plan, retrieval, evaluation, alignment and improvement loop controller (rows 1–6), and some \emph{new research} head toward an higher level ecosystem layer (rows 7–8) that would \emph{evaluate} and \emph{capitalize} complex question answers; allowing to orchestrate research question exploration and knowledge progress capitalization~\citep{zhangAiXivNextGenerationOpen2025, schmidgallAgentRxivCollaborativeAutonomous2025}.

\section{Discussion: limitations and research topics for solving more complex QA and problems}\label{sec_limits_and_research}
In the architectural patterns section, we listed the most frequent topics identified as challenge or limits of LLMs.
After reviewing the collected literature and identifying different solutions in this study, some limits seem tougher research limits to enable more complex QA and problems solving:
\begin{itemize}
    \item Complex question decomposition, robust and explainable.
    \item The hallucination problem which limits a clear expectation of credibility/truthfulness in "Alignment to human expectation \& values in answer" (mitigations in practice couple retrieval-grounding with attribution metrics and verifier loops; metrics like FActScore/RAGAS are now commonly used; researchers warn when using LLM-as-judge without calibration~\citep{zhengJudgingLLMasaJudgeMTBench2023}).
    \item The scalability problem, which today includes not only training scale but also the \emph{allocation of compute at inference time} (reasoning-time) and the ability to mix fast paths and deep thinking under budget~\citep{dohanLanguageModelCascades2022, schusterConfidentAdaptiveLanguage2022}.
    \item Data availability \& quality which limits "domain adaptation \& task specialization" and "bias" .
    \item Data multi-sensitivity in LLM is now an \emph{active} area with defense-in-depth deployments (retrieval gating, guard agents, offsite/federated adaptation, privacy-preserving inference, unlearning); see \autoref{sec_data_sensitivity}~\citep{debenedettiDefeatingPromptInjections2025, xiaoOffsiteTuningTransferLearning2023, mireshghallahQuantifyingPrivacyRisks2022, chenTHEXPrivacyPreservingTransformer2022}.
\end{itemize}

\emph{From 2022 to 2025} practice has converged from answering with a single LLM inference to \emph{agentic, retrieval‑grounded, verifier‑guided} pipelines that allocate extra \emph{reasoning‑time} on hard queries~\citep{liangetal.HolisticEvaluationLanguage2022, wangSelfConsistencyImprovesChain2022, cobbeTrainingVerifiersSolve2021}. Controllers plan–act–observe–reflect over tools (RAG, code, search), while verifiers/PRMs and self‑consistency select robust chains; for long‑form research/writing agents, this pattern materially improves coherence and factuality~\citep{shaoAssistingWritingWikipedialike2024, jiangUnknownUnknownsEngaged2024, xiongOutliningHeterogeneousRecursive2025, hanDeepResearcherTestTime2025}.

\subsection{Hallucination \& credibility}
Early debates about Galactica~\citep{taylorGalacticaLargeLanguage2022} and ChatGPT~\citep{rudolphChatGPTBullshitSpewer2023} shade the light of limits and credibility of such language models concerning hallucination.
It generates plausible-looking statements that are irrelevant or factually incorrect.
It predicts without giving clues about which part of a false claim goes wrong, even sources given are not trustworthy.
It even has difficulty to learn correct associations between entities from factual text corpus (e.g. Wikipedia).
Explainability of an answer with a supported citation is a pointer but does not mean it is true.
By late 2025, mitigations that combine \emph{retrieval grounding} with \emph{verifier/critic loops}, and that report \emph{attribution/groundedness} metrics alongside accuracy are widely adopted in complex QA; when using LLM‑as‑judge to scale evaluation, \emph{calibration against human preferences} is required to reduce bias~\citep{menickTeachingLanguageModels2022, nakanoWebGPTBrowserassistedQuestionanswering2022, cobbeTrainingVerifiersSolve2021, wangSelfConsistencyImprovesChain2022, zhengJudgingLLMasaJudgeMTBench2023}.
We have identified different topics of research to address the challenge of hallucinations such as:
\begin{itemize}
    \item Robust training \& prompting (self-consistency, context optimization, prompt tuning, denoising...).
    \item Hallucination detection~\citep{zhouDetectingHallucinatedContent2021}.
    \item Providing references, traceability, faithful explanation logic~\citep{chenLORENLogicRegularizedReasoning2022} or the emerging field of entailment tree explanation~\citep{ribeiroEntailmentTreeExplanations2022, liuRLETReinforcementLearning2022} or automated fact-checking~\citep{guoSurveyAutomatedFactChecking2022}; in retrieval-grounded settings, practitioners also track \emph{FActScore}/\emph{RAGAS}-style attribution and groundedness.
    \item Identifying faithfulness performance per tasks/domain~\citep{liangetal.HolisticEvaluationLanguage2022} and biases, to better ensemble experts~\citep{choubeyMoFEMixtureFactual2021}.
    \item Contrastive learning to reduce hallucination~\citep{sunContrastiveLearningReduces2022}.
    \item Reinforcement learning \emph{seems} one of the strongest area of research to reduce hallucination in QA.
\end{itemize}

\subsection{Compute, scaling... Costs}
More than 8 million TPUv4 hours was the time taken to train PalM 540B parameters model~\citep{tayTranscendingScalingLaws2022}.
For a far smaller model "T5 11B paremeters" and its variants, the cost of the project is estimated \$10 millions~\citep{sharirCostTrainingNLP2020}.
Those models continue to scale.
Time of compute for training is therefore reserved to few organization.
Operational costs for usage (inference) are less impressive~\citep{liangetal.HolisticEvaluationLanguage2022} but limits the use cases considering inference latency and minimal required hardware.
We can apply standard model size reductions like quantization, distillation, pruning, early stopping at training...
Different research avenues try to reduce this computing and costs required and inverse this scaling law :
\begin{itemize}
    \item \textbf{Frozen PLM techniques}: we presented in previous sections prompt tuning and parameter‑efficient tuning (PEFT) in our first survey in 2022; these continue to be the most practical path to specialize models under tight budgets (e.g., QLoRA, LoftQ)~\citep{dettmersQLoRAEfficientFinetuning2023, liLoftQLoRAFineTuningawareQuantization2023}.
    \item \textbf{Retrieval augmented LLM}: keeping a maximum of information out of model while making it easy to access and update without any re-train has an important potential but is often less efficient and may require equal computation when comparing "total additional answer generation time" vs "training", new techniques try to close the gap~\citep{dejongPrecomputedMemoryOnthefly2023}.
    \item \textbf{Scaling in-context learning}: in-context learning highly improves tasks efficiency without re-training but is limited by maximum length input constraints due to quadratic complexity in computation and memory; different techniques allow efficient scaling~\citep{pressTrainShortTest2022, haoStructuredPromptingScaling2022, choPromptAugmentedLinearProbing2022, martins$infty$formerInfiniteMemory2022}.
    \item \textbf{Serving/runtime efficiency}: employing FlashAttention-2/3 (bandwidth-efficient attention), paged attention/KV caching (compact context reuse), speculative decoding (parallel token prediction), alongside sparse \textbf{Mixture-of-Experts} routing, reduces per-answer cost without compromising model quality.
    \item \textbf{Mixture of denoisers}: in "Transcending Scaling Laws with 0.1\% Extra Compute"~\citep{tayTranscendingScalingLaws2022}, same model is train at half the budget to reach the same performance using mixture-of-denoisers (" U-PaLM achieves the same performance as the final PaLM 540B model at around half its computational budget - saving approx. 4.4 million TPUv4 hours").
UL2, proposes an unification of LM learning paradigms~\citep{tayUL2UnifyingLanguage2022}.
    \item \textbf{Improve pruning techniques}: \citet{frantarSparseGPTMassiveLanguage2023} reduce LLM size by more than 50\% with only one-shot, without any retraining, and nearly no loss in accuracy.
    \item \textbf{Mixture of experts, parameters sharing and routing techniques}: "Switch Transformers: Scaling to Trillion Parameter Models with Simple and Efficient Sparsity"~\citep{fedusSwitchTransformersScaling2022} demonstrated in 2022 the usage of internal routing to expert layers in large language models to limit compute to part of the whole model to allow to scale model without augmenting compute. By the end of 2025, it is used in many architecture and research still improve it.
    \item \textbf{Knowledge distillation} improvement~\citep{blakeneyReduceReuseRecycle2022, zaheerTeacherGuidedTraining2022, wahleCohesiveDistillationArchitecture2023} and \textbf{dynamic composition} of model~\citep{xuSurveyDynamicNeural2022} to compose optimal and smaller models; \textbf{RL‑free preference optimization} (e.g., DPO/ORPO/KTO) is widely used in post‑training to align to preferences at lower complexity~\citep{rafailovDirectPreferenceOptimization2023, hongORPOMonolithicPreference2024, ethayarajhModelAlignmentProspect2024}.
    \item \textbf{Adaptive computation and training}: all samples are equally computed in standard training. Easy samples could be less worked out than hard ones; adaptive computing enables sample‑dependent computation (e.g., CALM)~\citep{schusterConfidentAdaptiveLanguage2022}.
Chinchilla~\citep{hoffmannTrainingComputeOptimalLarge2022} demonstrated that we can highly reduce inference budget while improving accuracy by using the same training budget with much more data on a much smaller LLM.
    \item \textbf{Reasoning‑time compute and verifiers}: move some “scaling” from training into inference via cascades and confidence triggers~\citep{dohanLanguageModelCascades2022, schusterConfidentAdaptiveLanguage2022}, combine multi‑sample chains with \emph{self‑consistency}~\citep{wangSelfConsistencyImprovesChain2022}, score intermediate steps with \emph{process‑level verifiers/PRMs}~\citep{cobbeTrainingVerifiersSolve2021}, and—when training resources are available—use \emph{reinforcement fine‑tuning} (e.g., GRPO) to explicitly incentivize reasoning~\citep{guoDeepSeekR1IncentivizesReasoning2025}.
    \item \textbf{Dedicated hardware}: language models are typically accelerated by running on GPU but an area of research investigate dedicated hardware (e.g. FPGA, ASICS, ) for saving energy and costs~\citep{hongDFXLowlatencyMultiFPGA2022}.
\end{itemize}

\subsection{Data availability \& quality}
The skills and training datasets section, as well as human feedback, highlighted the need for specialized data and of high quality, to acquire skills and domain knowledge as well as to calibrate to requester intents.
Large language models requires huge volume of data to develop each skills and domains targeted.
Wealthy organization can spend millions to clean or produce data but even them are limited.
\begin{itemize}
    \item \textbf{Frugal and rich data with AI supervision}: we saw in previous sections techniques like dynamic least-to-most requiring far less data for training while improving accuracy and skills, or active learning identifying best examples for improvement\citep{diaoActivePromptingChainofThought2023}.
Those data could be more and more automatically generated like in Auto-CoT~\citep{zhangAutomaticChainThought2022}, CAI~\citep{baiConstitutionalAIHarmlessness2022}.
    \item \textbf{Simulation, distillation and code interpreter} (program execution): simulation~\citep{liuMindsEyeGrounded2022, jacksonNaturalLanguageSimulations2023}, existing models~\citep{wahleCohesiveDistillationArchitecture2023}, and code interpreters~\citep{haluptzokLanguageModelsCan2022} can provide infinite examples in different domains of logic and knowledge.
We also uncovered that code interpreter allow to learn logic transferable to many domains.
    \item \textbf{More signals}: better leverage all available symbolic data which have clear signals, already structured (Linked Open Data) or re-structured~\citep{yuanReStructuredPretraining2022}.
    \item \textbf{Open dataset}: QA and IR field has continuously progressed through the availability of existing and new open datasets~\citep{jerniteDataGovernanceAge2022} for target skills and knowledge.
    \item \textbf{Teacher‑generated “textbooks/datasets” for reasoning}: high‑quality synthetic curricula distilled from stronger teachers continue to be effective for math/coding/reasoning specialists~\citep{gunasekarTextbooksAreAll2023}.
\end{itemize}

\subsection{Data multi-sensitivity usage \& protection}\label{sec_data_sensitivity}
Sensitive data in CQA systems spans personal data, trade secrets, regulated information, and safety‑critical content; a \emph{single item can carry multiple sensitivities}.
Literature outlines the need for a defense-in-depth approach (multiple layers of protection) which we organize here across the data plane, control plane, and learning stack.
Leakage risks arise at \emph{training}, \emph{indexing/retrieval}, and \emph{generation/tool‑use} time, through membership/attribute inference and inversion attacks, prompt‑injection jailbreaks targeting RAG or agents, and poisoning of retrieval indexes~\citep{mireshghallahQuantifyingPrivacyRisks2022, debenedettiDefeatingPromptInjections2025}.

Accordingly, state‑of‑the‑art deployments use \emph{defense‑in‑depth} with controls aligned to a threat‑model. The checklist below summarize patterns recommended across privacy and security evaluations and deployment guidance for LLM systems~\citep{mireshghallahQuantifyingPrivacyRisks2022, debenedettiDefeatingPromptInjections2025, xiaoOffsiteTuningTransferLearning2023, chenTHEXPrivacyPreservingTransformer2022}:
(1) keep sensitive data \emph{out of weights} and behind a policy‑enforced retriever;
(2) sanitize/privatize representations and \emph{gate} retrieval;
(3) surround RAG/agents with guardrails and verifiers;
(4) adapt with offsite/federated training or cryptographic inference where needed;
(5) support unlearning and \emph{auditable} provenance.

\paragraph{Data‑plane controls (store/index/evidence).}
\begin{itemize}
 \item \textbf{Access control at retrieval time.} Keep sensitive corpora outside the model and enforce row/field/passage‑level policies in the retriever (ABAC/LBAC) so only authorized content is retrieved into the context window; this avoids baking access decisions into model weights and supports per‑user purpose limitation.
 \item \textbf{Segmentation by sensitivity / tenancy.} Where policies and risk profiles diverge, maintain separate indices or fine-tuned endpoints per sensitivity tier or tenant, with a front-end router enforcing who may query which segment. This complements retrieval-time access control rather than replacing it.
 \item \textbf{Privacy‑preserving embeddings and redaction.} Prevent embedding‑inversion and unintended memorization by (i) sanitizing/PII‑redacting at index time and (ii) using privacy‑preserving representation learning (e.g., TextHide, TextFusion) that makes recovering plain text from embeddings substantially harder~\citep{huangTextHideTacklingData2020, zhouTextFusionPrivacyPreservingPretrained2022, kimPrivacypreservingTextEmbedding2022}.
\end{itemize}

\paragraph{Control‑plane and runtime guards (retrieval \texorpdfstring{$\rightarrow$}{→} generation).}
Controls that mediate how retrieved context influences tools and generation:
\begin{itemize}
 \item \textbf{Guardrails against prompt‑injection and policy violations.} Enforce multi‑stage filters and guard agents around search/tool use, and apply \emph{retrieval gating}/verifier‑in‑the‑loop so untrusted context cannot steer tool calls or exfiltrate secrets~\citep{debenedettiDefeatingPromptInjections2025, xiangGuardAgentSafeguardLLM2024}.
 \item \textbf{Self‑/corrective‑RAG.} Couple generation with critique and retrieval‑backed verification to reduce ungrounded disclosure; keep attributions for auditability (see \S\ref{sec_evaluation}).
\end{itemize}

\paragraph{Learning \& compute with minimal exposure.}
Mechanisms to adapt models and run inference while minimizing raw-data exposure:
\begin{itemize}
 \item \textbf{Offsite/partitioned fine‑tuning.} Adapt models to private data without exposing full weights or raw data (e.g., Offsite‑Tuning)~\citep{xiaoOffsiteTuningTransferLearning2023}.
 \item \textbf{Federated learning.} Keep data at source and aggregate updates centrally to avoid raw‑data sharing for domain adaptation~\citep{chenFedMatchFederatedLearning2021}.
 \item \textbf{Cryptography \& secure computation.} Where latency budgets allow, use privacy-preserving transformer inference/training via homomorphic encryption/secure protocols (e.g., THE-X implements encrypted Softmax/GELU/LayerNorm paths)~\citep{chenTHEXPrivacyPreservingTransformer2022}; include \emph{functional encryption} for role-restricted queries when applicable~\citep{xuPrivacyPreservingMachineLearning2021}.
 \item \textbf{Right‑to‑be‑forgotten.} Support \emph{machine unlearning} to remove specific data influence from adapted models or indices~\citep{bourtouleMachineUnlearning2020}.
\end{itemize}

\paragraph{Propositions to use blockchain}
Used as an audit or coordination layer rather than a data store, a permissioned blockchain can add \emph{verifiable provenance and usage accounting} without placing raw data on‑chain: (i) immutable logs of retrieval/training/inference \emph{events}, (ii) access‑policy enforcement and incentives via smart contracts, and (iii) privacy‑preserving attestations (e.g., zk‑proof‑backed claims) that a request complied with policy~\citep{gerenBlockchainLargeLanguage2025}.
These patterns complement (not replace) store‑level access control and cryptographic/privacy mechanisms above.

\textbf{Governance and safety context}. Regulatory requirements (e.g., logging, risk management) are tightening; in parallel, organizations increasingly combine policy filters/guard agents with private adaptation methods (offsite/federated) for multi‑tenant RAG/agent systems~\citep{xiaoOffsiteTuningTransferLearning2023, xiangGuardAgentSafeguardLLM2024}.

\subsection{Decomposition of very complex QA and explainability}
The need for decomposition in complex questions is central in their solving process because it breaks down a non solvable complex question (problem) down to solvable questions.
As we have seen in chain-of-thought and dynamic least-to-most, those traceable steps improve solving capacities of a given model but also makes the answer auditable, truthful and explainable in case of errors.
However nearly all examples of papers reviewed related to decomposition are factoid questions.
\citet{duaSuccessivePromptingDecomposing2022}  used "\textit{Who kicked the longest field goal in the first half?}" as the main example of complex question, it is a factoid question.
What if we ask "\textit{What are the most adapted LM hybrid architectures to answer complex questions ?}".
That would be a non-factoid question, requiring multiple sources, reasoning...
ChatGPT could provide an answer but reasoning would be nor explainable nor a robust decomposition process with an acceptable and auditable scientific methodology.
We could learn this decomposition behaviour by cloning a human process~\citep{yangChainThoughtImitation2022} or learn to discover it through a human collaboration.
Iterated decomposition~\citep{reppertIteratedDecompositionImproving2023}, a human-in-the-loop workflow for process supervision using a language model allows to address new type of problems and enables users to inspect and intervene in the LM’s reasoning process.
In 2025, \emph{long‑form agentic systems} (e.g., STORM/Co‑STORM, WriteHere, Deep Researcher) improved decomposition and revision on open‑ended research/writing tasks by coupling retrieval, planning, and verifier‑guided revision~\citep{shaoAssistingWritingWikipedialike2024, jiangUnknownUnknownsEngaged2024, xiongOutliningHeterogeneousRecursive2025, hanDeepResearcherTestTime2025}.
Open challenges remain: robust planning under ambiguity, reproducible evaluation beyond task‑specific win‑rates (tie‑back to HELM’s multi‑metric view), and controlling inference‑time cost without hurting factuality~\citep{liangetal.HolisticEvaluationLanguage2022, wangSelfConsistencyImprovesChain2022, cobbeTrainingVerifiersSolve2021}.
These approaches still require substantial human expert feedback, domain‑specific data and calculations; this is an important limitation to generalize the resolution of complex non‑factoid questions across domains and methodologies.

\section{Conclusion}
In this paper, we present a comprehensive survey of language model hybrid architectures for answering complex questions.
We review the various skills required and typical approach, datasets and metrics that are used, the current limits of large language models for complex QA, the potential of hybrid and meta architectures, better training, inferring, and prompting strategies for this goal.
We also identify the main challenges and research avenues for solving more complex questions including knowledge capitalization.
We identify the need to address multi-sensitivity data defense-in-depth in LLM architectures.
Finally, we outline research topics and highlight the potential of exploration in this field.
Compared to the initial review dated 2022, the default architecture for robust complex QA is now \emph{agentic (with tools), retrieval‑grounded, verifiable and aligned}: plan–act–observe–reflect controllers that orchestrate tools and retrieval (including graph‑structured search), allocate extra \emph{reasoning‑time} when needed (with process supervision), and report evidence/attribution alongside answers.
This paper aimed to provide a comprehensive and useful resource for readers interested in the development of complex non-factoid question answering.

\appendix

\begin{table}[htbp]
\centering
\footnotesize
\resizebox{\textwidth}{!}{%
\begin{tabular}{llrrrrrrrrr}
\hline
Task (BBEH) & Random & Llama & Gemma2 & Gemini 2.0 & Gemini 2.0 & GPT-4o & Distill & DeepSeek & o3-mini \\
- skill(s) &  & 3.1 8B IT & 27B IT & Flash-Lite & Flash & & R1 Qwen 32B & R1 & (high) \\
\multicolumn{1}{c}{} & Score (\%) & Score (\%) & Score (\%) & Score (\%) & Score (\%) & Score (\%) & Score (\%) & Score (\%) & Score (\%) \\
\hline
BoardgameQA& 33.3 & 31.5 & 39.5 & 29.5 & 42.5 & 41.0 & 36.0 & \textbf{75.5} & 53.0 \\
- multi-step + in-context-learning & \\
Boolean Expressions & 20.0 & 18.0 & 25.0 & 24.0 & 27.0 & 22.5 & 17.5 & 55.5 & \textbf{67.0} \\
- multi-step & \\
Causal Understanding & 38.0 & 37.0 & 45.5 & 52.5 & 52.0 & 54.0 & \textbf{54.5} & \textbf{54.5} & 54.0 \\
- causal, logical, counterfactual & \\
DisambiguationQA & 21.0 & 36.7 & 45.0 & 50.0 & 48.3 & 51.7 & 52.5 & 50.0 & \textbf{58.3} \\
- commonsense, linguistics & \\
Dyck Languages & 1.4 & 4.5 & 2.0 & 6.5 & 14.0 & 8.0 & 18.0 & \textbf{56.0} & 55.0 \\
- errors in reasoning & \\
Geometric Shapes & 6.2 & 25.5 & 31.0 & 30.0 & 35.0 & 22.5 & 4.5 & 1.5 & \textbf{52.5} \\
- dealing-with-distractors & \\
Hyperbaton & 0.0 & 2.0 & 4.0 & 6.5 & 4.5 & 7.5 & 3.0 & 6.0 & \textbf{32.0} \\
- inductive reasoning + against-strong-prior & \\
Movie recommendation & 10.0 & 30.0 & 40.0 & 51.5 & 59.5 & 61.0 & 46.0 & 59.5 & \textbf{84.0} \\
- knowledge association & \\
SARC Triples & 12.5 & 16.5 & 21.0 & 27.0 & 37.5 & \textbf{38.5} & 22.0 & 28.5 & 24.0 \\
- compositional reasoning & \\
Word Sorting & 4.3 & 2.5 & 3.5 & 12.5 & 26.0 & 22.0 & 36.0 & 68.0 & \textbf{77.5} \\
- against-strong-prior + finding-errors & \\
Web of Lies & 5.5 & 5.5 & 6.5 & 14.0 & 18.5 & 14.5 & 13.0 & 29.5 & \textbf{43.0} \\
- multi-step & \\
SportQA & 0.0 & 1.5 & 10.0 & 18.5 & 23.0 & 25.0 & 19.5 & \textbf{29.0} & 26.5 \\
- knowledge + compositional reasoning & \\
Shuffled objects & 14.3 & 9.5 & 12.0 & 15.0 & 9.0 & 14.0 & 2.0 & 6.0 & \textbf{49.5} \\
- long-context  + long-range-dep + needle-in-haystack & \\
Buggy tables & 0.0 & 0.0 & 0.5 & 1.5 & 3.5 & 0.5 & 0.5 & 4.5 & \textbf{59.5} \\
- in-context-learning + needle-in-haystack & \\
Temporal sequences & 0.0 & 9.5 & 1.5 & 1.0 & 0.5 & 0.0 & 0.5 & 0.0 & \textbf{68.5} \\
- temporal reasoning & \\
\hline
\end{tabular}%
}
\caption{Big bench \textbf{extra hard} tasks for LLM compared across recent SOTA models.}
\label{tab:bbeh}
\end{table}

\bibliography{BibSurveyCQALM}\label{sec_biblio}

\end{document}